\documentclass[Afour,sageh,times]{sagej}

\usepackage{moreverb,url}

\usepackage[colorlinks,bookmarksopen,bookmarksnumbered,citecolor=red,urlcolor=red]{hyperref}

\newcommand\BibTeX{{\rmfamily B\kern-.05em \textsc{i\kern-.025em b}\kern-.08em
T\kern-.1667em\lower.7ex\hbox{E}\kern-.125emX}}

\def\volumeyear{2016}

\usepackage[para,online,flushleft]{threeparttable}
\usepackage{textcomp}
\usepackage{booktabs}
\usepackage{multirow}
\usepackage{adjustbox}
\usepackage{graphicx}
\usepackage{siunitx}
\usepackage[utf8]{inputenc}
\usepackage{rotating}
\usepackage{ulem}
\usepackage{natbib}
\newtheorem{problem}{Problem}
\newtheorem{subproblem}{Problem}[problem]
\newtheorem{formulation}{Formulation}
\newtheorem{subformulation}{Formulation}[formulation]
\newtheorem{theorem}{Theorem}
\newtheorem{lemma}[theorem]{Lemma}

\newtheorem{definition}{Definition}

\newtheorem{remark}{Remark}
\newtheorem{example}{Example}
\newtheorem{subexample}{Example}[example]

\usepackage{graphicx}
\usepackage{amsmath, amssymb, mathtools}
\usepackage{siunitx}
\usepackage{subcaption}
\usepackage{caption}
\usepackage{color}
\usepackage{theoremref}
\usepackage[noadjust]{cite}
\usepackage{algorithm}
\usepackage[noend]{algpseudocode}

\usepackage{enumitem}
\usepackage{comment}
\usepackage{makecell}
\usepackage{tabularx}
\usepackage{float}

\usepackage{soul}

\newcommand{\xuan}[1]{\textcolor{black}{#1}}
\newcommand{\jimrev}[1]{\textcolor{black}{#1}}

\newcommand{\xuanrev}[1]{\textcolor{black}{#1}}

\title{Towards Tighter Convex Relaxation of Mixed-Integer Programs: Leveraging Logic Network Flow for Task and Motion Planning}

\author{Xuan Lin\affilnum{1}, Jiming Ren*\affilnum{1}, Yandong Luo*\affilnum{1}, Weijun Xie\affilnum{2}, and Ye Zhao\affilnum{1}}

\affiliation{\affilnum{1}George W. Woodruff School of Mechanical Engineering, Georgia Institute of Technology, USA\\
\affilnum{2}Milton Stewart School of Industrial and Systems Engineering, Georgia Institute of Technology, USA \\
$^*$These authors contributed equally to this work.
}

\corrauth{Ye Zhao, Georgia Institute of Technology, Atlanta, GA 30332, USA.\\
Email: yezhao@gatech.edu}

\begin{document}

\begin{abstract}
\xuan{This paper proposes an optimization-based task and motion planning framework, named ``Logic Network Flow," that integrates temporal logic specifications into mixed-integer programs for efficient robot planning. Inspired by the Graph-of-Convex-Sets formulation, temporal predicates are encoded as \jimrev{polyhedral} constraints on each edge of a network flow model, instead of as constraints between nodes in traditional Logic Tree formulations. 
We further propose a network-flow-based Fourier-Motzkin elimination procedure that removes continuous flow variables while preserving convex relaxation tightness, 
leading to provably tighter convex relaxations and fewer constraints than Logic Tree formulations. 
For temporal logic motion planning with piecewise-affine dynamical systems, comprehensive experiments across vehicle routing, multi-robot coordination, and temporal logic control on dynamical systems using point-mass and linear inverted pendulum models demonstrate computational speedups of up to several orders of magnitude, \xuanrev{alongside reduced memory consumption}. Hardware demonstrations with quadrupedal robots validate real-time replanning capabilities under dynamically changing environmental conditions. The project website is at \url{https://logicnetworkflow.github.io/}.}
\end{abstract}

\maketitle

\section{Introduction}
\label{Sec:introduction}

\jimrev{Task and motion planning (TAMP) with temporal logic specifications enables robots to execute complex tasks with formal guarantees of motion safety and task satisfaction \citep{plaku2016motion, zhao2024survey, li2021reactive, he2015towards, shamsah2023integrated}. Temporal logics provide expressive specification languages for describing high-level robotic behaviors that extend beyond reaching predefined goal states, allowing users to encode sequential, conditional, safety, persistence, and liveness requirements in a mathematically rigorous manner. }

\jimrev{Among the most widely used temporal logics, Linear Temporal Logic (LTL) specifies qualitative temporal relationships over discrete event sequences and is commonly employed for high-level task planning. Metric Temporal Logic (MTL) extends LTL by introducing explicit timing constraints, enabling the specification of deadlines and bounded-time requirements. Signal Temporal Logic (STL) further generalizes these capabilities by reasoning over continuous-valued signals and is particularly suitable for integrating task-level specifications with robot dynamics and control. Existing approaches typically solve LTL-constrained planning using automata-based methods, whereas MTL and STL specifications are formulated as mixed-integer convex programs (MICPs) \citep{cardona2023mixed, sun2022multi}. These formulations enable the simultaneous optimization of task sequencing and motion generation while preserving the semantics of the temporal logic specifications.}

\jimrev{Despite their strong expressive power and formal guarantees, temporal-logic-constrained motion planning remains computationally challenging. The underlying optimization problems are NP-hard, and although modern MICP solvers based on Branch and Bound (B\&B) often solve practical instances efficiently, their worst-case computational complexity remains exponential. Consequently, even with recent advances in improving solver performance \citep{kurtz2021more, kurtz2022mixed}, solving large-scale temporal logic planning problems frequently requires minutes or even hours, rendering them impractical for real-time robotics applications.}

\begin{figure}[t!]
    \centering
    \includegraphics[width=0.48\textwidth]{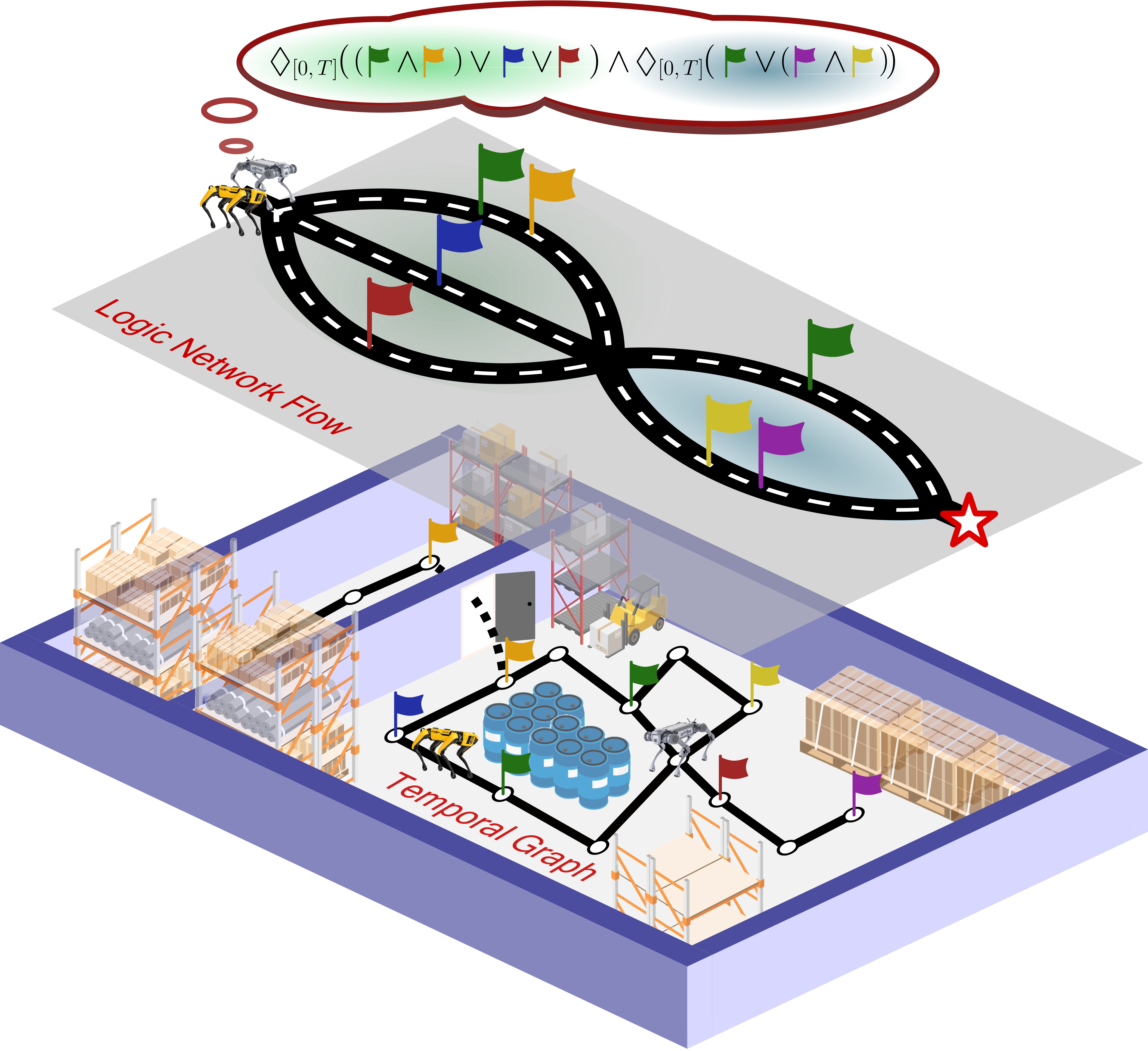}
     \caption{Conceptual illustration of multi-robot coordination in a factory warehouse using Logic Network Flow (LNF). The LNF representation of a conjunctive-disjunctive temporal logic specification is visualized at the top, with colored flags indicating different predicates on the LNF edges. Two quadrupedal robots navigate through the warehouse, visualized at the bottom, with the underlying temporal graph overlaid on the factory floor. Flags on the temporal graph indicate locations where specific task predicates are satisfied, showing the correspondence between the logic specification and physical locations.}
\label{fig:example4}
\end{figure} 

In this paper, we address this challenge by introducing a novel optimization formulation named ``Logic Network Flow (LNF)." Inspired by the Graph-of-Convex-Sets (GCS) formulation \citep{marcucci2024graphs}, we transform temporal logic specifications into network flow constraints to achieve tighter convex relaxations. Compared to the conventional Logic Tree (LT) approaches \citep{wolff2014optimization, raman2014model}, which place temporal logic predicates at leaf nodes of a tree structure, the novelty of LNF lies in its usage of flow conservation constraints to correlate the predicates placed on the edges of a network flow.

Furthermore, we propose a network-flow-based Fourier-Motzkin procedure that eliminates flow variables, replacing them with a series of inequality constraints that provably tighten the convex relaxations compared to the LT formulations. This effectively projects the original formulation onto a lower-dimensional space, leading to our final formulation that maintains tighter convex relaxations without introducing additional variables. This is in contrast with the GCS formulation \citep{marcucci2024graphs}, which achieves tighter convex relaxations at the expense of introducing additional variables compared to baseline formulations.

Through achieving tighter convex relaxations without introducing additional variables, our formulation demonstrates better computational performance in the B\&B process and faster discovery of better incumbent solutions with computational speedups of up to several orders of magnitude \xuanrev{and reduced memory consumption} compared to the benchmark methods. In addition, \jimrev{through} integrating LNF with dynamical systems, our framework offers greater flexibility than traditional motion planning approaches such as \citep{kuindersma2016optimization, marcucci2023motion, lin2019optimization}, which typically require predefined goal states. Instead, our formulation synthesizes motion plans directly from temporal logic specifications, allowing behaviors that cannot be expressed as simply reaching fixed destinations, while still accommodating complex dynamics such as Piecewise Affine (PWA) \xuanrev{dynamics} \citep{sontag1981nonlinear}.

We evaluate our formulation by comparing it against the standard LT approach for temporal logic motion planning integrated with PWA \xuanrev{dynamics}, as reviewed by \citep{belta2019formal}. Specifically, we experiment on robotics problems that often go beyond the simple examples demonstrated in prior temporal logic planning literature, 
including Vehicle Routing Problems with Time Windows (VRPTW) for multi-robot coordination \citep{karaman2008vehicle}, optimal motion planning under temporal logic specifications using point-mass \citep{belta2019formal} and linear inverted pendulum models \citep{gu2024walking}, and multi-robot search-and-rescue scenarios. The formulation is further validated through hardware experiments with quadrupedal robots demonstrating real-time replanning capabilities under dynamically changing environmental conditions.

Our main contributions are summarized as:
\begin{enumerate}
\item A novel optimization formulation, LNF, that provides provably tighter convex relaxations and fewer constraints compared to existing LT approaches, resulting in computational speedups up to orders of magnitude \xuanrev{and reduced memory consumption}.
\item A network-flow-based Fourier-Motzkin elimination procedure that reduces the number of continuous variables and constraints for our temporal logic formulations, while being applicable to broader GCS formulations.
\item Methods for integrating LNF with \xuanrev{PWA dynamics}, including specialized discrete graph formulations. 
\item Extensive experiments to validate computational speed advantages in various robotic systems \xuanrev{formulated using PWA dynamics}, as well as hardware demonstrations using quadrupedal robots. 
\end{enumerate}

A conference version of this work is presented in \citep{lin2025optimization}. The current article significantly advances the original paper in several critical aspects. First, we introduce a network-flow-based Fourier-Motzkin elimination procedure that substantially improves computational efficiency by reducing the number of variables and constraints \xuanrev{(Section~\ref{sect:variable_elimination})}. Second, we extend the framework to handle \xuanrev{general PWA dynamics} under temporal logic constraints, while maintaining the Dynamic Network Flow (DNF) formulation as a specialized formulation for planning on Temporal Graphs \xuanrev{(Section~\ref{Sec:DNF})}. Third, we provide rigorous theoretical analysis including formal proofs that our LNF formulation achieves tighter convex relaxations compared to the LT formulations \xuanrev{(Section~\ref{Sec:proof_of_tighter_relaxation})}, as well as proofs of soundness and completeness \xuanrev{(Appendix~\ref{appendix_soundness})}. \xuanrev{Fourth, we provide extensive additional experiments with no counterpart in the conference paper, including substantially broader experiments on temporal graphs (Section~\ref{Sec:temporal_graph_experiment}), a realistic multi-robot search-and-rescue application minimizing terrain-roughness-induced falling risk and time-varying overheating risk (Section~\ref{Sec:search_rescue}), new experiments on point-mass systems (Section~\ref{sec:point_mass_navigation}) and linear inverted pendulum models (Section~\ref{sec:LIP_experiment}), a dedicated study of infeasibility detection (Section~\ref{Sec:infeasibility}), two ablation studies isolating the source of LNF's tighter convex relaxation (Sections~\ref{Sec:ablation_encoding} and~\ref{Sec:ablation_source}), and two hardware demonstrations with quadrupedal robots covering multi-robot delivery under sequential ordering and collision-avoidance constraints and single-robot real-time replanning under human perturbations and unforeseen obstacles (Section~\ref{sec:experiment_hardware}). These experiments demonstrate substantial speedups over the LT baseline across all problem scales, along with reduced memory consumption.}

The remainder of this paper is organized as follows. Section \ref{Sec:related_work} reviews prior work on motion planning under temporal logic constraints. Section \ref{Sec:background} presents the mathematical preliminaries for formulating temporal logic constraints into MILPs. Section \ref{Sec:LNF} introduces our LNF formulation as well as the network-flow-based Fourier-Motzkin elimination procedure. Section \ref{Sec:DNF} describes general approaches for integrating LNF with PWA \xuanrev{dynamics} as well as specialized formulations using DNF. Section \ref{Sec:experiments} presents computational results and experimental validation through simulations and hardware demonstrations. We conclude in Section \ref{Sec:conclusion} with a discussion of limitations and future work.

\section{Related Work}
\label{Sec:related_work}
\subsection{\xuan{Automata-based Methods for Temporal Logic Planning}}

Originating in software verification~\citep{baier2008principles}, formal methods based on temporal logic provide mathematical guarantees of the correctness of complex task specifications. A range of robotic studies have therefore leveraged temporal-logic-based methods for TAMP, such as control synthesis over dynamic systems~\citep{belta2017formal}, multi-agent coordination~\citep{kloetzer2009automatic}, and human-robot interaction~\citep{kress2021formalizing}.

Automaton construction is a common approach to synthesize the reachability set of a robot to find feasible motion plans that satisfy temporal-logic specifications, where the specification is converted to a finite state machine featuring the actions to satisfy the logical requirements~\citep{gastin2001fast}. 
However, building automata leads to significant computational costs, with the automaton size growing exponentially as the specification becomes more complex~\citep{wolff2014optimization}. 
This scalability issue becomes particularly challenging for large-scale robotic systems, motivating dedicated algorithms such as~\citep{kantaros2020stylus}. 
When limited to a fragment of the LTL specifications, the satisfiability of temporal logic games can be verified in polynomial time complexity with respect to the problem size using reactive GR(1) synthesis \citep{piterman2006synthesis, kress2009temporal, liu2013synthesis}, and when satisfiable, the corresponding automaton is constructed. However, the resultant automaton size remains prohibitively large for complex systems and specifications \citep{liu2013synthesis}.
While automata-based frameworks focus primarily on finding feasible solutions, researchers have extended them to achieve optimality~\citep{ren2024ltl,  wolff2012optimal, smith2011optimal, ulusoy2013optimality}.
These methods typically struggle with complex specifications or high-dimensional systems. 
To address these challenges, Kurtz and Lin~\citep{kurtz2023temporal} cast LTL motion planning as a shortest path problem in a GCS, enabling efficient solutions through convex optimization. 
Finally, these works focus on MTL and STL specifications, whereas generating  automata for MTL and STL remains challenging~\citep{brihaye2017mightyl, raman2015reactive, belta2019formal}. In contrast, our framework does not require building automata, but directly encodes the logic specifications into a unified optimization formulation, \jimrev{avoiding} the exponential growth in representation size of automata approaches.

\subsection{\xuan{Optimization-based Methods for Temporal Logic Planning}}

Optimization-based methods offer an alternative approach to handling temporal logic specifications. Rather than constructing automata, optimization-based techniques directly encode temporal logic constraints into mathematical programs~\citep{kurtz2022mixed, belta2019formal, raman2014model, wolff2014optimization}, and synthesize with underlying dynamic systems using a unified formulation to achieve both logical constraint satisfaction and trajectory optimization. \jimrev{Mixed-integer convex programming} (MICP) has emerged as a particularly effective framework for this approach. MICP formulations for temporal logic constraints are fundamentally NP-hard due to their combinatorial nature. However, researchers have developed several approaches to reduce computational complexity.

\subsubsection{Smoothing Approximation Methods}
Smoothing-based methods transform a discrete logical satisfaction problem into a continuous optimization problem, which can be solved by gradient-based techniques. Gilpin et al.~\citep{gilpin2020smooth} introduce a smooth robustness metric for STL that measures logical soundness on top of the gradient-based method.
Mehdipour et al.~\citep{mehdipour2019average} develop an average-based robustness for continuous-time STL specifications that provides a smoother quantitative semantics. Pant et al.~\citep{pant2018fly} propose a fly-by-logic approach for multi-drone fleets leveraging smooth approximations to enable real-time control with temporal logic objectives. While this method improves computational speed, it sacrifices the guarantees of completeness.

\subsubsection{Binary Variable Reduction Methods}
Many other studies also focus on improving MICP solver efficiency while maintaining completeness. One approach is to minimize the total number of binary variables. For example, Kurtz and Lin~\citep{kurtz2022mixed} propose disjunction encodings that use logarithmic numbers of binary variables, thereby improving scalability for long-horizon specifications. 

\subsubsection{Convex Relaxation Tightening Methods}
Another approach focuses on designing MICP formulations with tighter convex relaxations to improve B\&B efficiency. Marcucci and Tedrake~\citep{marcucci2019mixed} compare several formulation approaches for optimal control \xuanrev{under PWA dynamics}. Their later work~\citep{marcucci2024graphs, marcucci2023motion} introduces GCS, designing optimization formulations with tighter convex relaxations for motion planning around obstacles leveraging a structure similar to network flows. The benefits of tighter convex relaxations for motion planning under LTL are further explored by Kurtz and Lin~\citep{kurtz2021more,kurtz2023temporal}.

In this paper, we present LNF, a novel formulation that provides provably tighter convex relaxations for temporal logic specifications while maintaining both completeness and soundness guarantees. Our approach significantly improves computational efficiency for temporal logic planning problems compared to existing methods.

\subsection{\xuan{Applications of Temporal Logic Planning to Multi-Robot Coordination and Bipedal Locomotion}}

Recent years have witnessed increasing applications of temporal logic in robotic motion planning and control~\citep{kress2009temporal, plaku2016motion}. Among these diverse applications of temporal logic, two particularly interesting research directions have emerged: multi-agent coordination leveraging robot teaming, and integration with complex dynamic models for platforms such as legged robots.

\subsubsection{Multi-Robot Coordination} 
Multi-robot coordination under temporal logic specifications has been studied across different types of teaming. For homogeneous teams, researchers have developed specialized approaches for specific platforms: Pant et al.~\citep{pant2018fly} and Cardona et al.~\citep{cardona2023temporal} utilize temporal logic specifications for coordination of drone swarms, while Kress-Gazit et al.~\citep{kress2008courteous} and Scher et al.~\citep{scher2020warehouse} incorporate car-like behaviors and dynamics for navigation in urban environments or warehouses. Recent work has increasingly focused on heterogeneous teams that combine different robot types to leverage their complementary capabilities \citep{shamsah2025terrain, cao2022leveraging}. Schillinger et al.~\citep{schillinger2018simultaneous} introduce a framework that simultaneously handles task allocation and motion planning for heterogeneous multi-robot systems under temporal logic specifications. 
Zhou et al.~\citep{zhou2022reactive} introduce specific planning strategies for teams combining quadrupedal and wheeled robots.

\subsubsection{Motion Planning for Bipedal Locomotion} 
Bipedal locomotion planning has traditionally leveraged optimization-based approaches with predefined goals, such as~\citep{kuindersma2016optimization} and~\citep{huang2023efficient}. More recently, researchers have proposed integrating temporal logic with more complicated dynamics models for bipedal locomotion \citep{shamsah2023integrated, jiang2023abstraction}. Zhao et al.~\citep{zhao2022reactive} present a framework for reactive task and motion planning for whole-body dynamic locomotion. Gu et al.~\citep{gu2025robust} propose an STL-guided model predictive control approach for bipedal locomotion resilient to external perturbations. Ren et al.~\citep{ren2025accelerating} accelerate STL-based bipedal locomotion planning through Benders decomposition, separating logical constraints from nonlinear kinematics and dynamics. These works demonstrate that temporal logic can effectively describe complex locomotion tasks while respecting the dynamics constraints of robots. In our work, we further advance this direction by integrating LNF with the Linear Inverted Pendulum model, allowing bipedal walking robots to synthesize trajectories under temporal logic constraints on terrains with predefined safe contact regions.

\begin{table*}[t]
\centering
\caption*{\large\textbf{Nomenclature}}
\vspace{0.3cm}
\setlength{\fboxsep}{8pt}
\setlength{\fboxrule}{0.6pt}
\fbox{%
\begin{minipage}{0.975\textwidth}
\vspace{6pt}
\renewcommand{\arraystretch}{1.15}
\begin{minipage}[t]{0.49\textwidth}
\centering
\begin{tabular}{@{} l p{5.7cm} @{}}
\specialrule{1.2pt}{0pt}{2pt}
\multicolumn{2}{@{}l}{\textit{System Dynamics}} \\
\midrule
$k$ & Discrete time step, $k = 0, 1, \ldots, T$ \\
$T$ & Planning horizon \\
$\boldsymbol{x}_k$, $\boldsymbol{u}_k$ & System state and control input at time step $k$ \\
$\boldsymbol{\xi}$ & System trajectory (run) \\
$\mathcal{D}^i$ & $i$-th polytope region for PWA dynamics \\
$\boldsymbol{A}^i, \boldsymbol{B}^i$ & System matrices for mode $i$ \\
$\boldsymbol{H}^i_1, \boldsymbol{H}^i_2, \boldsymbol{h}^i$ & Polytope constraint matrices/vector for mode $i$ \\
\addlinespace
\specialrule{1.2pt}{2pt}{2pt}
\multicolumn{2}{@{}l}{\textit{Temporal Logic}} \\
\midrule
$\varphi$ & Temporal logic formula \\
$\pi$ & Atomic predicate \\
$\Pi$ & Set of atomic predicates \\
$z^{\pi^i_k}$ & Binary predicate variable for $\pi^i$ at time step $k$ \\
$\boldsymbol{z}^{\pi}$ & Vector of binary predicates across all predicates and time steps \\
$\wedge, \vee, \neg$ & Logical AND, OR, NOT \\
$\square_{[k_1,k_2]}$ & Operator ``Always" \\
$\lozenge_{[k_1,k_2]}$ & Operator ``Eventually" \\
$\mathcal{U}_{[k_1,k_2]}$ & Operator ``Until" \\
\bottomrule
\end{tabular}
\end{minipage}
\hfill
\begin{minipage}[t]{0.49\textwidth}
\centering
\begin{tabular}{@{} l p{5.7cm} @{}}
\specialrule{1.2pt}{0pt}{2pt}
\multicolumn{2}{@{}l}{\textit{Logic Tree (LT)}} \\
\midrule
$\mathcal{T}^{\varphi}$ & LT for formula $\varphi$ \\
$z^{\varphi_i}$ & Binary variable for internal node $\varphi_i$ \\
$\boldsymbol{z}^{\varphi}$ & Vector of internal node binary variables \\
\addlinespace
\specialrule{1.2pt}{2pt}{2pt}
\multicolumn{2}{@{}l}{\textit{Logic Network Flow (LNF)}} \\
\midrule
$\mathcal{F}^{\varphi}$ & LNF for formula $\varphi$ \\
$\mathcal{G}$ & Directed graph $(\mathcal{V}, \mathcal{E})$ \\
$v_s, v_t$ & Source and target vertices \\
$\mathcal{E}^{\text{in}}_v, \mathcal{E}^{\text{out}}_v$ & Incoming/outgoing edges of vertex $v$ \\
$\mathcal{P}_e$ & Predicates associated with edge $e$ \\
$y_e$ & Binary edge traversal variable \\
$\boldsymbol{\omega}_e$ & Flow variable for edge $e$ \\
$\boldsymbol{v}^+_e, \boldsymbol{v}^-_e$ & Artificially designed indicator vectors \\
\addlinespace
\specialrule{1.2pt}{2pt}{2pt}
\multicolumn{2}{@{}l}{\textit{Dynamic Network Flow (DNF)}} \\
\midrule
$\mathcal{G}_d$ & Temporal graph \\
$\mathcal{V}_d$ & Set of vertices in temporal graph representing spatial locations \\
$\mathcal{E}_d$ & Set of directed edges in temporal graph \\
$T_d, C_d$ & Travel time and edge capacity functions for each edge \\
$\theta_k$ & Time-varying edge traversal cost \\
$r_e$ & DNF flow variable \\
$\mathcal{P}^{\pi}$ & Vertices with physical locations satisfying predicate $\pi$ \\
\bottomrule
\end{tabular}
\end{minipage}

\vspace{8pt}
\begin{center}
\begin{tabular}{@{} l p{6.4cm} @{\hspace{1cm}} l p{6.4cm} @{}}
\specialrule{1.2pt}{0pt}{2pt}
\multicolumn{4}{@{}l}{\textit{Abbreviations}} \\
\midrule
B\&B  & Branch and Bound              & LT   & Logic Tree \\
DNF   & Dynamic Network Flow          & MICP & Mixed-Integer Convex Programming \\
F-M   & Fourier-Motzkin (Elimination) & MILP & Mixed-Integer Linear Programming \\
GCS   & Graph-of-Convex-Sets          & MIP  & Mixed-Integer Programming \\
LIP   & Linear Inverted Pendulum      & MTL  & Metric Temporal Logic \\
LNF   & Logic Network Flow           & PWA  & Piecewise-Affine \\
LP    & Linear Programming            & STL  & Signal Temporal Logic \\
\bottomrule
\end{tabular}
\end{center}
\vspace{6pt}
\end{minipage}%
}
\end{table*}

\section{Background}
\label{Sec:background}
\subsection{Piecewise-Affine Dynamic Systems}
Consider a discrete-time dynamical system in the form of
\begin{equation}
  \boldsymbol{x}_{k+1} = f(\boldsymbol{x}_k, \boldsymbol{u}_k)  
\label{eqn:dyn}
\end{equation}
where $\boldsymbol{x}_{k} \in \mathcal{X} \subseteq \mathbb{R}^{n_x} \times \mathbb{B}^{n_z}$ represents the state vector, consisting of continuous variables of size $n_x$ and binary variables of size $n_z$, with $\mathbb{B} = \{0,1\}$ and $k=0,1,\ldots,T$ denoting the time steps; $\boldsymbol{u}_k \in \mathcal{U} \subseteq \mathbb{R}^{n_u}$ represents the control input of size $n_u$. Given the control input at each time step and the initial state of the trajectory $\boldsymbol{x}_{0} \in \mathcal{X}_0$, where $\mathcal{X}_0$ is typically a singleton set containing only the initial condition, a run of the system is expressed as $\boldsymbol{\xi}=(\boldsymbol{x}_{0}\boldsymbol{u}_{0})(\boldsymbol{x}_{1}\boldsymbol{u}_{1})\cdots$ by rolling out Eqn.~\eqref{eqn:dyn}.

While Eqn.~\eqref{eqn:dyn} captures general nonlinear dynamics, a wide range of nonlinear dynamic systems can be effectively approximated \xuanrev{using Piecewise Affine (PWA) dynamics \citep{sontag1981nonlinear}}. \xuanrev{PWA dynamics} naturally translate into mixed-integer convex programs (MICPs) \citep{marcucci2019mixed}, making them ideal dynamics representations for our paper. Let $\boldsymbol{x} = \{\boldsymbol{x}_0, \ldots, \boldsymbol{x}_T\}$ and $\boldsymbol{u} = \{\boldsymbol{u}_0, \ldots, \boldsymbol{u}_{T-1}\}$ represent the state and control trajectories. We first define a collection of polytopes (\textit{i.e.}, bounded polyhedra): 
$\mathcal{D}^i \triangleq \{(\boldsymbol{x}_k, \boldsymbol{u}_k) \mid \boldsymbol{H}^i_1 \boldsymbol{x}_k + \boldsymbol{H}^i_2 \boldsymbol{u}_k \leq \boldsymbol{h}^i \}$, where $i \in \mathcal{I}$ denotes the index set. \xuanrev{The discrete-time PWA dynamics is expressed as:}
\begin{subequations}
\begin{equation}
\begin{aligned}
    \boldsymbol{x}_{k+1} &= \boldsymbol{A}^i \boldsymbol{x}_k + \boldsymbol{B}^i \boldsymbol{u}_k \\
    \boldsymbol{H}^i_1 \boldsymbol{x}_k &+ \boldsymbol{H}^i_2 \boldsymbol{u}_k \leq \boldsymbol{h}^i \ \ \text{for some } i \in \mathcal{I} 
\end{aligned}
\label{eqn:pwa}
\end{equation}
\end{subequations}
\noindent where $\boldsymbol{A}^i \in \mathbb{R}^{n_{x} \times n_{x}}$ and $\boldsymbol{B}^i \in \mathbb{R}^{n_{x} \times n_{u}}$ are the system matrices, and $\boldsymbol{H}^i_1 \in \mathbb{R}^{n_c \times n_{x}}$, $\boldsymbol{H}^i_2 \in \mathbb{R}^{n_c \times n_u}$, and $\boldsymbol{h}^i \in \mathbb{R}^{n_c}$ are the constraint matrices and vector, respectively.


\subsection{Temporal Logic Preliminaries}
In this paper, we focus on bounded-time temporal logic formulas built upon convex predicates, specifically Metric Temporal Logic (MTL) and Signal Temporal Logic (STL), where the maximum trajectory length $T$ to determine logic satisfiability is finite. We recursively define the syntax of temporal logic formulas as follows \citep{belta2019formal}:
$
\varphi \coloneqq \:
\pi \;|\; \neg\varphi \;|\; \varphi_1 \wedge \varphi_2 \;|\; \varphi_1 \vee \varphi_2 \;|\; \Diamond_{[k_1,k_2]}\;\varphi \;|\; \square_{[k_1,k_2]}\;\varphi \;|\; 
\varphi_1 \; \mathcal{U}_{[k_1,k_2]}\; \varphi_2
$, 
where the semantics consists of not only boolean operations “and” ($\wedge$) and “or” ($\vee$), but also temporal operators ``always" ($\square$), ``eventually" ($\lozenge$), and ``until" ($\mathcal U$). $\varphi$, $\varphi_1$ and $\varphi_2$ in the definition are formulas, and $\pi$ is an atomic predicate $\mathcal{X} \rightarrow \mathbb{B}$ whose truth value is defined by the sign of the convex function $g^\pi : \mathcal{X} \rightarrow \mathbb{R}$. We assume the convex function is a combination of linear functions, which can be expressed as $g^\pi(\boldsymbol{x}_k) = ({\boldsymbol{a}^{\pi}})^\top \boldsymbol{x}_k + b^\pi$. Let $\Pi = \{\pi^1, \ldots, \pi^{|\Pi|}\}$ be the set of $|\Pi|$ atomic predicates. Define the binary predicate variable $z^{\pi^i_k} \in \mathbb{B}$ for each predicate $\pi^i$ at time step $k$ such that:
\begin{gather}
\begin{aligned}
({\boldsymbol{a}^{\pi^i}})^\top \boldsymbol{x}_k + b^{\pi^i} \geq 0 &\Leftrightarrow z^{\pi^i_k}=1, \\  
({\boldsymbol{a}^{\pi^i}})^\top \boldsymbol{x}_k + b^{\pi^i} < 0 &\Leftrightarrow z^{\pi^i_k}=0
\label{eqn:predicate}
\end{aligned}
\end{gather}
Let vector $\boldsymbol{z}^{\pi}$ collect these binary predicates across all predicates and time steps into a single vector:
\begin{equation}
    \boldsymbol{z}^{\pi} = [z^{\pi^1_{0}}, z^{\pi^2_{0}}, \ldots, z^{\pi^{|\Pi|}_{0}}, z^{\pi^1_{1}}, \ldots, z^{\pi^{|\Pi|}_{T}}]^\top
\label{eqn:def_z_pi}
\end{equation}

A run $\boldsymbol{\xi}$ that satisfies a temporal logic formula $\varphi$ is denoted as $\boldsymbol{\xi} \models \varphi$. The satisfaction of a formula $\varphi$ having a state signal $\boldsymbol{x}$ starting from time step $k$ is defined inductively as in Table \ref{tab:STL_satisfy}.

\begin{table}[t!]
\centering
\caption {\label{tab:STL_satisfy} Validity semantics of Signal Temporal Logic}
\vspace{-0.1in}
\renewcommand{\arraystretch}{1.2}
\adjustbox{width=\columnwidth}{
\begin{tabular}{l c c} \\
\hline
$(\boldsymbol{x},k) \models \varphi_1 \wedge \varphi_2$ &$\Leftrightarrow$& $(\boldsymbol{x},k) \models \varphi_1 \wedge (\boldsymbol{x},k) \models \varphi_2$ \\
$(\boldsymbol{x},k) \models \varphi_1 \vee \varphi_2$ &$\Leftrightarrow$& $(\boldsymbol{x},k) \models \varphi_1 \vee (\boldsymbol{x},k) \models \varphi_2$ \\
$(\boldsymbol{x},k) \models \Diamond_{[k_1,k_2]}\varphi$ &$\Leftrightarrow$& $\exists {k^{'}\in[k+k_1,k+k_2]}, (\boldsymbol{x},k^{'}) \models \varphi$ \\
$(\boldsymbol{x},k) \models \square_{[k_1,k_2]}\varphi$ &$\Leftrightarrow$& $\forall {k^{'}\in[k+k_1,k+k_2]}, (\boldsymbol{x},k^{'}) \models \varphi$\\
$(\boldsymbol{x},k) \models {\varphi_1}\mathcal{U}_{[k_1,k_2]}{\varphi_2}$ &$\Leftrightarrow$& $\exists {k^{'}\in[k+k_1,k+k_2]}, (\boldsymbol{x},k^{'}) \models \varphi_2$ \\
&& $\wedge \ \forall {k^{''}\in[k+k_1,k^{'}]} (\boldsymbol{x},k^{''}) \models \varphi_1$ \\[0.5ex]
\hline
\end{tabular}}
\end{table}

In particular, STL provides a unique capability of admitting a \textit{robustness degree} \citep{fainekos2009robustness} of how strongly a formula is satisfied by a signal \citep{belta2019formal}. A positive robustness value indicates satisfaction, and the magnitude represents the margin of robustness against disturbances. \xuanrev{Table \ref{tab:robustness}} shows the semantics of the robustness degree of STL. 

\begin{table}[t!]
\centering
\caption{Robustness degree semantics of Signal Temporal Logic}
\label{tab:robustness}
\renewcommand{\arraystretch}{1.2}
\adjustbox{width=\columnwidth}{
\begin{tabular}{l c l} 
\hline
$\rho ^ \pi (\boldsymbol{x},k)$ & $=$ & $({\boldsymbol{a}^{\pi }})^\top \boldsymbol{x}_k + b^\pi$ \\
$\rho ^ {\neg\varphi} (\boldsymbol{x},k)$ & $=$ & $-\rho ^ \varphi  (\boldsymbol{x},k)$ \\
$\rho ^ {\varphi_1 \wedge \varphi_2} (\boldsymbol{x},k)$ & $=$ & ${\rm min} (\rho ^ {\varphi_1} (\boldsymbol{x},k),\rho ^ {\varphi_2} (\boldsymbol{x},k))$ \\
$\rho ^ {\varphi_1 \vee \varphi_2} (\boldsymbol{x},k)$ & $=$ & ${\rm max} (\rho ^ {\varphi_1} (\boldsymbol{x},k),\rho ^ {\varphi_2} (\boldsymbol{x},k))$ \\
$\rho ^ {\Diamond_{[k_1,k_2]}\varphi}(\boldsymbol{x},k)$ & $=$ & ${\max}_{k^{'}\in[k+k_1,k+k_2]} (\rho^{\varphi} (\boldsymbol{x},k^{'}))$ \\
$\rho ^ {\square_{[k_1,k_2]}\varphi}(\boldsymbol{x},k)$ & $=$ & ${\min}_{k^{'}\in[k+k_1,k+k_2]} (\rho^{\varphi} (\boldsymbol{x},k^{'}))$ \\
$\rho ^ {{\varphi_1}\mathcal{U}_{[k_1,k_2]}{\varphi_2}} (\boldsymbol{x},k)$ & $=$ & ${\max}_{k^{'}\in[k+k_1,k+k_2]}({\min}(\rho^{\varphi_2}(\boldsymbol{x},k^{'}),$ \\
& & $\quad{\min}_{k^{''}\in[k+k_1,k^{'}]}(\rho^{\varphi_1}(\boldsymbol{x},k^{''}))))$ \\[1ex]
\hline
\end{tabular}}
\end{table}

\subsection{Optimization-based Temporal Logic Motion Planning}

By combining PWA dynamics with temporal logic specifications, we define an integrated optimization problem for motion planning. We structure the objective function in the form of $f_{\text{obj}}(\boldsymbol{\xi}, \boldsymbol{z}^{\pi})=f_{\text{obj}}(\boldsymbol{x}, \boldsymbol{u}, \boldsymbol{z}^{\pi}) = \boldsymbol{x}^\top\boldsymbol{Q}\boldsymbol{x} + \boldsymbol{u}^\top\boldsymbol{R}\boldsymbol{u} + \boldsymbol{\Theta}^\top\boldsymbol{z}^{\pi}$, where the quadratic terms $\boldsymbol{x}^\top\boldsymbol{Q}\boldsymbol{x} + \boldsymbol{u}^\top\boldsymbol{R}\boldsymbol{u}$ aim to penalize high energy consumption and control effort, or maximize robustness measures. The linear term $\boldsymbol{\Theta}^\top\boldsymbol{z}^{\pi}$ incentivizes the satisfaction of specific predicates. For instance, it can be used to encode preferences for regions in the state space associated with lower risks of failure.



We present a formal statement of the optimization problem for motion planning and control under temporal logic constraints.



\begin{problem}[Temporal Logic Motion Planning with PWA Dynamics]
\label{prob1}
Given a system of the form \eqref{eqn:pwa} and a temporal logic specification $\varphi$, compute a control input sequence $\boldsymbol{u}$ such that the resulting trajectory $\boldsymbol{\xi}$ satisfies $\boldsymbol{\xi} \models \varphi$ and $f_{\text{obj}}(\boldsymbol{\xi}, \boldsymbol{z}^{\pi})$ is minimized.
\end{problem}

Given the problem statement above, we propose a unified formulation to solve it. The fundamental idea is to design two complementary sets of constraints: one instantiating the temporal logic specifications and the other capturing the \xuanrev{PWA dynamics}. The two constraint sets are then composed into a single formulation with the predicate variables $\boldsymbol{z}^{\pi}$ coupling both counterparts, such that the dynamics evolve in a manner consistent with the logical requirements. This integrated approach leads to the following optimization formulation:

\begin{formulation}
(Optimization Formulation for Temporal Logic Motion Planning with PWA Dynamics)
\begin{align*}
&\underset{\boldsymbol{\xi}, \boldsymbol{\gamma}, \boldsymbol{z}^{\pi}}{\text{minimize}} \quad f_{\text{obj}}(\boldsymbol{\xi}, \boldsymbol{z}^{\pi}) \nonumber \\
& \begin{aligned}
\text{s.t.} &\quad\;\; \text{Eqn. } \eqref{eqn:pwa}, \quad \boldsymbol{x}_0 \in \mathcal{X}_0 & \text{(PWA dynamics)} \\
& \quad\;\; \text{Eqn. } \eqref{eqn:predicate}, \quad \forall \: k, \forall \: \pi \in \Pi & \text{(Atomic predicates)} \\
& \quad\;\;
\mathcal{C}_{\text{logic}}(\boldsymbol{\gamma}, \boldsymbol{z}^{\pi}) \leq 0 & \text{(Temporal logics)}\\
\end{aligned}
\end{align*}
\label{eqn:general_formulation}
\end{formulation}

The temporal logic module is represented as a set of constraints denoted by $\mathcal{C}_{\text{logic}}(\boldsymbol{\gamma}, \boldsymbol{z}^{\pi}) \leq 0$. The vector $\boldsymbol{\gamma}$ comprises the variables specific to the logic formulation (\textit{e.g.}, edge variables in LNF or node variables in LT, as we will introduce later). 
In the following sections, we will explore different versions for the temporal logic formulations and PWA dynamics that fall under the umbrella of this general optimization structure.

In the next section, we study the LT approach, a conventional method to encode temporal logic constraints into mixed-integer programs (MIPs). This will provide the foundation for understanding our proposed LNF framework in Section IV.

\subsection{Logic Tree}
\jimrev{Logic Tree (LT) \citep{wolff2014optimization, raman2014model}, also known as the STL Tree \citep{kurtz2022mixed}, STL Parse Tree \citep{leung2023backpropagation}, or AND-OR Tree \citep{sun2022multi}, is a hierarchical structure that encodes a temporal logic formula's recursive syntax, decomposing it into nested logical and temporal operators to enable efficient optimization-based solving. We begin by formally defining the LT, followed by an example illustrating how a temporal logic formula is translated into this tree representation.}

\begin{definition}
\label{Def: Logic Tree}
A Logic Tree $T^\varphi$ constructed from a temporal logic specification $\varphi$ is defined as a tuple $(\circ, \Pi,  \mathcal{N}, \tau)$, where:
\begin{itemize}
\item $\circ \in \{\wedge, \vee\}$ denotes the combination type;
\item \jimrev{ $\Pi = \{\pi^1, \ldots, \pi^{|\Pi|}\}$ is the set of $|\Pi|$ predicates involved in the specification. Each leaf node corresponds to a predicate $\pi^i$ evaluated at time step $k$. A binary variable $z^{\pi_k^i}$ is associated with each leaf node to indicate whether the corresponding predicate is satisfied at that time step.}
\item $\mathcal{N}=\{T^{\varphi_0}, T^{\varphi_1}, \ldots, T^{\varphi_n}\}$ represents the set of $n+1$ internal nodes having at least one child, where the root node is denoted by $T^\varphi = T^{\varphi_0}$. Each node is associated with a temporal logic formula $\varphi_i$ and a combination type $\circ$. Similarly, each internal node is assigned a variable $z^{\varphi_i}$ to indicate the formula's validity.
\item $\tau = \{t^{\varphi_0} , t^{\varphi_1} , \ldots , t^{\varphi_n}\}$
is a list of starting times corresponding to each of the temporal logic formulas at the internal nodes.
\end{itemize}
\end{definition}

\begin{figure}[t!]
    \centering
    \includegraphics[width=0.45\textwidth]{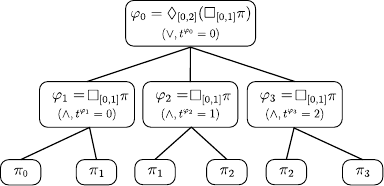}
     \caption{The LT for $\Diamond_{[0,2]}(\square_{[0,1]} \pi)$ given in Example \ref{exp:example1}.}
\label{fig:example}
\end{figure}

\begin{example}
\jimrev{Consider the specification $\Diamond_{[0,2]}(\square_{[0,1]} \pi)$ and let its corresponding LT be shown in Fig. \ref{fig:example}. 
The resulting LT contains a total of $10$ nodes, comprising $4$ internal nodes and $6$ leaf nodes. The root is a disjunction node, reflecting the semantics of the bounded ``eventually" operator $\Diamond_{[0,2]}$. Each of its three children is a conjunction node associated with the bounded ``always" operator $\square_{[0, 1]}$. The leaf nodes correspond to the evaluations of $\pi$ at the individual time steps.}
\label{exp:example1}
\end{example}

Given Definition~\ref{Def: Logic Tree}, now we reformulate Problem \ref{prob1} to an LT problem:

\begin{subproblem}[Temporal Logic Motion Planning with PWA Dynamics using Logic Tree]
Given a system of the form \eqref{eqn:pwa} and a temporal logic formula $\varphi$, construct an LT $T^\varphi$ according to Definition \ref{Def: Logic Tree}, and then compute a control input sequence $\boldsymbol{u}$ such that the resulting trajectory $\boldsymbol{\xi}$ satisfies $\boldsymbol{\xi} \models \varphi$ and $f_{\text{obj}}(\boldsymbol{\xi}, \boldsymbol{z}^{\pi})$ is minimized.
\label{prob1_1}
\end{subproblem}



To incorporate temporal logic specifications represented by an LT into an optimization framework, the approaches presented in \citep{wolff2014optimization} and \citep{raman2014model} formulate the problem as a \jimrev{mixed-integer convex program (MICP)}. In this formulation, binary variables are introduced for both the internal nodes, denoted by $z^{\varphi_i}, \forall i \in \{0, \ldots, n\}$, and the leaf nodes, denoted by $z^{\pi_k^i}, \forall t,i \in \{1, \ldots, |\Pi|\}$. For an internal node representing a conjunction, \textit{i.e.}, $\varphi = \wedge_{i=1}^p \varphi_i$, where each $\varphi_i$ corresponds to the suboperand of the child nodes, the MICP enforces the following set of constraints:
\begin{subequations}
\begin{align}
z^\varphi &\leq z^{\varphi_i}, \ \ i = 1, \ldots , p \label{eqn:tree1a} \\
z^\varphi &\geq 1-p+\sum_{i=1}^p z^{\varphi_i} \label{eqn:tree1b}
\end{align}
\label{eqn:tree1}%
\end{subequations}
%
Likewise, for an internal node corresponding to a disjunction, \textit{i.e.}, $\varphi = \vee_{i=1}^q \varphi_i$, the following constraints are imposed:
\begin{subequations}
\begin{align}
z^\varphi &\geq z^{\varphi_i}, \ \ i = 1, \ldots , q \label{eqn:tree2a} \\
z^\varphi &\leq \sum_{i=1}^q z^{\varphi_i} \label{eqn:tree2b}
\end{align}
\label{eqn:tree2}
\end{subequations}

On the root node, $z^{\varphi_0}=1$ must hold to ensure the full temporal logic specification is satisfied. By constructing the LT $T^\varphi$, we present the following formulation to solve Problem \ref{prob1_1}, which is identical to \xuanrev{Eqn.} 15 in \citep{belta2019formal}:
\begin{subformulation}
(Optimization Formulation for Temporal Logic Motion Planning with PWA Dynamics using Logic Tree)
\begin{align*}
&\underset{\boldsymbol{\xi}, \boldsymbol{z}^{\varphi}, \boldsymbol{z}^{\pi}} {\text{minimize}} \quad f_{\text{obj}}(\boldsymbol{\xi}, \boldsymbol{z}^{\pi}) \nonumber \\
& \begin{aligned}
\text{s.t.}& \quad\;\; \text{Eqn. } \eqref{eqn:pwa}, \ \ \boldsymbol{x}_0 \in \mathcal{X}_0 & \text{(PWA dynamics)} \\
& \quad\;\; \text{Eqn. } \eqref{eqn:predicate},\ \ \forall \: k, \forall \: \pi \in \Pi & \text{(Atomic predicates)} \\
& \left.\begin{aligned}
& \quad\; \text{Eqn. } \eqref{eqn:tree1} ~\eqref{eqn:tree2}, \ \ \forall \:T^{\varphi_i} \in \mathcal N \\
& \quad\; z^{\varphi_0}=1
\end{aligned}\right\} & \text{(LT constraints)}
\end{aligned} 
\end{align*}
\label{eqn:sub_formulation_1}
\end{subformulation}
\noindent where $\mathcal{C}_{\text{logic}}$ from Formulation \ref{eqn:general_formulation} is replaced by the specific LT constraints \eqref{eqn:tree1} and \eqref{eqn:tree2}, and the variables $\boldsymbol{\gamma}$ are replaced by $\boldsymbol{z}^{\varphi}$, which is given as:

\begin{equation}
    \boldsymbol{z}^{\varphi} = [z^{\varphi_0}, z^{\varphi_1}, \ldots, z^{\varphi_n}]^\top \in \mathbb{B}^{n+1}
\end{equation}

Constraints \eqref{eqn:tree1} and \eqref{eqn:tree2} guarantee that if all child variables $z^{\varphi_i}$ claim binary values, then the parent variable $z^{\varphi}$ will also take a binary value. By induction, this implies that as long as the leaf variables $\boldsymbol{z}^{\pi}$ are binary, all internal variables $\boldsymbol z^{\varphi}$ will be binary as well. Nevertheless, we explicitly impose binary constraints on all variables to ensure mathematical completeness of the formulation. \xuanrev{If $f_{\text{obj}}$ is linear in its variables, Formulation~\ref{eqn:sub_formulation_1} is a mixed-integer linear program (MILP); with a quadratic objective, it becomes a mixed-integer quadratic program (MIQP). Throughout this paper, we refer to its convex relaxation obtained by relaxing all binary variables to continuous variables in [0,1] as LT-relax. Similarly, the convex relaxation of Formulation~\ref{eqn:sub_formulation_2} (introduced in Section~\ref{Sec:LNF}) is referred to as LNF-relax.}

To illustrate this formulation, we apply it to Example \ref{exp:example1}:

\begin{subexample}[Example \ref{exp:example1} Continued]
We present Formulation \ref{eqn:sub_formulation_1} for Example \ref{exp:example1}. For the disjunction node at $\varphi_0$,~\eqref{eqn:tree2} translates to:
\begin{align*}
    z^{\varphi_0} = 1, \quad z^{\varphi_0} \geq z^{\varphi_i}, i=1,2,3, \quad z^{\varphi_0} \leq \sum_{i=1}^{3} z^{\varphi_i} 
\end{align*}
For the conjunction nodes at $\varphi_j$, $j \in \{1,2,3\}$,~\eqref{eqn:tree1} translates to:
\begin{align*}
    z^{\varphi_j} \leq z^{\pi_k}, \ \forall k \in I_j, \quad z^{\varphi_j} \geq -1+\sum_{k \in I_j} z^{\pi_k}
\end{align*}
where $I_1 = \{0,1\}$, $I_2 = \{1,2\}$, and $I_3 = \{2,3\}$. The constraints above substitute $\mathcal{C}_{\text{logic}}(\boldsymbol{\gamma}, \boldsymbol{z}^{\pi}) \leq 0$.
\label{example1_2}
\end{subexample}



\subsection{Branch and Bound}
\label{sec:BB}

\jimrev{Temporal logic motion planning is \xuanrev{NP-hard}, which is reflected in its MICP encoding \citep{karp2010reducibility}. To tackle this complexity, the Branch and Bound (B\&B) algorithm is a well-established approach for solving MILPs to optimality. By systematically exploring the solution space, B\&B guarantees convergence to the global optimum if a feasible solution exists, or definitively proves infeasibility otherwise. In this section, we briefly review B\&B, and refer readers to \citep{conforti2014integer} for a more detailed description. }

\jimrev{Let $P^*$ denote the optimal objective value of the original MICP. The B\&B algorithm constructs a search tree, where each node $P_i$ corresponds to a subproblem in which a subset of binary variables are fixed while the remaining binary variables are relaxed to their continuous domains. The root node, denoted by $P_0$, is obtained by relaxing all the binary variables. Solving the continuous convex relaxation associated with node $P_i$ yields a local lower bound $P_i^{\mathrm{lb}}$, on the optimal objective value of that subproblem. If the relaxed solution satisfies all integrality constraints, it is then taken as a feasible solution for the original MICP and can be used to update the global upper bound $P^{\mathrm{ub}}$, the objective value of the current incumbent solution. Otherwise, if the relaxed solution contains a fractional binary variable $z_j$, the algorithm branches on $z_j$ by creating two child nodes with the additional constraints $z_j = 0$ and $z_j = 1$, respectively.}

\jimrev{Before any feasible integer solution is found, $P^{\mathrm{ub}}$ is initialized to $+\infty$. The global lower bound is defined as the minimum lower bound among all active leaf nodes in the search tree,
$$
P^{\mathrm{lb}}=\min_{i\in\mathcal{L}} P_i^{\mathrm{lb}}
$$
where $\mathcal{L}$ denotes the set of active leaf nodes. During the search, any node satisfying $P_i^{\mathrm{lb}} \geq P^{\mathrm{ub}}$ is fathomed, since it cannot contain a better feasible solution than the incumbent. The algorithm terminates once $P^{\mathrm{lb}} = P^{\mathrm{ub}}$, 
at which point the incumbent is certified as the globally optimal solution, \textit{i.e.}, $P^{\mathrm{ub}} = P^*$ \citep{conforti2014integer}.}

\jimrev{To monitor the progress of the B\&B algorithm, we evaluate the relative optimality gap:
\begin{equation*}
    G_a = \frac{|P^{\mathrm{ub}} - P^{\mathrm{lb}}|}{|P^{\mathrm{ub}}|}
\end{equation*}
The solver typically terminates when bounds become completely tight ($G_a = 0$).
In addition, we evaluate the quality of the root relaxation using the root relaxation gap:
\begin{equation*}
    G_r = \frac{|P^{\mathrm{ub}} - P_0^{\mathrm{lb}}|}{|P^{\mathrm{ub}}|}
\end{equation*}
which measures the tightness of the root node continuous relaxation relative to the optimal integer objective.}


\section{Logic Network Flow}
\label{Sec:LNF}
This section presents the proposed Logic Network Flow (LNF) formulation. Section~\ref{sec:LNF_definition} provides a definition of LNF based on the LT, and Section~\ref{sect:formulation} develops the optimization formulation based on this representation, and Section~\ref{sect:variable_elimination} describes a network-flow-based Fourier–Motzkin variable elimination procedure that enhances computational efficiency of the optimization framework. \jimrev{Sections~\ref{Sec:proof_of_tighter_relaxation} and \ref{Sec:computational_complexity} theoretically ground this performance gain by proving a tighter convex relaxation and analyzing problem scale in terms of the quantity of variables and constraints.} 

\subsection{Definition}
\label{sec:LNF_definition}

Our approach begins with the definition of LNF, a new formulation for encoding temporal logic specifications.

\begin{definition}
\label{Def: Logic Network Flow}
A Logic Network Flow $\mathcal{F}^{\varphi}$ from the specification $\varphi$ is defined as a tuple $(\mathcal{G}, \mathcal{P}, \Pi)$, where:
\begin{itemize}
    \item $\mathcal{G} = (\mathcal{V}, \mathcal{E})$ is a directed graph with a vertex $v_{s} \in \mathcal{V}$ as the source vertex and a vertex $v_{t} \in \mathcal{V}$ as the target vertex.
    \item $\Pi = \{\pi^1, \ldots, \pi^{|\Pi|}\}$ is the set of $|\Pi|$ predicates involved in the specification (as in Definition \ref{Def: Logic Tree}). The binary vector $\boldsymbol{z}^{\pi}$ is same as the definition in \eqref{eqn:def_z_pi}, representing the truth values of all predicates in $\Pi$ at given times.
    \item $\mathcal{P}$ is a collection of sets of $n_e$ ($n_e \leq |\Pi|\cdot T$) possibly negated predicates that must hold true (or false if negated) to pass through each edge $e \in \mathcal{E}$, defined as $P_e \coloneqq \{\pi^i_k | \pi^i \in \Pi \text{ s.t. } z^{\pi^i_k}=1 \text{ (or } z^{\pi^i_k}=0 \text{ if } \pi^i \text{ is negated})\} \in \mathcal{P}$. 
\end{itemize}
\end{definition}

\jimrev{We provide a simple algorithm for translating an existing LT into an LNF in Appendix~\ref{appendix_LT_to_LNF}. Fig.~\ref{fig:example2} illustrates how conjunction and disjunction nodes in an LT are converted into their corresponding LNF structures using this algorithm.}
Given the definition above, we reformulate Problem \ref{prob1} using LNF.

\begin{subproblem}[Temporal Logic Motion Planning with PWA Dynamics using Logic Network Flow]
Given a system of form \eqref{eqn:pwa} and a temporal logic formula $\varphi$, construct a Logic Network Flow $\mathcal{F}^{\varphi}$ according to Definition \ref{Def: Logic Network Flow}, and then compute a control input sequence $\boldsymbol{u}$ such that the resulting trajectory $\boldsymbol{\xi}$ satisfies $\boldsymbol{\xi} \models \varphi$ and $f_{\text{obj}}(\boldsymbol{\xi}, \boldsymbol{z}^{\pi})$ is minimized.
\label{prob1_2}
\end{subproblem}

\begin{figure}[t!]
    \centering
    \includegraphics[width=0.47\textwidth]{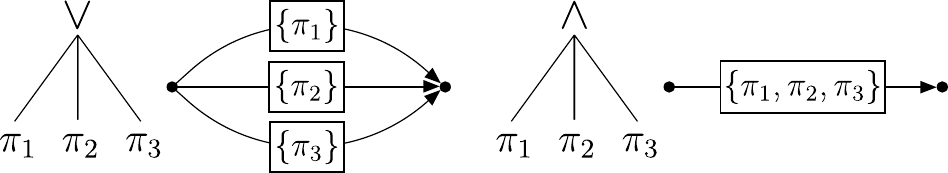}
     \caption{An illustration of the strategy to translate conjunction and disjunction combination types from an LT to an LNF in Algorithm \ref{Algorithm:build_node}.}
\label{fig:example2}
\end{figure} 

\subsection{Optimization Formulation}
\label{sect:formulation}

Given an LNF $\mathcal{F}^{\varphi}$, we propose an optimization formulation with a tighter convex relaxation than Formulation \ref{eqn:sub_formulation_1}. For each edge $e \in \mathcal{E}$ in the LNF, let $y_e \in \mathbb{B}$ specify if the edge is traversed by the flow, with a total of one unit of flow entering the network (\textit{i.e.}, $\sum_{e \in \mathcal{E}^{\rm{out}}_{v_s}} y_e = 1$). In addition, we associate a multi-dimensional continuous flow variable $\boldsymbol{\omega}_e \in [0, 1]^{|\Pi|}$ to each edge, with its in-flow at $v_s$ being $\boldsymbol{z}^{\pi}$. The idea of associating continuous variables with edges in network flow structures stems from the Graph-of-Convex-Sets (GCS) formulation \citep{marcucci2024graphs}, where continuous flow variables represent dynamics trajectories that must satisfy convex set constraints. Our LNF formulation adapts this concept to the logical domain, where $\boldsymbol{\omega}_e$ indicates predicate truth values.

We devise the convex set constraint to instantiate the logical requirements, such that if the in-flow passes through an edge $e$, we require $\boldsymbol{\omega}_e[i] = 1$ for all non-negated predicates $\pi_i \in P_e$, where $\pi_i$ is the $i^{\text{th}}$ predicate in $\Pi$; for negated predicates $\neg \pi_i \in P_e$, we require $\boldsymbol{\omega}_e[i] = 0$; and for predicates not contained in $P_e$, no constraint is imposed on $\boldsymbol{\omega}_e[i]$. These requirements can be expressed as two convex set constraints that need to be enforced simultaneously:
\begin{equation}
    \boldsymbol{\omega}_e \geq y_e\boldsymbol{v}^+_e, \quad \boldsymbol{\omega}_e \leq \boldsymbol{1}_{|\Pi|}-y_e\boldsymbol{v}^-_e
\label{eqn:edge}
\end{equation}
where $\boldsymbol{v}^+_e \in \mathbb{B}^{|\Pi|}$ and $\boldsymbol{v}^-_e \in \mathbb{B}^{|\Pi|}$ are artificially designed column vectors to help encode the logical constraints. Specifically, $\boldsymbol{v}^+_e[i] = 1$ if $\pi_i \in P_e$ , and 0 otherwise; and $\boldsymbol{v}^-_e[i] = 1$ if $\neg\pi_i \in P_e$, and 0 otherwise.

For logical consistency, we also demand that if all predicates in the set $P_e$ are satisfied, then the edge $e$ must be traversed ($y_e=1$). This is identical to constraint \eqref{eqn:tree1b}:
\begin{equation}
y_e \geq 1 - |P_e| + \sum_{\pi_i \in P_e} z^{\pi_i} + \sum_{\neg\pi_i \in P_e} (1-z^{\pi_i})
\label{eqn:edge_completeness}
\end{equation}

For each vertex $v \in \mathcal{V}$ with the input edges $\mathcal{E}_{v}^{\rm in} \subset \mathcal{E}$ and the output edges $\mathcal{E}_{v}^{\rm out} \subset \mathcal{E}$, flow conservation constraints are enforced on both $y_e$ and $\boldsymbol{\omega}_{e}$:
%
\begin{subequations}
\label{eqn:vertex}
\begin{align}
\sum_{e \in \mathcal{E}^{\rm in}_{v}} y_{e} &= \sum_{e \in \mathcal{E}^{\rm out}_{v}} y_{e} \label{eqn:vertex-a} \\
\sum_{e \in \mathcal{E}^{\rm in}_{v}} \boldsymbol{\omega}_{e} &= \sum_{e \in \mathcal{E}^{\rm out}_{v}} \boldsymbol{\omega}_{e} \label{eqn:vertex-b}
\end{align}
\end{subequations}
Flow conservation constraints are also applied to the source vertex $v_s$, where one unit of flow is injected:
%
%
\begin{subequations}
\label{eqn:input}
\begin{align}
1 &= \sum_{e \in \mathcal{E}^{\rm out}_{v_s}} y_{e} \label{eqn:input-a} \\
\boldsymbol{z}^{\pi} &= \sum_{e \in \mathcal{E}^{\rm out}_{v_s}} \boldsymbol{\omega}_{e} \label{eqn:input-b}
\end{align}
\end{subequations}

Overall, the LNF formulation aims to direct a continuous in-flow $\boldsymbol{z}^{\pi}$ from the source vertex to the target vertex. If $\boldsymbol{z}^{\pi}$ reaches the target vertex, then we conclude that the specification $\varphi$ is satisfied. 

\begin{figure}[t!]
    \centering
    \includegraphics[width=0.45\textwidth]{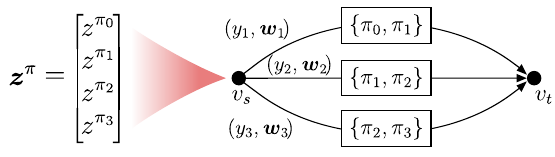}
     \caption{The LNF for $\Diamond_{[0,2]}(\square_{[0,1]} \pi)$, given in Example \ref{example2_1}.}
\label{fig:example2_1}
\end{figure} 

\begin{example}
Consider the same logic specification as in Example \ref{exp:example1}, whose corresponding LNF is illustrated by Fig. \ref{fig:example2_1}. This LNF consists of $3$ edges and $2$ vertices: $v_s$ and $v_t$. The variables in the LNF are $[y_1, y_2, y_3]$, $y_i \in \mathbb{B}$, and  $[\boldsymbol{\omega}_1, \boldsymbol{\omega}_2, \boldsymbol{\omega}_3]$, $\boldsymbol{\omega}_i \in \xuanrev{[0,1]}^4$. The top edge is associated with the predicate set $P_1=\{\pi_0,\pi_1\}$ containing only non-negated predicates, so the first inequality constraint in \eqref{eqn:edge} simplifies to $\boldsymbol{\omega}_1 \geq y_1[1, 1, 0, 0]^\top$.
Similar constraints are applied to the other two edges. For the source vertex $v_s$, the flow conservation constraints are given by $y_1+y_2+y_3=1$ and $\boldsymbol{z}^{\pi}=\boldsymbol{\omega}_1+\boldsymbol{\omega}_2+\boldsymbol{\omega}_3$. 
\label{example2_1}
\end{example}

Given a constructed LNF $\mathcal{F}^{\varphi}$, we reformulate Problem \ref{prob1} with the LNF framework:
\begin{subformulation}
(Optimization Formulation for Temporal Logic Motion Planning with PWA Dynamics using Logic Network Flow)
\begin{align*}
&\underset{\boldsymbol{\xi}, \boldsymbol{y}, \boldsymbol{\omega}, \boldsymbol{z}^{\pi}} {\text{minimize}} \quad f_{\text{obj}}(\boldsymbol{\xi}, \boldsymbol{z}^{\pi}) \nonumber \\
& \begin{aligned}
\text{s.t.}& \quad\;\; \text{Eqn. } \eqref{eqn:pwa}, \ \ \boldsymbol{x}_0 \in \mathcal{X}_0 & \text{(PWA dynamics)} \\
& \quad\;\; \text{Eqn. } \eqref{eqn:predicate},\ \ \forall \: k, \forall \: \pi \in \Pi & \text{(Atomic predicates)} \\
& \left.\begin{aligned}
& \quad\; \text{Eqn. } \eqref{eqn:edge},\ \ \forall e \in \mathcal{E} \\
& \quad\; \text{Eqn. } \eqref{eqn:vertex}, \eqref{eqn:input} \ \ \forall v \in \mathcal{V}
\end{aligned}\right\} & \text{(LNF constraints)}
\end{aligned} 
\end{align*}
\label{eqn:sub_formulation_2}
\end{subformulation}
\noindent Specifically, $\mathcal{C}_{\text{logic}}$ from Formulation \ref{eqn:general_formulation} is replaced by LNF constraints \eqref{eqn:edge}, \eqref{eqn:vertex}, and \eqref{eqn:input}, and variables $\boldsymbol{\gamma}$ are replaced by $\boldsymbol{y}$ and $\boldsymbol{\omega}$, which are defined as:
\begin{subequations}
\begin{align}
    \boldsymbol{y} &= [y_1, y_2, \ldots, y_{|\mathcal{E}|}]^\top \in \mathbb{B}^{|\mathcal{E}|} \\
    \boldsymbol{\omega} &= [\boldsymbol{\omega}_1^\top, \boldsymbol{\omega}_2^\top, \ldots, \boldsymbol{\omega}_{|\mathcal{E}|}^\top]^\top \in [0,1]^{|\mathcal{E}|\cdot|\Pi|}
\end{align}
\end{subequations}

\subsection{Network-flow-based Fourier-Motzkin Elimination}
\label{sect:variable_elimination}

While GCS admits tighter convex relaxations, it comes at the cost of introducing additional continuous flow variables, which in turn may increase the problem size and computational overhead. In \citep{marcucci2023motion}, the authors suggest taking advantage of parallel computing to improve the solving speed for the convex relaxations of GCS.

With GCS as the backbone of our established LNF formulation, we further improve its computational efficiency by introducing a novel network-flow-based Fourier–Motzkin procedure that reduces the number of variables in the formulation while preserving the tightness of the relaxation. This procedure eliminates the flow variables $\boldsymbol{\omega}_e$ and replaces the flow conservation constraints with inequality constraints. Although this technique could simplify general GCS formulations, in this paper we focus on how it refines our LNF formulation and improves computational performance without resorting to parallel computing.

We present here a simplified Fourier-Motzkin elimination process featuring the major changes from Formulation~\ref{eqn:sub_formulation_2}. For a rigorous mathematical treatment, we refer readers to Appendix \ref{appendix_project_one_by_one}.

Since constraints \eqref{eqn:edge} already provide explicit bounds on all flow variables $\boldsymbol{\omega}_e$, they can be used to obviate the flow conservation constraints \eqref{eqn:vertex-b} and \eqref{eqn:input-b}. Starting with constraint \eqref{eqn:input-b} for the source vertex, each $\boldsymbol{\omega}_e$ on the right-hand side is substituted by its corresponding upper and lower bounds from \eqref{eqn:edge}. This substitution removes $\boldsymbol{\omega}_e$ entirely and creates two sets of inequalities involving only variables $\boldsymbol{z}^{\pi}$ and $y_e$:
\begin{equation}
\boldsymbol{z}^{\pi} \geq \sum_{e \in \mathcal{E}^{\rm out}_{v_s}} y_e \boldsymbol{v}^+_e, \quad \boldsymbol{z}^{\pi} \leq \sum_{e \in \mathcal{E}^{\rm out}_{v_s}} \boldsymbol{1}_{|\Pi|} - y_e \boldsymbol{v}^-_e
\label{eqn:new_ineq_flow_conserv}
\end{equation}
Constraints \eqref{eqn:new_ineq_flow_conserv} are added to the formulation in place of \eqref{eqn:input-b}. Note that \eqref{eqn:new_ineq_flow_conserv} is equivalent to \eqref{eqn:edge} and \eqref{eqn:input-b} with respect to the feasible set for $\boldsymbol{z}^{\pi}$, and therefore this variable reduction preserves the tightness of the convex relaxation. See Appendix \ref{appendix_epigraph} for a detailed proof.

We next propagate the flow out of $v_s$ and to the subsequent vertices. Consider a vertex $v$ whose full set of input flow variables $\boldsymbol{\omega}_e$, where $e \in \mathcal{E}^{\rm in}_{v}$, are within the right-hand side of \eqref{eqn:input-b}. These variables can be replaced by the corresponding output flows from $v$ using the flow conservation constraint \eqref{eqn:vertex-b}:
\begin{equation*}
    \boldsymbol{z}^{\pi} = \sum_{e \in \mathcal{E}^{\rm out}_{v_s} \setminus \mathcal{E}^{\rm in}_{v}} \boldsymbol{\omega}_{e} + \sum_{e \in \mathcal{E}^{\rm out}_{v}} \boldsymbol{\omega}_{e}
\end{equation*}
This transformation yields new constraints with a similar structure to \eqref{eqn:input-b}, but with different $\boldsymbol{\omega}_e$ variables on the right-hand side. Following the substitution process described in the previous paragraph, we eliminate $\boldsymbol{\omega}_e$ by imposing its bounds from \eqref{eqn:edge}, which results in constraints involving only $\boldsymbol{z}^{\pi}$ and $y_e$. These derived constraints also become part of the formulation.

In addition, inequality constraints are generated between input and output edge variable $y_e$ by combining \eqref{eqn:edge} with flow conservation \eqref{eqn:vertex-b}.
When there are non-negated predicates on the input edges and negated predicates on the output edges, or vice versa, these inequality constraints become:
\begin{subequations}
\begin{align}
   \sum_{e \in \mathcal{E}^{\rm in}_{v}} y_e \boldsymbol{v}^+_e &\leq \sum_{e \in \mathcal{E}^{\rm out}_{v}} \boldsymbol{1}_{|\Pi|} - y_e \boldsymbol{v}^-_e \\
   \sum_{e \in \mathcal{E}^{\rm in}_{v}} \boldsymbol{1}_{|\Pi|} - y_e \boldsymbol{v}^-_e &\geq \sum_{e \in \mathcal{E}^{\rm out}_{v}} y_e \boldsymbol{v}^+_e
\end{align}
\label{eqn:ineq_contradict}%
\end{subequations}
\xuanrev{These inequalities turn out to be redundant, as they can be obtained by chaining the lower and upper bounds of \eqref{eqn:new_ineq_flow_conserv} at $v_s$ with those analogous to \eqref{eqn:new_ineq_flow_conserv} at $v$, both of which bound $\boldsymbol{z}^{\pi}$. They are therefore excluded from the formulation.}

\begin{figure*}[t!]
    \centering
    \includegraphics[width=0.99\textwidth]{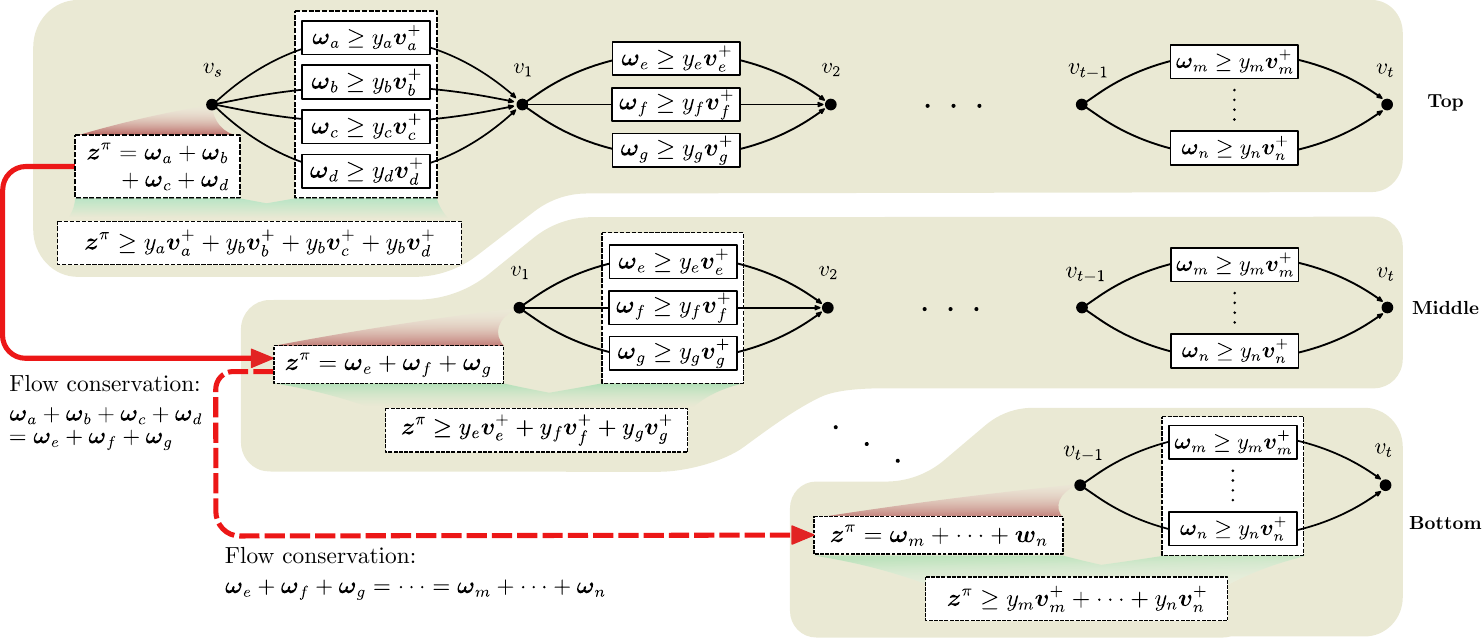}
     \caption{Illustration of the Network-Flow-Based Fourier-Motzkin variable elimination process. \textit{Top:} Starting from source vertex $v_s$, constraints \eqref{eqn:edge} ($\boldsymbol{\omega}_i \geq y_i \boldsymbol{v}^+_i$, $i=a,b,c,d$) and \eqref{eqn:input-b} ($\boldsymbol{z}^{\pi}=\sum_{i=a}^d\boldsymbol{\omega}_i$) are replaced by a single inequality \eqref{eqn:new_ineq_flow_conserv} ($\boldsymbol{z}^{\pi} \geq \sum_{i=a}^d y_i \boldsymbol{v}^+_i$), thereby eliminating flow variables $\boldsymbol{\omega}_a$ to $\boldsymbol{\omega}_d$. \textit{Middle:} Flow conservation at vertex $v_1$ ($\sum_{i=a}^d\boldsymbol{\omega}_i=\sum_{j=e}^g\boldsymbol{\omega}_j$) is applied, generating a new constraint with the same structure as \eqref{eqn:input-b} ($\boldsymbol{z}^{\pi}=\sum_{j=e}^g\boldsymbol{\omega}_j$). When combined with convex set constraints, this yields $\boldsymbol{z}^{\pi} \geq \sum_{j=e}^g y_j \boldsymbol{v}^+_j$. \textit{Bottom:} This process continues sequentially, propagating flows through the network until reaching target vertex $v_t$, at which point no further propagation is possible.}
\label{fig:F-M}
\end{figure*}

The flow conservation constraint \eqref{eqn:vertex-b} is  omitted after adding these derived constraints above. We continue this process sequentially, propagating flows through each vertex in the network until reaching $v_t$, where no further propagation is possible. Fig. \ref{fig:F-M} illustrates this process on an LNF with a simple structure.

At this point, the flow variables $\boldsymbol{\omega}_e$ and their associated constraints in \eqref{eqn:edge}, \eqref{eqn:vertex-b}, and \eqref{eqn:input-b} have been completely transformed to inequality constraints between $\boldsymbol{z}^{\pi}$ and $y_e$. This transformation leads to a formulation with significantly fewer continuous variables while retaining the same tightness of convex relaxation as the original formulation presented in Section \ref{sect:formulation}.
\begin{subexample}[Example \ref{example2_1} Continued] \label{exp:fm_continued}
We build on Example \ref{example2_1} to demonstrate the variable elimination procedure. The graph consists of two vertices: $v_s$ and $v_t$. Applying the inequality constraints \eqref{eqn:edge} to the flow conservation constraint at $v_s$ gives $\boldsymbol{z}^{\pi}=\boldsymbol{\omega}_1+\boldsymbol{\omega}_2+\boldsymbol{\omega}_3 \geq y_1[1, 1, 0, 0]^\top + y_2[0, 1, 1, 0]^\top + y_3[0, 0, 1, 1]^\top$. This produces the following set of constraints:
\begin{align*}
    y_1+y_2+y_3 &=1 \\
    y_1 &\leq \boldsymbol{z}^{\pi}[1], \\
    y_1+y_2 &\leq \boldsymbol{z}^{\pi}[2], \\ 
    y_2+y_3 &\leq \boldsymbol{z}^{\pi}[3], \\
    y_3 &\leq \boldsymbol{z}^{\pi}[4]
\end{align*}
For comparison, recall from Example \ref{example1_2} that the constraints derived from the three conjunction nodes in the LT formulation are:
\begin{alignat*}{2}
    z^{\varphi_1} &\leq \boldsymbol{z}^{\pi}[1], \quad  &z^{\varphi_1} &\leq \boldsymbol{z}^{\pi}[2], \\ 
    z^{\varphi_2} &\leq \boldsymbol{z}^{\pi}[2], \quad  &z^{\varphi_2} &\leq \boldsymbol{z}^{\pi}[3], \\
    z^{\varphi_3} &\leq \boldsymbol{z}^{\pi}[3], \quad  &z^{\varphi_3} &\leq \boldsymbol{z}^{\pi}[4]
\end{alignat*}
where the above two formulations share the same $y_i$ and $z^{\varphi_i}$. By examining these constraints from the LT with the last four constraints in the LNF, it is easy to conclude that the LNF formulation yields a tighter convex relaxation.
\end{subexample}
\begin{remark}
While this paper focuses on formulating temporal logic specifications, the proposed Fourier-Motzkin elimination process can be applied to simplify many other types of GCS formulations by removing flow variables and improving computational performance without sacrificing relaxation tightness. In Appendix \ref{app:gcs_application}, we provide an example by applying it to the minimum-time control of discrete-time linear systems from Chapter 10.1 of \citep{marcucci2024graphs}. \xuanrev{Its purpose is to underscore the generality of this network-flow-based Fourier-Motzkin pipeline as a systematic procedure for discovering strong valid inequalities. The coupled cuts obtained in Example~\ref{exp:fm_continued} (e.g., $y_1+y_2 \leq \boldsymbol{z}^{\pi}[2]$, which we generalize to inequalities \eqref{eqn:lnf_2} and \eqref{eqn:lnf_3} in the next section) could otherwise be hand-crafted as a problem-specific trick --- as we do for LT in Section~\ref{Sec:ablation_source} --- yet they arise automatically from the network flow formulation rather than through human investigation. In Appendix~\ref{app:gcs_application}, the same pipeline automatically uncovers tightening inequalities such as \eqref{eq:simplified_flow} for the minimum-time control problem.}
\end{remark}

\subsection{Proof of Tighter Convex Relaxation}
\label{Sec:proof_of_tighter_relaxation}

Following the empirical example from the previous section, here we present a theorem rigorously proving that the convex relaxations from the LNF formulations are tighter than those from the LT formulations. Specifically, we show that for any problem setup including cost coefficients $\boldsymbol{Q}$, $\boldsymbol{R}$, $\boldsymbol{\Theta}$, and initial conditions $\boldsymbol{x_0}$, the optimal cost of the convex relaxation of LNF is greater than or equal to that of LT. We denote the convex relaxation problems of LNF and LT as LNF-relax and LT-relax, respectively, with their corresponding optimal objective values as $f_{\text{LNF-r}}^{*}$ and $f_{\text{LT-r}}^{*}$. The following theorem formalizes this statement: 
\begin{theorem}
\xuanrev{Let the objective depend only on $\boldsymbol{\xi}$ and $\boldsymbol{z}^\pi$, and not on the logic-formulation variables ($\boldsymbol{y}$ in LNF, $\boldsymbol{z}^\varphi$ in LT). Throughout this paper we use $f_{obj}(\boldsymbol{\xi}, \boldsymbol{z}^\pi) = \boldsymbol{x}^\top \boldsymbol{Q} \boldsymbol{x} + \boldsymbol{u}^\top \boldsymbol{R} \boldsymbol{u} + \boldsymbol{\Theta}^\top \boldsymbol{z}^\pi$ with $\boldsymbol{Q} \succeq 0$, $\boldsymbol{R} \succeq 0$. Then for} any given $\boldsymbol{Q}$, $\boldsymbol{R}$, $\boldsymbol{\Theta}$, and $\boldsymbol{x_0}$, if the convex relaxation problem LNF-relax is feasible with an optimal cost $f_{\text{LNF-r}}^{*}$, then the convex relaxation problem LT-relax is also feasible with an optimal cost $f_{\text{LT-r}}^{*} \leq f_{\text{LNF-r}}^{*}$.
\label{theorem1}
\end{theorem}
\begin{proof}
Without loss of generality, we assume the logic specification takes the form of $\varphi = \bigwedge_{l=1}^{L}\bigvee_{m=1}^{M_l}\bigwedge_{n=1}^{N_{m,l}}\pi_{m,n}^l $, where $\pi_{m,n}^l$ is a predicate belonging to $\Pi$ (possibly negated). This conjunctive-disjunctive form naturally arises from common temporal logic specifications where multiple requirements must be satisfied simultaneously (outer conjunction), each potentially having alternative ways of satisfaction (disjunction), with specific conditions for each alternative (inner conjunction). Any propositional logic formula can be converted to this form through iterative application of distributivity and De Morgan's laws \citep{jackson2004clause}.
\begin{figure*}[t!]
    \centering
    \includegraphics[width=0.95\textwidth]{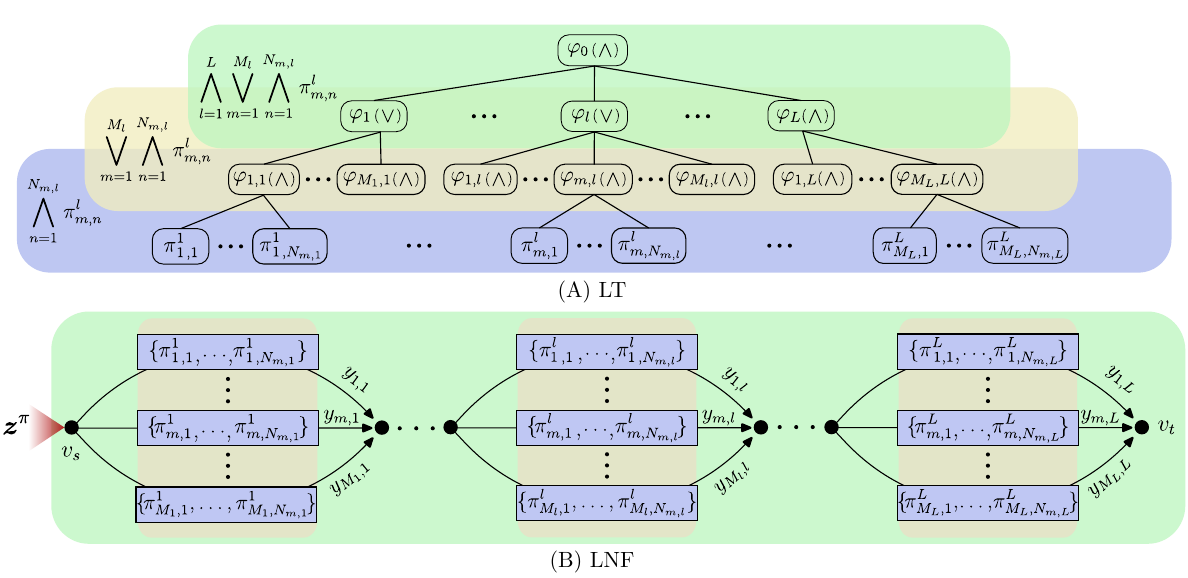}
     \caption{The representations of the conjunctive-disjunctive form $\varphi =\bigwedge_{l=1}^{L}\bigvee_{m=1}^{M_l}\bigwedge_{n=1}^{N_{m,l}}\pi_{m,n}^l$. (A) LT representation; (B) LNF representation constructed by Algorithm~\ref{Algorithm:build_node}.}
\label{fig:disjunctive_NF}
\end{figure*} 
This is reflected in the LT  shown in Fig. \ref{fig:disjunctive_NF} (A): 
the root node $\varphi_0$ is a conjunction of suboperands $\varphi_l$ $(l = 1, \dots, L)$; each $\varphi_l$ is a disjunction of formula $\varphi_{m,l}$ $(m = 1, \dots, M_l)$, with each formula itself a conjunction of predicates $\pi_{m,n}^l$ $(n = 1, \dots, N_{m,l})$. 
Building on this conjunctive-disjunctive structure, we formulate the corresponding convex relaxation problem LT-relax as:
\allowdisplaybreaks
\begin{subequations}
\begin{alignat}{2}
&\underset{\boldsymbol{\xi}, \boldsymbol{z}^{\varphi}, \boldsymbol{z}^{\pi}}{\textit{minimize}} \hspace{-1em} && \hspace{1em} \quad \boldsymbol{x}^\top\boldsymbol{Q}\boldsymbol{x} + \boldsymbol{u}^\top\boldsymbol{R}\boldsymbol{u} + \boldsymbol{\Theta}^\top\boldsymbol{z}^{\pi} \nonumber \\
&\textit{s.t.} \quad &&\text{for } l = 1,\ldots, L: \nonumber \\
& &&\quad z^{\varphi_l} = 1 \label{eqn:lt_1} \\
& &&\quad z^{\varphi_l} \geq z^{\varphi_{m,l}}, \quad m = 1,\ldots,M_l \label{eqn:lt_2} \\ 
& &&\quad z^{\varphi_l} \leq \sum_{m=1}^{M_l} z^{\varphi_{m,l}} \label{eqn:lt_3} \\ 
& &&\text{for } m = 1,\ldots, M_l: \nonumber \\
& &&\quad z^{\varphi_{m,l}} \geq 1 - |P_{m,l}| + \sum_{\pi_j \in P_{m,l}} z^{\pi_j} \nonumber \\
& &&\qquad \qquad + \sum_{\neg\pi_j \in P_{m,l}} (1-z^{\pi_j}) \label{eqn:lt_conj} \\
& &&\quad \text{for } j = 1,\ldots,|\Pi|: \nonumber \\
& &&\quad \quad z^{\pi_j} \geq z^{\varphi_{m,l}}, \quad \text{if } \pi_j \in P_{m,l} \label{eqn:lt_4} \\ 
& &&\quad \quad z^{\pi_j} \leq (1-z^{\varphi_{m,l}}), \quad \text{if } \neg\pi_j \in P_{m,l} \label{eqn:lt_5} \\ 
& &&\eqref{eqn:pwa}, \ \boldsymbol{x}_0 \in \mathcal{X}_0, \quad \eqref{eqn:predicate}, \ \forall \: k, \forall \: \pi \in \Pi \label{eqn:lt_6}
\end{alignat}
\end{subequations}
where $P_{m,l}$ denotes the set of possibly negated atomic predicates $\{\pi^l_{m,1}, \ldots, \pi^l_{m,N_{m,l}}\}$ at the leaf nodes of the suboperand rooted at $\varphi_{m,l}$.

With Algorithm \ref{Algorithm:build_node}, we construct the LNF for $\varphi$ as shown in Fig. \ref{fig:disjunctive_NF} (B). In particular, Algorithm \ref{Algorithm:build_node} collects all predicates at leaf nodes under each $\varphi_{m,l}$ to constitute a predicate set $P_{m,l}$ for each edge, coinciding with our definition of $P_{m,l}$ in the LT formulation. By applying Fourier-Motzkin elimination to the formulation for this structure through constraints \eqref{eqn:edge}, \eqref{eqn:edge_completeness}, \eqref{eqn:vertex}  and \eqref{eqn:input} (detailed process omitted for brevity), we arrive at the convex relaxation problem LNF-relax:
%
\allowdisplaybreaks
\begin{subequations} \label{eqn:LNF_conj_disj_form}
\begin{alignat}{2}
&\underset{\boldsymbol{\xi}, \boldsymbol{y}, \boldsymbol{z}^{\pi}}{\textit{minimize}} \hspace{-1em} && \hspace{1em} \quad \boldsymbol{x}^\top\boldsymbol{Q}\boldsymbol{x} + \boldsymbol{u}^\top\boldsymbol{R}\boldsymbol{u} + \boldsymbol{\Theta}^\top \boldsymbol{z}^{\pi} \nonumber \\ 
&\textit{s.t.} \quad &&\text{for } l = 1,\ldots,L: \nonumber \\
& &&\quad \sum_{m=1}^{M_l} y_{m,l} = 1 \label{eqn:lnf_1} \\ 
& &&\text{for } j = 1,\ldots,|\Pi|: \nonumber \\
& &&\quad z^{\pi_j} \geq \sum_{\substack{m=1,\ldots,M_l \\ \pi_j \in P_{m,l}}} y_{m,l}  \label{eqn:lnf_2} \\
& &&\quad z^{\pi_j} \leq \xuanrev{1-}\sum_{\substack{m=1,\ldots,M_l \\ \neg\pi_j \in P_{m,l}}}y_{m,l}   \label{eqn:lnf_3} \\
& &&\text{for } m = 1,\ldots,M_l: \nonumber \\
& &&\quad y_{m,l} \geq 1 - |P_{m,l}| + \sum_{\pi_j \in P_{m,l}} z^{\pi_j} \nonumber \\
& &&\qquad \qquad + \sum_{\neg\pi_j \in P_{m,l}} (1-z^{\pi_j}) \label{eqn:lnf_5} \\
& &&\eqref{eqn:pwa}, \ \boldsymbol{x}_0 \in \mathcal{X}_0, \quad \eqref{eqn:predicate}, \ \forall \: k, \forall \: \pi \in \Pi \label{eqn:lnf_4}
\end{alignat}
\end{subequations}

Let $(\boldsymbol{\xi}^*, \boldsymbol{y}^*, \boldsymbol{z}^{\pi*})$ be any optimal solution for LNF-relax. We construct a solution for LT-relax by setting $\boldsymbol{z}^{\pi} = \boldsymbol{z}^{\pi*}$, $\boldsymbol{\xi} = \boldsymbol{\xi}^*$, and $z^{\varphi_{m,l}} = y_{m,l}^*$, $z^{\varphi_l} = 1$ for $l = 1,\ldots, L$ and $m = 1,\ldots, M_l$.
We aim to show that this solution is also feasible for LT-relax and achieves the same objective value, and therefore prove that $f_{\text{LT-r}}^{*} \leq f_{\text{LNF-r}}^{*}$. 

First, constraint \eqref{eqn:lnf_1} requests $\sum_{m=1}^{M_l} y_{m,l}^* = 1$, $\forall l = 1,\ldots, L$, which simultaneously guarantees the satisfaction of \eqref{eqn:lt_1}, \eqref{eqn:lt_2} and \eqref{eqn:lt_3} in LT-relax\footnote{\xuanrev{If one needs to show that \eqref{eqn:lnf_1} is equivalent to \eqref{eqn:lt_1}, \eqref{eqn:lt_2} and \eqref{eqn:lt_3}, an additional condition is needed requiring the disjunctive branches to be mutually exclusive. In Example~\ref{exp:example1}, for instance, this requires replacing $\pi_0\pi_1 \vee \pi_1\pi_2 \vee \pi_2\pi_3$ with
$\pi_0\pi_1 \vee \neg\pi_0\,\pi_1\pi_2 \vee\neg\pi_0\neg\pi_1\,\pi_2\pi_3$, so that exactly one branch is
satisfiable.}}. Next, we check constraints \eqref{eqn:lt_4} and \eqref{eqn:lt_5}. For non-negated predicates, constraint \eqref{eqn:lnf_2} implies $z^{\pi_j*} \geq \sum_{m: \pi_j \in P_{m,l}} y_{m,l}^* \geq y_{m.l}^* = z^{\varphi_{m,l}}$ for any $(l,j,m) \in \{(l,j,m) \mid l \in \{1,\ldots,L\}, \pi_j \in P_{m,l}\}$, which corresponds to constraint \eqref{eqn:lt_4}. For negated predicates, constraint \eqref{eqn:lnf_3} analogously infers the satisfaction of \eqref{eqn:lt_5}. Last, the dynamics constraints \eqref{eqn:pwa} and predicate constraints \eqref{eqn:predicate} in \eqref{eqn:lt_6} automatically hold true since $(\boldsymbol{\xi}^*, \boldsymbol{z}^{\pi*})$ are the optimal values shared by the both formulations. \qed


\end{proof}

It is obvious to show that the converse of Theorem \ref{theorem1} does not hold by finding a solution such that $f_{\text{LT-r}}^{*} < f_{\text{LNF-r}}^{*}$. Consider a counter-example with arguments $L=1$ and $M=2$, predicate sets $P_1=\{\pi_1, \pi_2\}$ and $P_2=\{\pi_2, \pi_3\}$, and uniform cost coefficients $\boldsymbol{\Theta}=[1, 1, 1]$. The atomic predicates evaluate conditions on a single integrator system $x_{k+1} = x_k + u_k$ with state and input bounds $|x_k|\leq 1, |u_k| \leq 1$. The validity mapping for the predicates are $\pi_1$: $x_1 = 1$, $\pi_2$: $x_2 = 0$, and $\pi_3$: $x_1 = -1$. These predicates are encoded using the following big-M constraints: $2z^{\pi_1}-1 \leq x_1 \leq 1$ for $\pi_1$, $z^{\pi_2}-1 \leq x_2 \leq 1-z^{\pi_2}$ for $\pi_2$, and $-1 \leq x_1 \leq 1 -2z^{\pi_3}$ for $\pi_3$. LT-relax obtains an optimal solution ($y_1^* = 0.5$, $y_2^* = 0.5$, $z^{\pi_1*} = 0.5$, $z^{\pi_2*} = 0.5$, $z^{\pi_3*} = 0.5$) with a cost of $f_{\text{LT-r}}^{*} = 1.5$. This solution, however, is infeasible for LNF-relax, which instead finds an optimal solution ($y_1^* = 1$, $y_2^* = 0$, $z^{\pi_1*} = 1$, $z^{\pi_2*} = 1$, $z^{\pi_3*} = 0$) with a cost of $2$. The binary solutions for both formulations achieve a cost of $2$, but there is a $25\%$ gap for LT whereas LNF has a zero gap. 

\begin{remark}
The flow conservation constraint plays a key role in tightening the convex relaxation. Intuitively, it ``couples" the predicates across different branches of a disjunction node, yielding multiple terms on the right-hand side of inequalities \eqref{eqn:lnf_2} and \eqref{eqn:lnf_3}. This coupling \xuanrev{would not naturally arise} if LT were used as the underlying data structure, since in an LT, each node has only a single parent node, leading to one-to-one inequality constraints as seen in \eqref{eqn:lt_4} and \eqref{eqn:lt_5}. \xuanrev{We empirically confirm this mechanism in Section~\ref{Sec:ablation_source} by manually building the aggregated inequalities \eqref{eqn:lnf_2} and \eqref{eqn:lnf_3} into the LT formulation for a specific problem, and show that the resulting formulation reproduces LNF's root relaxation gap exactly across all problem sizes. Additionally, the linear term $\boldsymbol{\Theta}^\top \boldsymbol{z}^{\pi}$ plays an important role in realizing the tightening effect: the aggregated inequalities \eqref{eqn:lnf_2} and \eqref{eqn:lnf_3} restrict the feasible values of $\boldsymbol{z}^{\pi}$ in the relaxation, which increases the relaxed optimal cost through $\boldsymbol{\Theta}^\top \boldsymbol{z}^{\pi}$.}
\end{remark}
\begin{remark}
\xuan{Formulation \eqref{eqn:LNF_conj_disj_form} is directly implementable in standard MIP solvers for temporal logic specifications in conjunctive-disjunctive form $\varphi = \bigwedge_{l=1}^{L}\bigvee_{m=1}^{M_l}\bigwedge_{n=1}^{N_{m,l}}\pi_{m,n}^l$. This is the formulation used throughout our experimental evaluation in Section~\ref{Sec:experiments}.}
\end{remark}
We also show that the LNF formulation is sound, meaning that any solution from the LNF formulation satisfies the original problem specification, and complete, meaning that it finds a solution whenever the specification is satisfiable. The proofs are provided in Appendix \ref{appendix_soundness}.

\subsection{Computational Complexity}
\label{Sec:computational_complexity}

\jimrev{The computational complexity of MICPs is often dominated by the number of binary variables, since the worst-case computational time grows exponentially with their number.} Both LNF and LT formulations introduce binary variables: LNF has edge variables $\boldsymbol{y}$ and atomic predicates $\boldsymbol{z}^{\pi}$; LT has node variables $\boldsymbol{z}^{\varphi}$ and atomic predicates $\boldsymbol{z}^{\pi}$. The atomic predicates $\boldsymbol{z}^{\pi}$ appear identically in both formulations, and their dimension is related to the planning horizon length and the number of predicates at each time step.

In addition to $\boldsymbol{z}^{\pi}$, LNF assigns a binary variable $y_e$ to each edge $e \in \mathcal{E}$, resulting in $O(|\mathcal{E}|)$ binary edge variables, where $|\mathcal{E}|$ is the total number of child nodes across all disjunction nodes in its corresponding LT. In contrast to LNF, LT formulation introduces a binary variable $z^{\varphi_i}$ for each internal node $\varphi_i$ in the tree. Bounded-time temporal operators like $\Diamond_{[t_1,t_2]}$ lead to $t_2-t_1+1$ internal nodes per operator. Therefore, the number of internal variables is proportional to both the number of operators and the size of their associated time bounds. In most practical logic specifications, the additional binary variables introduced by LNF ($\boldsymbol{y}$) and LT ($\boldsymbol{z}^{\varphi}$) are approximately equal in number. The only exception occurs in cases where edges contain single predicates, where LNF may introduce additional edge variables compared to LT.

Both LNF and LT formulations involve continuous variables $\boldsymbol{\xi}$ associated with the underlying dynamics. Assuming they use the same dynamics formulation, the numbers of these variables are identical. It is worth noting that the Fourier-Motzkin process removes the continuous flow variables $\boldsymbol{\omega}$ from the LNF formulation, avoiding the additional computational overhead that would arise if they were retained.





Regarding constraints, LNF incorporates flow conservation constraints (\ref{eqn:vertex-a}) and (\ref{eqn:input-a}) for each vertex in $\mathcal{V}$. The Fourier-Motzkin process also generates $O(|\Pi|)$ inequality constraints per vertex as it propagates the flow from source $v_s$ sequentially towards target $v_t$. However, many of these constraints are redundant and can be eliminated. Therefore, while the theoretical number of constraints is $O(|\mathcal{V}| \cdot |\Pi|)$, it typically requires significantly fewer constraints in practice. The number of constraints for LT formulation is also proportional to the number of internal nodes in the tree, as each internal node imposes individual constraints relating its associated variable to its parent.
For the specification $\varphi$ assumed in Theorem \ref{theorem1}, within each disjunction section $l$ ($l=1,\ldots,L$), LNF assigns one inequality per unique predicate, resulting in a total of $O(|\Pi|)$ constraints per section and $O(L\cdot|\Pi|)$ constraints overall. In comparison, LT defines one inequality for each instance of a predicate across the $M_l$ conjunctive terms, leading to $O(M_l\cdot|\Pi|)$ constraints per section and $O(\sum_{l=1}^L M_l\cdot|\Pi|)$ constraints in total.

Overall, LNF formulations provide tighter convex relaxations compared to LT formulations, while generally not increasing the number of variables and constraints. This typically results in faster solving speeds, as demonstrated in our experimental results in Section \ref{Sec:experiments}.

\section{Modeling of Dynamical Systems}
\label{Sec:DNF}
While the LNF framework provides an efficient representation for temporal logic specifications, its practical application to robotics requires integration with appropriate dynamics models. In this section, we review a series of existing approaches of modeling robot dynamics that can be integrated with our LNF framework. This integration forms a comprehensive framework, allowing users to select the dynamical system representation that best suits their specific robotic applications. 

\subsection{Abstraction of Dynamics in Discrete Configuration Space using Dynamic Network Flow}
\label{sec:dnf_1}

The first approach abstracts the dynamics constraint in Eqn.~\eqref{eqn:dyn} through a Dynamic Network Flow (DNF) representation \citep{aronson1989survey}. A similar strategy has been adopted in \citep{kurtz2021more} for motion planning under MTL specifications. 
This approach begins with constructing a \textit{temporal graph} where the discrete configuration space $\mathcal{X} = \{\boldsymbol{p}^1, \boldsymbol{p}^2, \ldots, \boldsymbol{p}^m\}$ is represented by a set of vertices $\{p^1, p^2, \ldots, p^m\}$. Each vertex $p^i$ corresponds to a physical location $\boldsymbol{p}^i$, and the vertices are connected by directed edges (\textit{e.g.}, from $p^i$ to $p^j$) with associated travel times and costs. This graph structure is then transformed into a DNF by expanding it along the time dimension. Although this concept has appeared in several works (\textit{e.g.}, \citep{yu2013multi}), we provide an explicit definition of temporal graph here for completeness.

\begin{definition}
(Temporal Graph)
A Temporal Graph is a tuple $\mathcal{G}_d = (\mathcal{V}_d, \mathcal{E}_d, T_d, C_d, \theta, \mathcal{V}_s)$, where:
\begin{itemize}
    \item $\mathcal{V}_d = \{p^1, p^2, \ldots, p^m\}$ is a finite set of vertices where each vertex $p^i$ represents a spatial location $\boldsymbol{p}^i \in \mathcal{X}$ in the configuration space.
    \item $\mathcal{E}_d \subseteq \mathcal{V}_d \times \mathcal{V}_d$ is a set of directed edges, each $e_d = (p^i, p^j) \in \mathcal{E}_d$ corresponding to a movement from location $\boldsymbol{p}^i$ to location $\boldsymbol{p}^j$.
    \item $T_d: \mathcal{E}_d \rightarrow \mathbb{R}^+$ is a function that assigns a positive travel time to each edge.
    \item $C_d: \mathcal{E}_d \rightarrow \mathbb{N}^+$  assigns a positive integer capacity to each edge indicating the maximum number of robots that can traverse the edge at the same time. 
    \item $\theta_k: \mathcal{E}_d \rightarrow \mathbb{R}^+$ is a function that assigns a (potentially time-varying) positive cost to each edge representing the cost of traversing that edge at time step $k$, including self-loops representing the cost of remaining at a vertex for one time unit at time step $k$.
    \item $\mathcal{V}_s \subseteq \mathcal{V}_d$ is a set of source vertices aggregating initial positions of all robots.
\end{itemize}
\label{def:temporal_graph}
\end{definition}
\begin{remark}
Motion planning on temporal graphs under temporal logic constraints is NP-hard. Consider a simple temporal graph with only two vertices, where one vertex represents ``true" and the other represents ``false." The robot transitions to, or remains at, the ``true" vertex if a logical condition is satisfied, and moves to the ``false" vertex otherwise. Solving a Boolean satisfiability (SAT) problem reduces to finding a valid robot trajectory satisfying the encoded logical formula on this specific temporal graph.
\end{remark}
\begin{figure}[t!]
    \centering
    \includegraphics[width=0.5\textwidth]{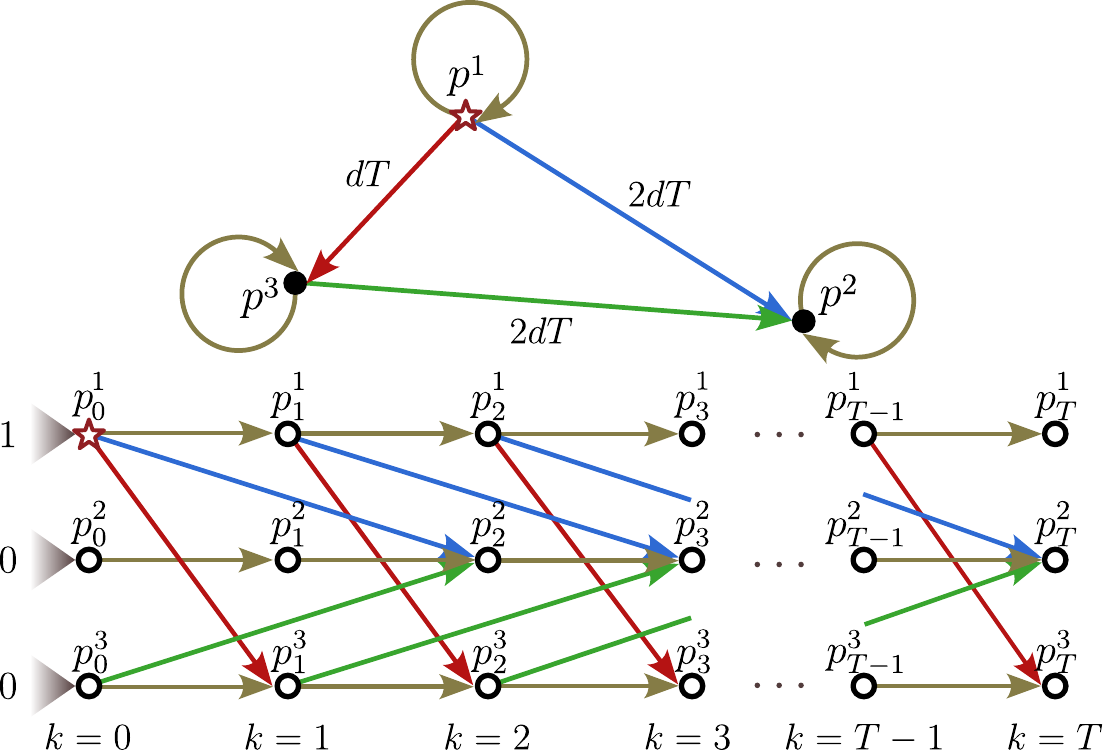}
     \caption{Conversion from a temporal graph to a DNF. \textit{Upper}: A temporal graph consisting of three vertices $p^1$, $p^2$, $p^3$ with travel times $T_d(p^1,p^2)=2dT$, $T_d(p^1,p^3)=dT$, and $T_d(p^3,p^2)=2dT$. \textit{Lower}: The converted DNF with time horizon $T$, where each vertex $p^i$ is expanded into an array of time-indexed vertices $p^i_0, p^i_1, \ldots, p^i_T$. Red and green arrows represent robots moving from one location to another, while golden arrows indicate robots staying at the same location. The robot initially starts from vertex $p^1$ depicted by a star icon.}
\label{fig:dnf}
\end{figure} 
A temporal graph can be constructed for a robotic system of interest by collecting a library of trajectories offline, followed by building a graph structure based on the library. This is also referred to as experience-based planning \citep{coleman2015experience}, where a database stores memorized trajectories that can be retrieved and adapted to generate new trajectories. This setup is particularly beneficial for large configuration spaces that include many invariant constraints, such as planning within a (mostly) static environment.

We first present the construction of a single robot system. This simplification obviates the need to consider collision avoidance between robots. It starts with identifying the locations in the configuration space that the temporal logic specifications may require the robot to visit (\textit{i.e.}, those corresponding to atomic predicates for which $z^{\pi^i_k}=1$). For each pair of connected locations $\boldsymbol{p}^i$ and $\boldsymbol{p}^j$, we generate a feasible trajectory 
that \jimrev{satisfies} the robot's dynamics constraints using motion planning algorithms. Each feasible trajectory matches a directed edge $e = (p^i, p^j)$ in the temporal graph, associated with a travel time $T_d(e)$ representing a (preferably tight) upper bound on the actual traversal time from $\boldsymbol{p}^i$ to $\boldsymbol{p}^j$, a cost $\theta(e)$ reflecting factors like energy consumption or terrain traversability, and a capacity $C_d(e)=1$ limiting passage to a single robot. Additionally, we assign a cost $\theta(p^i)$ to each vertex $p^i$ representing the cost of remaining at that location for a unit of time, to capture factors such as weather severity.

With an established temporal graph, we proceed to build a DNF with a time horizon $T$. For each vertex $p^i \in \mathcal{V}_d$, we index it by discrete time steps to produce a new array of vertices $p^i_0, p^i_1, \ldots, p^i_T$, where each $p^i_k$ corresponds to a visiting location $\boldsymbol{p}^i$ at time $k \cdot dT$. The time discretization $dT$ is selected as the greatest common divisor of all travel times such that $T_d(e)/dT$ take integer values (rounding up $T_d(e)$ if necessary to meet integer multiples of $dT$). If traversing along an edge $e = (p^i, p^j)$ takes $T_d(e)/dT$ time steps, edges are connected from $p^i_k$ to $p^j_{k+T_d(e)/dT}$, for any $k$ such that $k+T_d(e)/dT \leq T$. Moreover, for each vertex $p^i \in \mathcal{V}_d$ and time step $k$ such that $k+1 \leq T$, edges are also introduced from $p^i_k$ to $p^i_{k+1}$, representing the robot's staying at the same location between consecutive time steps. Fig.~\ref{fig:dnf} illustrates this conversion from a temporal graph to a DNF.

DNF can be formulated as an LP problem, similar to a standard network flow optimization. Let $\mathcal{E}_{\text{DNF}}$ denote the set of all edges in the DNF. For each edge $e \in \mathcal{E}_{\text{DNF}}$ corresponding to $(p^i, p^j) \in \mathcal{E}_d$ starting at time step $k$, we associate a continuous flow variable $r_e \in [0, 1]$, upon which we impose flow conservation constraints on source vertices $\mathcal{V}_s$ and all other vertices through:
\begin{subequations}
\begin{align}
    \sum_{e \in \mathcal{E}^{\rm in}_{p^i_k}} r_e &= \sum_{e \in \mathcal{E}^{\rm out}_{p^i_k}} r_e, \; \forall p^i \in \mathcal{V}_d, \; k = 1, \ldots, T \\
    \sum_{e \in \mathcal{E}^{\rm out}_{p_0^i}} r_e &= 1, \; \forall p^i \in \mathcal{V}_s \\
    \sum_{e \in \mathcal{E}^{\rm out}_{p_0^i}} r_e &= 0, \; \forall p^i \in \mathcal{V}_d \setminus \mathcal{V}_s
\end{align}
\label{eqn:DNF_constraints}%
\end{subequations}
where $\mathcal{E}^{\rm in}_{p_k^i}$ and $\mathcal{E}^{\rm out}_{p_k^i}$ denote the sets of incoming and outgoing edges, respectively, for vertex $p_k^i$.

Also for each edge $e \in \mathcal{E}_{\text{DNF}}$ corresponding to $e' \in \mathcal{E}_d$, the flow incurs a cost $\Tilde{\theta}(e) \cdot r_e$, proportional to the amount of flow $r_e$, where $\Tilde{\theta}(e) = \theta_k(e')$. This cost expression applies to both robots moving between different locations in the temporal graph and those remaining at the same locations. Therefore, the overall objective function is:
\begin{equation}
f_{\text{obj}}(\boldsymbol{r}) = \sum_{e \in \mathcal{E}_{\text{DNF}}} \Tilde{\theta}(e) \cdot r_e
\label{eqn:obj}
\end{equation}
where $\boldsymbol{r} = [r_{e_1}, r_{e_2}, \ldots, r_{e_{|\mathcal{E}_{\text{DNF}}|}}]^\top \in [0,1]^{|\mathcal{E}_{\text{DNF}}|}$.

To define the atomic predicates in discrete vertex space of DNF, for each predicate $\pi$ we associate the set $\mathcal{P}^\pi \subseteq \mathcal{V}_d$ containing vertices $p$ whose physical locations in the temporal graph $\boldsymbol{p}$ satisfy $({\boldsymbol{a}^{\pi}})^\top \boldsymbol{p} + b^\pi \geq 0$. At any time step, the predicate $\pi$ is considered satisfied if the robot traverses an edge that leads into a vertex in $\mathcal{P}^\pi$. This is expressed as:
\begin{equation}
\bigvee\nolimits_{p \in \mathcal{P}^\pi, \, e \in \mathcal{E}_{p_k}^{\rm in}} r_e \Leftrightarrow z^{\pi_k}=1
\label{eqn:predicate_dnf}
\end{equation}
\noindent In essence, the edge variables $r_e$ in the DNF, or their disjunctions, can be directly considered as the atomic predicates $\boldsymbol{z}^{\pi}$.

Through constraints \eqref{eqn:DNF_constraints}, DNF defines a special type of dynamics on the discrete space $\mathcal{X}$. The robot positions evolve following $\boldsymbol{p}^j_{k+T_d(e_d)/dT} = \boldsymbol{p}^i_k + \boldsymbol{u}_k$, where $e_d = (p^i, p^j)$ and $\boldsymbol{u}_k$ represent the discrete transition chosen at time step $k$. This discrete transition system can be viewed as a special case of the PWA dynamics. With the DNF formulation, the state variables $\boldsymbol{\xi}$ are now superseded by flow variables $\boldsymbol{r}$, and the general atomic predicate \eqref{eqn:predicate} is replaced by constraint \eqref{eqn:predicate_dnf}. We have:

\begin{subformulation}
(Optimization Formulation for Temporal Logic Motion Planning with DNF Dynamics)
\begin{align*}
&\underset{\boldsymbol{r}, \boldsymbol{\gamma}, \boldsymbol{z}^{\pi}}{\text{minimize}} \quad \eqref{eqn:obj} \nonumber \\
& \begin{aligned}
\text{s.t.} &\quad\;\; \text{Eqn. } \eqref{eqn:DNF_constraints} & \text{(DNF dynamics)} \\
& \quad\;\; \text{Eqn. } \eqref{eqn:predicate_dnf},\ \ \forall \: k, \forall \: \pi \in \Pi & \text{(Atomic predicates)} \\
& \quad\;\; \mathcal{C}_{\text{logic}}(\boldsymbol{\gamma}, \boldsymbol{z}^{\pi}) \leq 0 & \text{(Temporal logics)}\\
\end{aligned}
\end{align*}
\label{eqn:logic_dnf_formulation}
\end{subformulation}
Finally, we show a lemma arguing that 
\xuanrev{for the single-robot DNF ($|\mathcal{V}_s| = 1$) with $\boldsymbol{z}^{\pi}$ enforced as binary values},
the DNF subset of formulation \eqref{eqn:general_formulation}, including the objective function \eqref{eqn:obj}, and constraints \eqref{eqn:DNF_constraints} and \eqref{eqn:predicate_dnf}, possesses a tight LP relaxation. That is, we do not need to explicitly require $\boldsymbol{r}$ to be binary, as their optimal values will inherently take binary values. This justifies the selection of DNF to represent dynamics models, as when integrated with LNF, it preserves the tightness of the convex relaxation. The proof follows directly from standard network flow theory, which is omitted here for brevity, and readers can refer to Chapter 11.12 of \citep{ahuja1994network}. 

\begin{lemma}
\label{lemma:dnf_tight_lp}
(Tight LP Relaxation of DNF)
\xuanrev{For a single-robot DNF with $|\mathcal{V}_s| = 1$},
the optimal solution  $\boldsymbol{r}$ of the following formulation takes binary values:
\begin{align*}
&\underset{\boldsymbol{r}}{\text{minimize}} \quad \eqref{eqn:obj} \nonumber \\
& \begin{aligned}
\text{s.t.} &\quad\;\; \text{Eqn. } \eqref{eqn:DNF_constraints}, \ \ \xuanrev{|\mathcal{V}_s| = 1} & \text{(DNF dynamics)} \\
& \quad\;\; \text{Eqn. } \eqref{eqn:predicate_dnf},\ \ \forall \: k, \forall \: \pi \in \Pi & \text{(Atomic predicates)} \\
& \quad\;\; \boldsymbol{z}^{\pi} \text{ fixed to binary values} &
\end{aligned}
\end{align*}
\end{lemma}

\begin{remark} \label{remark:binary_encoding}
Rather than enforcing $\boldsymbol{z}^{\pi}$ as binary variables, an alternative is to keep $\boldsymbol{z}^{\pi}$ continuous and require the LNF edge variables $y_e$ to be binary, since disjunctions are the source of discrete decisions in the logic structure that requires binary variables. \xuanrev{This aligns with the encoding strategy of \citet{kurtz2022mixed}.} Furthermore, we can employ encoding techniques that use fewer binary variables, as also proposed by~\citet{kurtz2022mixed}, where only $\lceil \log_2(n) \rceil$ binary variables are demanded for $n$ disjunctive choices. This reduction in binary variables \xuanrev{could} lead to additional computational speedups.
\end{remark}

\subsection{Incorporating Multi-robot Planning with Collision Avoidance on Dynamic Network Flow}
\label{sec:dnf_2}
Collision avoidance is an indispensable factor to consider in multi-robot motion planning, whether homogeneous (\textit{e.g.}, a team of ground vehicles) or heterogeneous (\textit{e.g.}, ground and aerial platforms, or manipulators mounted at different locations). To address potential conflicts, pairwise collision checking is conducted on the temporal graphs generated from trajectory libraries of different robots. 
Three types of collision scenarios are identified in our study:

\begin{enumerate}
\item \textit{Vertex-to-vertex collisions.}
This occurs when two robots attempt to occupy the same spatial location simultaneously, or when two robots attempt to occupy different spatial locations but are too close to each other given the robots' physical dimensions.
\item \textit{Vertex-to-edge collisions.} This occurs when a stationary robot is at a spatial location that interferes with another robot's trajectory.
\item \textit{Edge-to-edge collisions.} This occurs when robots are moving head-to-head along the same path in opposite directions, or robots are moving along different paths but colliding at a finite or infinite number of intersection points given the robots' physical dimensions.
\end{enumerate}

All three types of collisions can be resolved through designing additional constraints in the DNF formulation. Consider $R$ robots in the system, where each robot $\ell \in \{1,...,R\}$ has its own temporal graph $\mathcal{G}_d^\ell = (\mathcal{V}_d^\ell, \mathcal{E}_d^\ell, T_d^\ell, C_d^\ell, \theta^\ell, \mathcal{V}_s^\ell)$ with corresponding flow variables $r_e^\ell$ for the edges in the associated DNF. We build conflict sets $\mathcal{C}$ comprising robot-vertex pairs $(\ell,p)$ and robot-edge pairs $(\tilde{\ell}, e)$ that cannot be active simultaneously. For vertex-to-vertex collisions, $\mathcal{C}$ contains only robot-vertex pairs $(\ell,p)$; \xuanrev{\textit{e.g.}, $\mathcal{C} = \{(\ell_1, p), (\ell_2, p')\}$ encodes that robot $\ell_1$ occupying vertex $p$ and robot $\ell_2$ occupying a nearby vertex $p'$ would collide}. For edge-to-edge collisions, $\mathcal{C}$ contains only robot-edge pairs $(\tilde{\ell}, e)$; \xuanrev{ \textit{e.g.}, $\mathcal{C} = \{(\ell_1, e), (\ell_2, e')\}$ encodes that robot $\ell_1$ traversing edge $e$ and robot $\ell_2$ traversing an intersecting edge $e'$ would collide.} For vertex-to-edge collisions, either type may appear in $\mathcal{C}$. Inspired by \citep{yu2016optimal}, we apply the following constraint for every such conflict set:
\begin{equation}
\sum_{(\ell,p) \in \mathcal{C}} \sum_{e \in \mathcal{E}^{\rm in}_{p_k}} r_e^\ell + \sum_{(\Tilde{\ell}, e) \in \mathcal{C}} r_{e}^{\Tilde{\ell}} \leq 1, \; \forall k
\label{eqn:DNF_collision_avoidance}
\end{equation}
This constraint ensures that at most one element from each conflict set can be active at any given time step, which prevents all three types of collisions.

In particular, when the robot team is homogeneous, a single DNF is sufficient to represent all robots. In this scenario, vertex-to-vertex collisions can be avoided by limiting the input degree of each vertex to at most one ($\sum_{e \in \mathcal{E}^{\rm in}_{p_k}} r_e \leq 1$), while edge-to-edge collisions can be eliminated by setting the edge capacity $C_d(e) = 1$.
However,  \eqref{eqn:DNF_collision_avoidance} breaks the tight LP relaxation property established in Lemma $2$ for single-robot DNF. Therefore, all edge variables $r_e^\ell$ in \eqref{eqn:DNF_collision_avoidance} must be explicitly set as binary variables.

\subsection{Incorporating Dynamics in Continuous Configuration Space}
\label{sec:continuous_dynamics_bigM}

While the prior approaches abstract the dynamics in discrete configuration space as a special case of PWA dynamics, here we present the big-M formulation for the general class of PWA dynamics in continuous configuration space, although alternative formulations, such as the convex hull formulation, also exist. Readers are referred to \citep{marcucci2019mixed} for a comprehensive comparison of different PWA formulations.

We model \xuanrev{PWA dynamics} using the standard big-M approach \citep{bemporad1999control} because it is widely received due to its ease of implementation and general effectiveness with modern MIP solvers, though it can yield loose convex relaxations when the big-M constants are large. For each polytope $\mathcal{D}^i$, $i \in \mathcal{I}$, we introduce a binary variable $\delta_{k}^i$ to indicate if the system state $\boldsymbol{x}_k$ and input $\boldsymbol{u}_k$ at time step $k$ lie within the polytope. The following constraints ensure that exactly one polytope constraint is satisfied for the system state and control input at any time step:
\begin{subequations}
\begin{align}
    \boldsymbol{x}_{k+1} - \boldsymbol{A}^i\boldsymbol{x}_k - \boldsymbol{B}^i\boldsymbol{u}_k  &\geq -\sum_{j \in \mathcal{I}\backslash\{i\}} \delta^j_k\boldsymbol{M}^{ij}_1 \label{eqn:pwa_as_bigM_1} \\
    \boldsymbol{x}_{k+1} - \boldsymbol{A}^i\boldsymbol{x}_k - \boldsymbol{B}^i\boldsymbol{u}_k &\leq \sum_{j \in \mathcal{I}\backslash\{i\}} \delta^j_k \boldsymbol{M}^{ij}_2 \label{eqn:pwa_as_bigM_2} \\
    \boldsymbol{H}^i_1\boldsymbol{x}_k + \boldsymbol{H}^i_2\boldsymbol{u}_k &\leq \boldsymbol{h}^i + \sum_{j \in \mathcal{I}\backslash\{i\}} \delta^j_k \boldsymbol{M}^{ij}_3 \label{eqn:pwa_as_bigM_3} \\
    \sum_{i\in\mathcal{I}} \delta^i_k &= 1 \label{eqn:pwa_as_bigM_4}
\end{align} 
\label{eqn:pwa_as_bigM}%
\end{subequations}
The constraints above collectively instantiate the big-M implementation for constraint~\eqref{eqn:pwa}. 
Constraints \eqref{eqn:pwa_as_bigM_1}, \eqref{eqn:pwa_as_bigM_2}, and \eqref{eqn:pwa_as_bigM_3} enforce that if polytope $\mathcal{D}^i$ is active (\textit{i.e.}, $\delta^i_k=1$), the associated dynamics and domain constraints of mode $i$ are satisfied. Otherwise, when $\delta^i_k=0$, these constraints become inactive due to the big-M terms. Meanwhile, constraint \eqref{eqn:pwa_as_bigM_4} ensures that only one polytope is active at each time step.

Smaller $\boldsymbol{M}^{ij}_1$, $\boldsymbol{M}^{ij}_2$, $\boldsymbol{M}^{ij}_3$ values provide tighter convex relaxations, which generally leads to better computational performance. These values for each pair of modes $(i,j) \in \mathcal{I}\times\mathcal{I}$ can be minimized through:
\begin{align*}
    \begin{bmatrix}
    \boldsymbol{M}^{ij}_1 \\
    \boldsymbol{M}^{ij}_2 \\
    \boldsymbol{M}^{ij}_3
\end{bmatrix}
:= \max_{(\boldsymbol{x}_k, \boldsymbol{u}_k)\in\mathcal{D}^j}
\begin{bmatrix}
    -(\boldsymbol{A}^i + \boldsymbol{A}^j)\boldsymbol{x}_k - (\boldsymbol{B}^i + \boldsymbol{B}^j)\boldsymbol{u}_k \\
    (\boldsymbol{A}^i - \boldsymbol{A}^j)\boldsymbol{x}_k + (\boldsymbol{B}^i - \boldsymbol{B}^j)\boldsymbol{u}_k \\
    - \boldsymbol{h}^i+\boldsymbol{H}^i_1\boldsymbol{x}_k + \boldsymbol{H}^i_2\boldsymbol{u}_k
\end{bmatrix}
\end{align*}
such that when mode $j\neq i$ is selected ($\delta^j_k=1$), the right-hand sides of constraints \eqref{eqn:pwa_as_bigM_1}, \eqref{eqn:pwa_as_bigM_2}, and \eqref{eqn:pwa_as_bigM_3} become sufficiently large to render these constraints inactive.

\xuanrev{Eqn.} \eqref{eqn:predicate} can be used straightforwardly to define atomic predicates for the big-M formulation of PWA dynamics. \xuan{When $\boldsymbol{H}^i_2=\boldsymbol{0}$ for all $i \in \mathcal{I}$, the big-M formulation builds a direct connection between binary variables and polytope membership.
This is because constraint \eqref{eqn:pwa_as_bigM_4} enforces that exactly one $\delta^i_k$ equals 1, which makes constraint \eqref{eqn:pwa_as_bigM_3} equivalent to $\boldsymbol{H}^i_1\boldsymbol{x}_k \leq \boldsymbol{h}^i$ when $\delta^i_k=1$.}
Therefore, by defining $g^{\pi^i}(\boldsymbol{x}_k) = -\boldsymbol{H}^i_1\boldsymbol{x}_k + \boldsymbol{h}^i$, the binary variable $\delta^i_k$ functions exactly same as the predicate variable $z^{\pi^i_k}$.

\xuan{Combining the big-M formulation for the general class of PWA dynamics with our LNF, we present the complete optimization formulation for Problem \ref{prob1_2}:}
\begin{subformulation}
(Optimization Formulation for Temporal Logic Motion Planning with Big-M Formulation of PWA Dynamics)
\begin{align*}
&\underset{\boldsymbol{\xi}, \boldsymbol{\gamma}, \boldsymbol{z}^{\pi}}{\text{minimize}} \quad f_{\text{obj}}(\boldsymbol{\xi}, \boldsymbol{z}^{\pi}) \nonumber \\
& \begin{aligned}
\text{s.t.} &\quad\;\; \text{Eqn. } \eqref{eqn:pwa_as_bigM}, \quad \boldsymbol{x}_0 \in \mathcal{X}_0 & \text{(big-M PWA dynamics)} \\
& \quad\;\; \text{Eqn. } \eqref{eqn:predicate},\ \ \forall \: k, \forall \: \pi \in \Pi & \text{(Atomic predicates)} \\
& \quad\;\; \mathcal{C}_{\text{logic}}(\boldsymbol{\gamma}, \boldsymbol{z}^{\pi}) \leq 0 & \text{(Temporal logics)}\\
\end{aligned}
\end{align*}
\label{eqn:logic_bigM_formulation}
\end{subformulation}

\begin{remark}
\label{remark1}
Instead of using the full PWA dynamics formulation in \eqref{eqn:pwa_as_bigM}, we can leverage Bézier curves to model the coordinates of interest for planning (such as CoM position), which can reduce the number of variables and constraints. This approach is used in the GCS formulation \citep{marcucci2023motion} for MIP of motion planning around obstacles, where the convex hull property of Bézier curves ensures obstacle avoidance. The assignment of predicates follows a similar approach to the DNF formulation in \eqref{eqn:predicate_dnf}, where binary variables indicate which polytope each curve segment occupies. We adopt this approach in our hardware experiments in Section \ref{sec:experiment_hardware}.
\end{remark}

\section{Experiments}
\label{Sec:experiments}
We conduct comprehensive experiments to evaluate the computational advantages of the LNF formulation compared to the LT formulation. Our experimental evaluation includes four types of studies that demonstrate the versatility and scalability of our approach: (i) vehicle routing problems with time windows using DNF models, (ii) multi-robot coordination using pre-computed trajectory libraries, (iii) optimal motion planning with PWA dynamics using big-M method, and (iv) hardware experiments demonstrating real-time replanning capabilities with quadrupedal robots traveling in dynamic environments.

For each study, we formulate the task specification using both our proposed LNF Formulation \ref{eqn:sub_formulation_2} and the baseline LT Formulation \ref{eqn:sub_formulation_1}, comparing their performance across the convex relaxation tightness measured by the root relaxation gap, the computational efficiency measured by solution times, \xuanrev{and the memory consumption measured by peak memory usage.} We conduct multiple trials with randomized parameters to ensure statistical significance. For a comprehensive comparison, we evaluate both formulations using two state-of-the-art MIP solvers: Gurobi $12.0$ and CPLEX 22.1.1.0, both under default settings. All experiments run on a computer with 12th Gen Intel Core i7-12800H CPU and 16GB RAM. \xuanrev{We measure peak memory consumption using the standard monitoring tool \texttt{psrecord}, which periodically samples the memory consumption of the solver process and records the maximum usage over the entire solving run.}

\subsection{Path Planning on Temporal Graph Formulated as Dynamic Network Flow}
\label{Sec:temporal_graph_experiment}
\jimrev{First, we evaluate our LNF formulation on temporal graphs formulated as DNF, as described in Section~\ref{sec:dnf_1}. We provide two sets of experiments featuring single- and multi-robot scenarios, respectively. The first experiment considers a task involving multiple target groups, where a robot must visit at least one target from each group. Following the setup in~\citep{kurtz2021more} to establish a direct comparison baseline, we benchmark our proposed Formulation~\ref{eqn:logic_dnf_formulation} against their LT formulation, which is identical to our Formulation~\ref{eqn:sub_formulation_1}.}

The workspace is modeled as a $16 \times 16$ grid world populated with randomly distributed targets and obstacles. \jimrev{In the temporal graph, the center of each grid cell corresponds to a vertex, and an edge is introduced between two cells whenever the robot can move directly between them. Horizontal and vertical motions generate edges with duration $2dT$, while diagonal motions generate edges with duration $3dT$.} We set the number of target groups as $N_g$, and the number
of targets in each group as $N_t$. Similar to~\citep{kurtz2021more}, we vary $N_g$ from $2$ to $5$, set the planning horizon to $10N_g$ time steps, and place $2N_g$ obstacles in the environment. The robot must visit at least one target among each group while avoiding all obstacles. This temporal logic specification is given by:
\begin{equation}
\begin{aligned}
\varphi_{\text{single}}
=&\; \square_{[0,T]} \neg \text{obstacle} \\
&\land \bigwedge_{i=1}^{N_g}
\left[
\Diamond_{[0,T]}
\left(
\bigvee_{j=1}^{N_t}
\square_{[0,1]} \text{target}^{i,j}
\right)
\right]
\end{aligned}
\label{eqn:multi_target_spec}
\end{equation}
where $\text{target}^{i,j}$ denotes the $j$-th target in group $i$. We place $N_t = 3$ targets in each group, and $T$ is the planning horizon.

\begin{figure}[t!]
    \centering
    \includegraphics[width=0.48\textwidth]{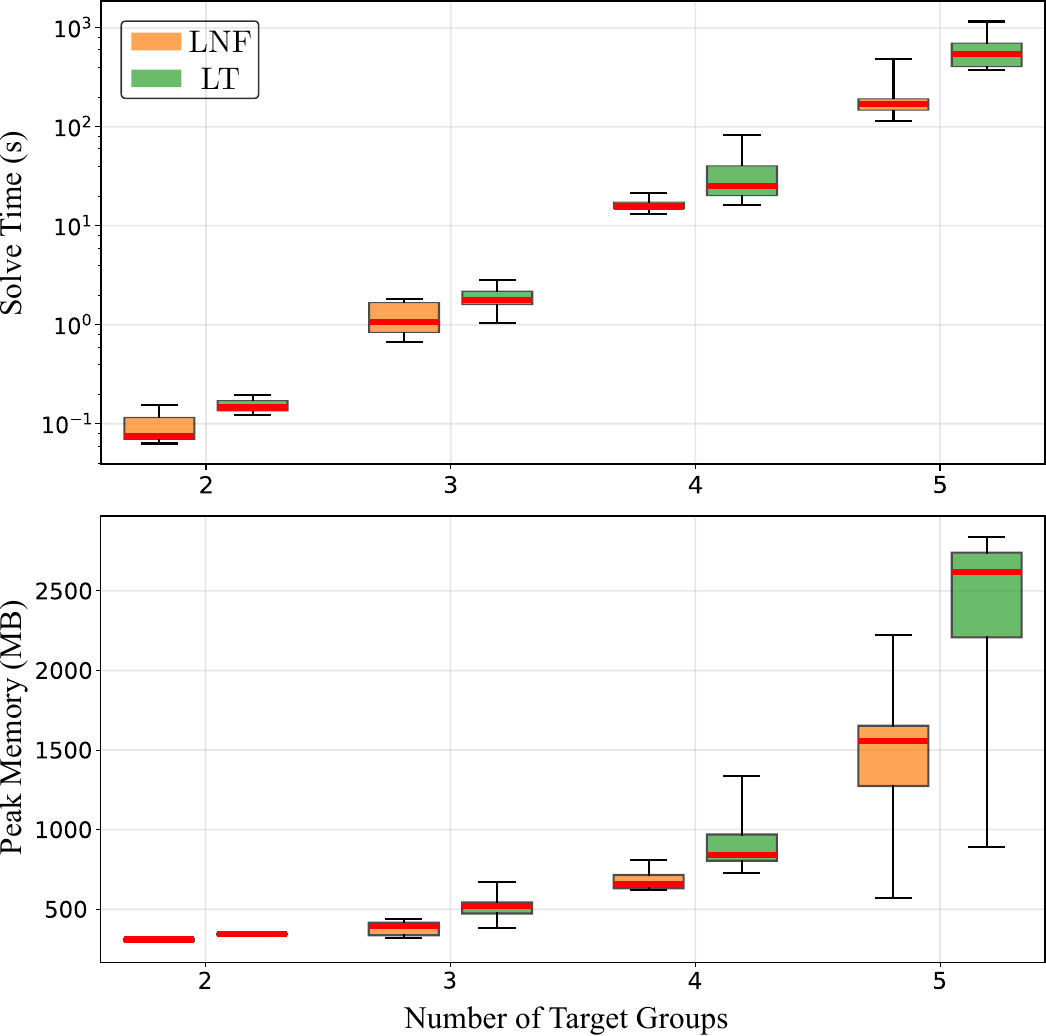}
    \caption{Scalability comparison between the LNF and LT formulations for multi-target scenarios with varying number of target groups $N_g$. $x$-axis: number of target groups ($N_g$); $y$-axis (T
    op): solving time to prove global optimality (seconds); \jimrev{$y$-axis (Bottom): peak memory consumption (MB).}}
\label{fig:single_robot}
\end{figure}

\textbf{Results} 
We conduct multiple trials for each $N_g$ under different target and obstacle placements. \xuanrev{The results are collected in Fig.~\ref{fig:single_robot}, where the upper plot reports the solving time to prove global optimality, and the lower plot reports the peak memory consumption. In each box plot, the horizontal red line marks the median, the colored box spans the upper and lower quartiles, and the whiskers indicate the full range.}

For smaller problem sizes ($N_g = 2,3$), all solving times of the baseline formulation are within a few seconds, which is consistent with~\citep{kurtz2021more}. However, our proposed LNF formulation demonstrates consistent speedups \jimrev{and lower momory consumption} across all problem sizes. For smaller problem sizes such as $N_g = 2, 3$, LNF achieves a median solving time that is $2 \times$ faster than LT. As the problem size increases, the improvement becomes more significant: for $N_g = 5$, LNF achieves a 3$\times$ speedup with a median solving time of $170$~s compared to $540$~s for LT. We attribute the superior performance of LNF to the significantly tighter convex relaxations. The average root relaxation gaps for LNF are $10.16\%$, $24.27\%$, $37.46\%$, and $51.78\%$ for $N_g = 2, 3, 4, 5$, respectively, while those for LT are $31.05\%$, $46.81\%$, $56.49\%$, and $65.89\%$. This shows that LNF consistently achieves tighter relaxation gaps than LT for each problem size. This allows the B\&B process to prune the tree more efficiently and avoid unnecessary branching by detecting earlier if certain nodes are not worth further exploration.

\xuanrev{The lower plot of Fig.~\ref{fig:single_robot} shows that LNF also consistently consumes less peak memory than LT, with the gap widening as the problem size grows. For small problems ($N_g = 2$) the two formulations are comparable, whereas for $N_g = 5$ LNF requires a median of roughly $1600$ MB versus $2700$ MB for LT, a $1.7\times$ reduction. We attribute this reduction to a combination of fewer constraints in the LNF formulation and its faster solving time: the smaller model size requires less memory for storage, while the faster convergence implies that fewer B\&B nodes need to be managed during the optimization process.}

\begin{figure}[t!]
    \centering
    \includegraphics[width=0.45\textwidth]{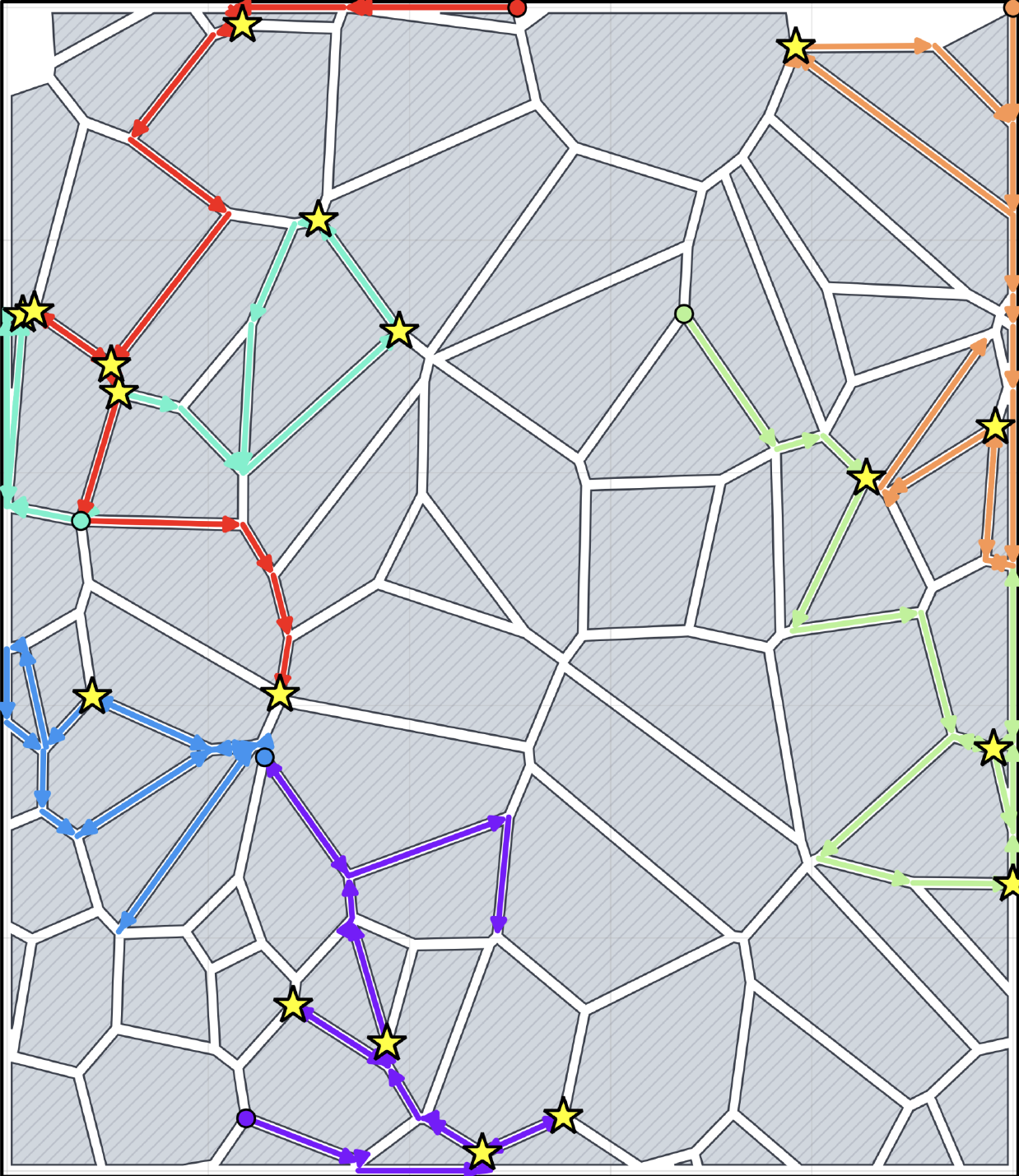}
    \caption{Example VRPTW environment with planned robot trajectories. The designated locations are shown by yellow stars. The initial positions of the robots are marked by the colored circles. Different colored paths represent trajectories for different robots visiting designated locations within their time windows.}
\label{fig:vrptw_map}
\end{figure}

\vspace{\baselineskip}
The second set of experiments considers the Vehicle Routing Problem with Time Windows (VRPTW), where a fleet of $R$ homogeneous robots must visit specific locations within their designated time windows $[0,T]$. Each location has a required service duration. To create a challenging environment with intertwined pathways, we segment the map into polygons and shrink them to create paths and intersections. Each intersection corresponds to a vertex in the temporal graph, and each path represents an edge. The travel time for each path is computed by dividing the travel distance by a constant speed. An example of such an environment and the resulting planned paths is shown in Fig.~\ref{fig:vrptw_map}.

We examine two variants of specifications. The first specification requires at least one robot to visit each location within its time window and execute the task for a specified duration:
\begin{equation}
\varphi_{\text{vrptw}} = \bigwedge_{t=1}^K \bigvee_{r=1}^R (\Diamond_{[0,T]} \Box_{[0,2]} \pi^{r,t})
\end{equation}
where $\pi^{r,t}$ is a predicate indicating that robot $r$ is at the target location specified for task $t$.

The second specification captures sequencing dependencies between robots, where one robot must first reach a location before another robot can access a different location. This coordination is expressed as:
\begin{equation}
\begin{split}
\varphi_{\text{sequential}} = \bigwedge_{i=1}^{N_{\text{seq}}} \bigvee_{r_1,r_2 \in R} &((\neg \pi^{r_2,B^i} \mathcal{U} \pi^{r_1,A^i}) \wedge \\ &(\Diamond_{[0,T]}\Box_{[0,2]}\pi^{r_2,B^i}))
\end{split}
\end{equation}
where $\pi^{r_1,A^i}$ indicates robot $r_1$ is at location $A^i$, $\pi^{r_2,B^i}$ indicates robot $r_2$ is at location $B^i$, and $N_{\text{seq}}$ denotes the number of sequential dependencies. \jimrev{For each sequential constraint $i$, there are a pair of locations $(A^i, B^i)$ and a pair of robots $(r_1, r_2)$ such that robot $r_2$ eventually reaches location $B^i$, provided that robot $r_1$ has previously reached location $A^i$.}

\xuanrev{Unlike the previous experiment, these two specifications involve multiple robots, whereas Lemma~\ref{lemma:dnf_tight_lp} guarantees a tight LP relaxation only for the single-robot setting. The relaxation nevertheless remains tight here, because in both cases we do not define any predicate as a disjunction of edges. Instead, each predicate corresponds to a single self-loop edge $e$, so the disjunction in \eqref{eqn:predicate_dnf} degenerates to the single equivalence $r_e = 1 \Longleftrightarrow z^{\pi_k} = 1$. Fixing $z^{\pi_k}$ to a binary value therefore fixes $r_e$ to $0$ or $1$, leaving the underlying network flow structure intact and the LP relaxation tight.}


\begin{figure*}[!htbp]
    \centering
    \includegraphics[width=1\textwidth]{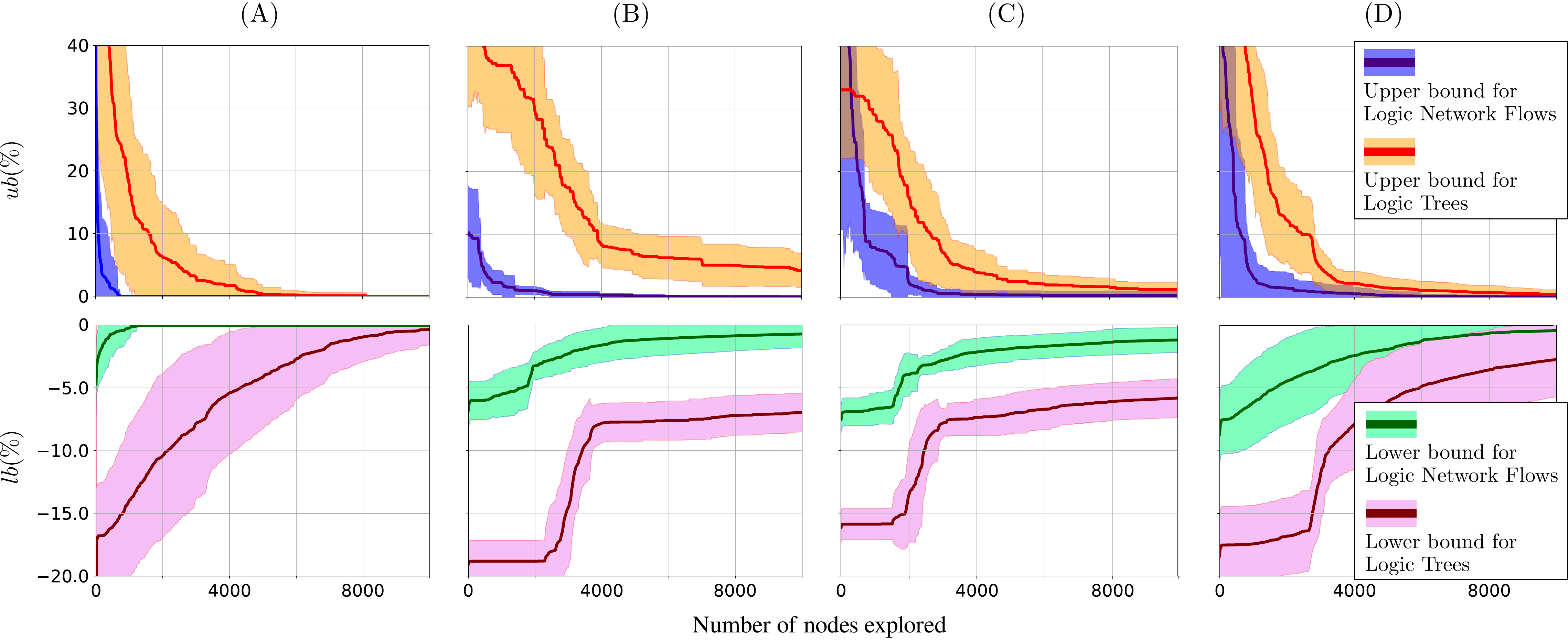}
     \caption{Solved upper and lower bounds plotted against the number of nodes explored during the B\&B process for four tasks between LNF and LT. Curves are the median and shaded regions show the variance. In general, LNF discovers the same bound by exploring less number of nodes. Four figures on the top show the upper bounds in percentage values, computed by $(\text{upper bound} - \text{global optimum})/\text{global optimum} \times 100\%$. The bottom four figures show the lower bounds in percentage values, computed by $(\text{lower bound} - \text{global optimum})/\text{global optimum} \times 100\%$. (A) shows $\varphi_{\text{vrptw}}$ with $3$ Robots, $9$ Tasks and $T=50$; (B) shows $\varphi_{\text{vrptw}}$ with $6$ Robots, $18$ Tasks and $T=70$; (C) shows $\varphi_{\text{sequential}}$ with $3$ Robots, $9$ Tasks and $T=50$; and (D) shows $\varphi_{\text{sequential}}$ with $6$ Robots, $18$ Tasks and $T=60$.}
\label{fig:task1}
\end{figure*} 

We conduct a series of experiments with increasing problem sizes. The first scenario involves $3$ robots serving $9$ tasks over a planning horizon of $T=50$ on a temporal graph containing $93$ vertices and $264$ edges. We then scale up to $6$ robots handling $18$ tasks with $T=70$ on a temporal graph containing $182$ vertices and $534$ edges. For each problem setup, $50$ randomized trials are conducted to ensure statistical significance. Task locations are randomly selected from the vertices of the graph. To model varying traversal difficulty, we assign random costs to the edges of the DNF following a uniform distribution in $[0,1]$, where higher costs represent greater traversal difficulties, such as rougher terrain or stronger wind gusts. We also introduce costs uniformly sampled from $[0, 1]$, assigned to the self-loop edges. This represents costs incurred when a robot remains stationary, such as exposure to extreme environmental temperatures. Similar to the previous experiment, we compare our LNF formulation against~\citep{kurtz2021more}.



\begin{figure}[H]
  \centering
  \includegraphics[width=\columnwidth]{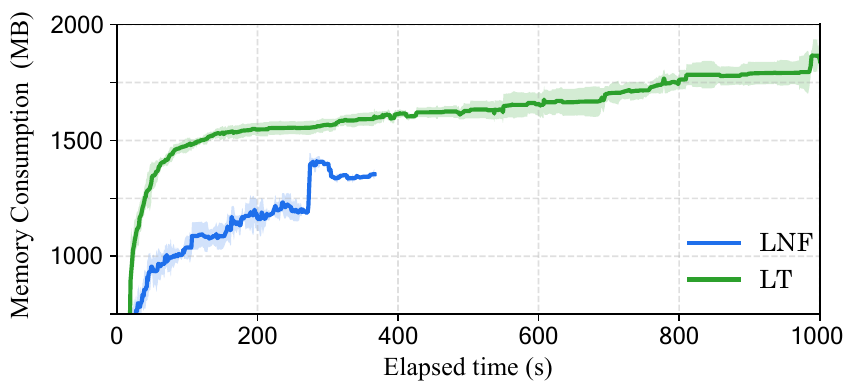}
  \caption{\xuanrev{Memory consumption versus elapsed solving time for the LNF and LT
  formulations on the VRPTW $3$-robot problem. Bold curves are
  the median across trials and shaded bands area the median absolute deviation.}}
  \label{fig:mem_vs_time}
\end{figure}

\textbf{Results} Tables~\ref{Tab:results_vrptw} and \ref{Tab:results_sequential} show the computational results for $\varphi_{\text{vrptw}}$ and $\varphi_{\text{sequential}}$ specifications, respectively,  across different problem sizes over 50 trials. The table reports the number of binary and continuous variables, and the number of constraints for each specification using LNF and LT formulations. The row ``$G_r$" records the root relaxation gap between the relaxed MICP formulations and the global optimal solution as explained in Section~\ref{sec:BB}. By analyzing the solver logs, we track several key performance metrics: ``T-Find ($10\%$ gap)" denotes the time to discover a solution within $10\%$ of optimality, ``T-Find (optimal)" reports when the global optimal solution is first found, ``T-Prove ($10\%$ gap)" indicates the time to prove a $10\%$ optimality gap, ``T-Prove (optimal)" represents the total time to prove global optimality, \xuanrev{and ``Peak Mem" reports the maximum memory consumption during solving.} Due to the exponential worst-case complexity leading to high variance in solving times, all timing results are reported as \textit{median $\pm$ median absolute deviation}. In addition, Fig.~\ref{fig:task1} shows the upper and lower bounds discovered by the Gurobi solver, plotted against the number of nodes explored during the B\&B process. These plots demonstrate the efficacy of discovering bounds through the B\&B process, regardless of the size of convex relaxations.
\xuanrev{Fig.~\ref{fig:mem_vs_time} plots the memory consumption against elapsed solving time for the LNF and LT formulations on the VRPTW $3$-robot case over $50$ trials. For both formulations the memory consumption grows as the solve progresses. The LNF curve consistently remains below that of LT and terminates substantially earlier, reflecting both its lower peak memory usage and its faster convergence.}

Our results indicate that optimization with LNF is more effective than that with LT in finding better upper and lower bounds, which is evidenced by the significant reduction in the number of nodes explored to achieve bounds of the same quality. We again attribute this to the tighter convex relaxations allowing the B\&B process to prune the search tree more efficiently and avoid unnecessary exploration of suboptimal regions. This efficacy enables LNF formulations to solve consistently faster than LT across all problem sizes, achieving tenfold speedups overall. Remarkably, for all $200$ instances tested ($50$ trials across four different planning problems), LNF demonstrates faster solving speeds than LT. \xuanrev{Beyond solving speed, LNF also consumes less peak memory than LT in every setting, with reductions ranging from $1.7\times$ to $4.3\times$.}

We also note that the network-based Fourier-Motzkin (F-M) elimination process is essential to the significant speedup. As shown in Table~\ref{Tab:results_vrptw}, the F-M elimination reduces the number of continuous variables by orders of magnitude (\textit{e.g.}, from $1,524,490$ to $48,994$ for the $6$-robot VRPTW case), while maintaining the same relaxation tightness. Our preliminary work \citep{lin2025optimization} shows that LNF formulations without F-M elimination outperform LT for problems with approximately $10,000$ continuous variables. However, LT becomes faster for larger problem scales due to its fewer continuous variables. The F-M elimination addresses this limitation by reducing the number of variables and constraints by orders of magnitude, and therefore maintaining LNF's computational advantage even for large-scale problems. Importantly, the solver tends to find high-quality solutions (within $10\%$ of optimality) much faster than proving global optimality, which may be sufficient for many practical applications requiring real-time planning.


\renewcommand{\arraystretch}{1.3}
\newcolumntype{C}[1]{>{\centering\let\newline\\\arraybackslash\hspace{0pt}}m{#1}}
\begin{table*}[t]
\setlength{\tabcolsep}{2pt}
\centering
\footnotesize
\caption{Computation results for specification $\varphi_{\text{vrptw}}$ with different problem sizes in $50$ trials (timeout limit: $10,000$ seconds)}
\label{Tab:results_vrptw}
\begin{tabular}
{@{}l|l|C{1.2cm}C{1.4cm}C{1.2cm}C{1.0cm}|C{1.2cm}C{1.4cm}C{1.2cm}C{1.0cm}@{}}
\Xhline{3\arrayrulewidth}
\multicolumn{2}{c|}{} &
\multicolumn{4}{c|}{\begin{tabular}[c]{c}$\varphi_{\text{vrptw}}$ \\ $3$ Robots, $9$ Tasks, T=$50$\end{tabular}} &
\multicolumn{4}{c}{\begin{tabular}[c]{c}$\varphi_{\text{vrptw}}$ \\ $6$ Robots, $18$ Tasks, T=$70$\end{tabular}} \\ \cline{3-10}
\multicolumn{2}{c|}{} & \multirow{2}{*}{LNF} & \multirow{2}{*}{\begin{tabular}[c]{@{}c@{}}LNF-w/o\\F-M\end{tabular}} & \multirow{2}{*}{LT} & \multirow{2}{*}{\begin{tabular}[c]{@{}c@{}}\xuanrev{Improv}\\\xuanrev{-ement}\end{tabular}} & \multirow{2}{*}{LNF} & \multirow{2}{*}{\begin{tabular}[c]{@{}c@{}}LNF-w/o\\F-M\end{tabular}} & \multirow{2}{*}{LT} & \multirow{2}{*}{\begin{tabular}[c]{@{}c@{}}\xuanrev{Improv}\\\xuanrev{-ement}\end{tabular}} \\
\multicolumn{2}{c|}{} & & & & & & & & \\ \hline
\multicolumn{2}{l|}{\# of binary vars} & $245$ & $245$ & $245$ & N.A. & $1242$ & $1242$ & $1242$ & N.A. \\ \hline
\multicolumn{2}{l|}{\# of cont. vars} & $17645$ & $242800$ & $17656$ & N.A. & $48994$ & $1524490$ & $49013$ & N.A. \\ \hline
\multicolumn{2}{l|}{\# of constr.} & $9614$ & $241264$ & $11897$ & N.A. & $26376$ & $1527738$ & $32283$ & N.A. \\ \hline
\multicolumn{2}{l|}{\multirow{3}{*}{$G_r$ (\%)}} & \textbf{$6.7$} & $6.7$ & $31.2$ & & \textbf{$8.1$} & $8.1$ & $38.0$ & \\
\multicolumn{2}{l|}{} & $\pm$ & $\pm$ & $\pm$ & N.A. & $\pm$ & $\pm$ & $\pm$ & N.A. \\
\multicolumn{2}{l|}{} & $3.1$ & $3.1$ & $4.3$ & & $2.0$ & $2.0$ & $1.7$ & \\ \hline
\multirow{15}{*}{Gurobi} & \multirow{3}{*}{T-Find ($10$\% gap)} & \textbf{$1.1$} & $266.0$ & $5.7$ & & \textbf{$49.2$} & $2580.9$ & $1178.8$ & \\
& & $\pm$ & $\pm$ & $\pm$ & \textbf{$5\times$} & $\pm$ & $\pm$ & $\pm$ & \textbf{$24\times$} \\
& & $0.4$ & $89.5$ & $2.3$ & & $5.3$ & $1161.4$ & $244.9$ & \\ \cline{2-10}
& \multirow{3}{*}{T-Find (optimal)} & \textbf{$1.8$} & $643.8$ & $22.6$ & & \textbf{$570.0$} & \multirow{3}{*}{\textit{Timeout}} & \multirow{3}{*}{\textit{Timeout}} & \\
& & $\pm$ & $\pm$ & $\pm$ & \textbf{$12\times$} & $\pm$ & & & \textbf{$>$$18\times$} \\
& & $0.6$ & $329.1$ & $14.9$ & & $245.9$ & & & \\ \cline{2-10}
& \multirow{3}{*}{T-Prove ($10$\% gap)} & \textbf{$1.2$} & $337.7$ & $22.4$ & & \textbf{$49.4$} & $3675.0$ & $1990.6$ & \\
& & $\pm$ & $\pm$ & $\pm$ & \textbf{$19\times$} & $\pm$ & $\pm$ & $\pm$ & \textbf{$40\times$} \\
& & $0.4$ & $117.3$ & $14.5$ & & $6.5$ & $893.3$ & $810.0$ & \\ \cline{2-10}
& \multirow{3}{*}{T-Prove (optimal)} & \textbf{$1.7$} & $1202.6$ & $52.9$ & & \textbf{$779.2$} & \multirow{3}{*}{\textit{Timeout}} & \multirow{3}{*}{\textit{Timeout}} & \\
& & $\pm$ & $\pm$ & $\pm$ & \textbf{$31\times$} & $\pm$ & & & \textbf{$>$$13\times$} \\
& & $0.8$ & $849.9$ & $34.6$ & & $409.8$ & & & \\
\cline{2-10}
& \multirow{3}{*}{\xuanrev{Peak Mem (MB)}} & \xuanrev{\textbf{$974$}} & \xuanrev{$4385$} & \xuanrev{$1762$} & & \xuanrev{\textbf{$2686$}} & \xuanrev{$12092$} & \xuanrev{$4498$} & \\
& & $\pm$ & $\pm$ & $\pm$ & \xuanrev{\textbf{$1.8\times$}} & $\pm$ & $\pm$ & $\pm$ & \xuanrev{\textbf{$1.7\times$}} \\
& & \xuanrev{$100$} & \xuanrev{$293$} & \xuanrev{$122$} & & \xuanrev{$103$} & \xuanrev{$586$} & \xuanrev{$585$} & \\
\hline
\multirow{3}{*}{CPLEX} & \multirow{3}{*}{T-Prove (optimal)} & \textbf{$13.0$} & \multirow{3}{*}{\textit{Timeout}} & $4475.7$ & & \textbf{$1127.0$} & \multirow{3}{*}{\textit{Timeout}} & \multirow{3}{*}{\textit{Timeout}} & \\
& & $\pm$ & & $\pm$ & \textbf{$344\times$} & $\pm$ & &  & \textbf{$>$$9\times$} \\
& & $9.6$ & & $3501.7$ & & $566.0$ & & & \\ \Xhline{3\arrayrulewidth}
\end{tabular}
\end{table*}

\renewcommand{\arraystretch}{1.3}
\newcolumntype{C}[1]{>{\centering\let\newline\\\arraybackslash\hspace{0pt}}m{#1}}
\begin{table*}[t]
\setlength{\tabcolsep}{2pt}
\centering
\footnotesize
\caption{Computation results for specification $\varphi_{\text{sequential}}$ with different problem sizes in $50$ trials (timeout limit: $10,000$ seconds)}
\label{Tab:results_sequential}
\begin{tabular}{@{}l|l|C{1.2cm}C{1.4cm}C{1.2cm}C{1.0cm}|C{1.2cm}C{1.4cm}C{1.2cm}C{1.0cm}@{}}
\Xhline{3\arrayrulewidth}
\multicolumn{2}{c|}{} &
\multicolumn{4}{c|}{\begin{tabular}[c]{c}$\varphi_{\text{sequential}}$ \\ $3$ Robots, $9$ Tasks, T=$50$\end{tabular}} &
\multicolumn{4}{c}{\begin{tabular}[c]{c}$\varphi_{\text{sequential}}$ \\ $6$ Robots, $18$ Tasks, T=$60$\end{tabular}} \\ \cline{3-10}
\multicolumn{2}{c|}{} & \multirow{2}{*}{LNF} & \multirow{2}{*}{\begin{tabular}[c]{@{}c@{}}LNF-w/o\\F-M\end{tabular}} & \multirow{2}{*}{LT} & \multirow{2}{*}{\begin{tabular}[c]{@{}c@{}}\xuanrev{Improv}\\\xuanrev{-ement}\end{tabular}} & \multirow{2}{*}{LNF} & \multirow{2}{*}{\begin{tabular}[c]{@{}c@{}}LNF-w/o\\F-M\end{tabular}} & \multirow{2}{*}{LT} & \multirow{2}{*}{\begin{tabular}[c]{@{}c@{}}\xuanrev{Improv}\\\xuanrev{-ement}\end{tabular}} \\
\multicolumn{2}{c|}{} & & & & & & & & \\ \hline
\multicolumn{2}{l|}{\# of binary vars} & $475$ & $475$ & $475$ & N.A. & $1035$ & $1035$ & $1035$ & N.A. \\ \hline
\multicolumn{2}{l|}{\# of cont. vars} & $16903$ & $235403$ & $16914$ & N.A. & $41967$ & $1085247$ & $41986$ & N.A. \\ \hline
\multicolumn{2}{l|}{\# of constr.} & $9853$ & $243678$ & $21361$ & N.A. & $23637$ & $1113762$ & $53889$ & N.A. \\ \hline
\multicolumn{2}{l|}{\multirow{3}{*}{$G_r$ (\%)}} & \textbf{$16.7$} & $16.7$ & $37.2$ & & \textbf{$18.4$} & $18.4$ & $40.1$ & \\
\multicolumn{2}{l|}{} & $\pm$ & $\pm$ & $\pm$ & N.A. & $\pm$ & $\pm$ & $\pm$ & N.A. \\
\multicolumn{2}{l|}{} & $2.7$ & $2.7$ & $3.7$ & & $2.0$ & $2.0$ & $1.9$ & \\ \hline
\multirow{15}{*}{Gurobi} & \multirow{3}{*}{T-Find ($10$\% gap)} & \textbf{$10.1$} & $320.8$ & $59.8$ & & \textbf{$62.2$} & $418.6$ & $600.5$ & \\
& & $\pm$ & $\pm$ & $\pm$ & \textbf{$6\times$} & $\pm$ & $\pm$ & $\pm$ & \textbf{$10\times$} \\
& & $4.3$ & $51.9$ & $30.3$ & & $16.7$ & $177.5$ & $314.7$ & \\ \cline{2-10}
& \multirow{3}{*}{T-Find (optimal)} & \textbf{$32.6$} & $1345.6$ & $339.7$ & & \textbf{$432.5$} & \multirow{3}{*}{\textit{Timeout}} & \multirow{3}{*}{\textit{Timeout}} & \\
& & $\pm$ & $\pm$ & $\pm$ & \textbf{$10\times$} & $\pm$ & & & \textbf{$>23\times$} \\
& & $20.0$ & $626.5$ & $158.6$ & & $127.0$ & & & \\ \cline{2-10}
& \multirow{3}{*}{T-Prove ($10$\% gap)} & \textbf{$14.1$} & $893.7$ & $226.3$ & & \textbf{$195.8$} & $1970.3$ & $1833.9$ & \\
& & $\pm$ & $\pm$ & $\pm$ & \textbf{$16\times$} & $\pm$ & $\pm$ & $\pm$ & \textbf{$9\times$} \\
& & $6.8$ & $387.4$ & $59.1$ & & $80.2$ & $29.7$ & $728.2$ & \\ \cline{2-10}
& \multirow{3}{*}{T-Prove (optimal)} & \textbf{$55.5$} & $2694.0$ & $437.2$ & & \textbf{$1664.2$} & \multirow{3}{*}{\textit{Timeout}} & \multirow{3}{*}{\textit{Timeout}} & \\
& & $\pm$ & $\pm$ & $\pm$ & \textbf{$8\times$} & $\pm$ & & & \textbf{$>6\times$} \\
& & $38.5$ & $1017.9$ & $234.7$ & & $1105.7$ & & & \\
\cline{2-10}
& \multirow{3}{*}{\xuanrev{Peak Mem (MB)}} & \xuanrev{\textbf{$1152$}} & \xuanrev{$8140$} & \xuanrev{$2046$} & & \xuanrev{\textbf{$3303$}} & \xuanrev{$23338$} & \xuanrev{$14131$} & \\
& & $\pm$ & $\pm$ & $\pm$ & \xuanrev{\textbf{$1.8\times$}} & $\pm$ & $\pm$ & $\pm$ & \xuanrev{\textbf{$4.3\times$}} \\
& & \xuanrev{$123$} & \xuanrev{$887$} & \xuanrev{$95$} & & \xuanrev{$312$} & \xuanrev{$1496$} & \xuanrev{$73$} & \\
\hline
\multirow{3}{*}{CPLEX} & \multirow{3}{*}{T-Prove (optimal)} & \textbf{$42.1$} & \multirow{3}{*}{\textit{Timeout}} & $595.7$ & & \textbf{$5837.0$} & \multirow{3}{*}{\textit{Timeout}} & \multirow{3}{*}{\textit{Timeout}} & \\
& & $\pm$ & & $\pm$ & \textbf{$14\times$} & $\pm$ & & & \textbf{$>2\times$} \\
& & $17.6$ & & $276.7$ & & $2553.0$ & & & \\ \Xhline{3\arrayrulewidth}
\end{tabular}
\end{table*}

\subsection{Path Planning with Trajectory Libraries: A Search and Rescue Application}
\label{Sec:search_rescue}
We showcase a practical application of our framework combining LNF and DNF with a multi-robot search and rescue assignment, where three bipedal robots coordinate to complete multiple missions in a disastrous environment. We demonstrate that our LNF formulation of temporal logic specifications discovers optimal coordination plans significantly faster than the baseline LT approach, enabling rapid mission planning for time-critical applications such as disaster response.

\begin{figure*}[hbt!]
    \centering
    \includegraphics[width=1.0\textwidth]{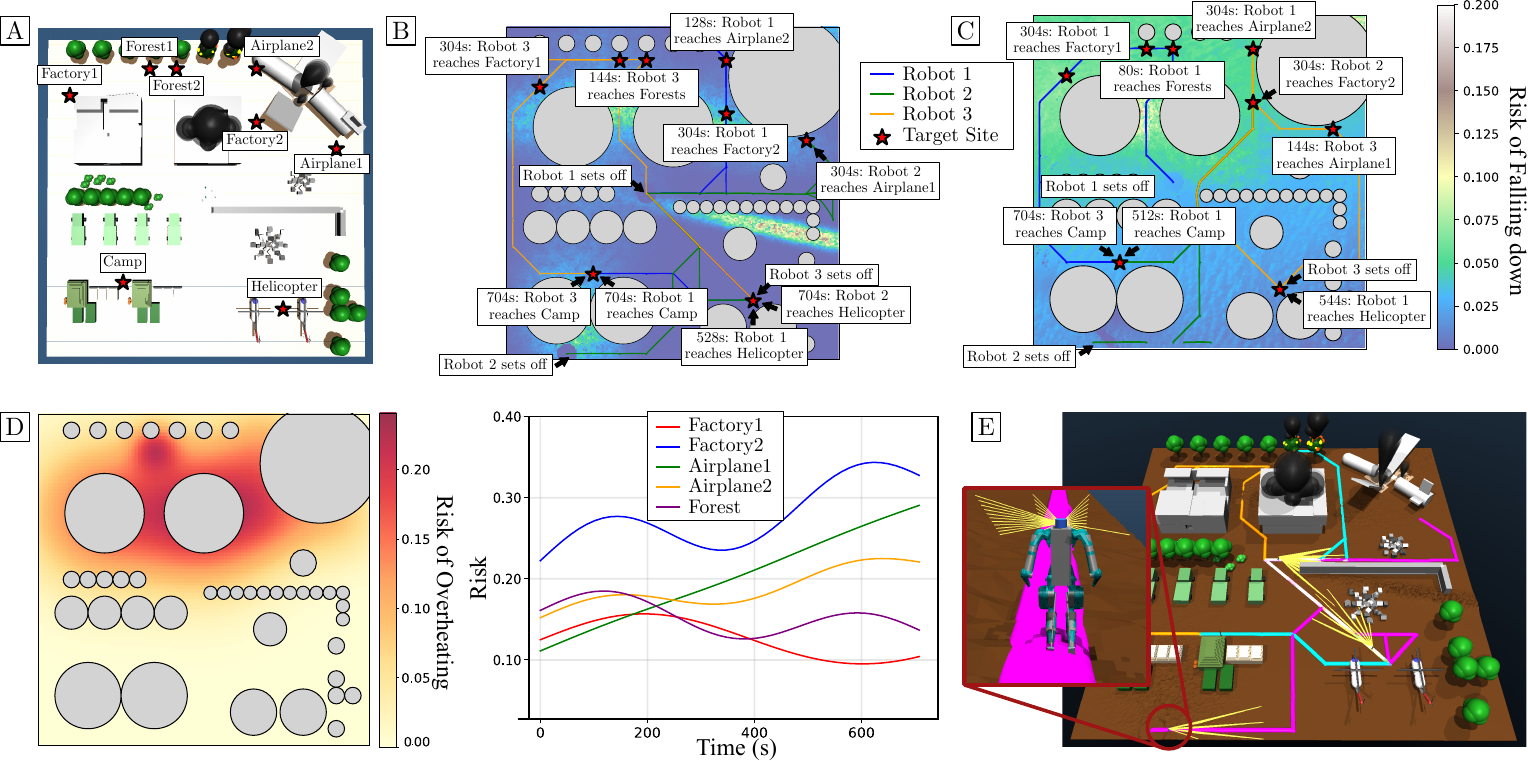}
     \caption{Search and rescue mission planning with three bipedal robots in a disaster environment. (A) Overview of the $100 \times 100$~m disaster zone containing mission targets: crashed airplane at ($40$~m, $16$~m), factory houses on fire at ($-40$~m, $32$~m) and ($16$~m, $24$~m), trees on fire at ($-8$~m, $40$~m) and ($-16$~m, $40$~m), search crew base camp at ($-24$~m, $-24$~m), and helicopter landing zone at ($24$ m, $-32$ m). Red stars indicate target locations, with walls, debris, and trees serving as obstacles. (B and C) Optimized robot trajectories for two terrain profiles with distinct roughness patterns. Color intensity represents risk of falling due to terrain roughness. Three robots start from ($-8$~m, $0$~m), ($-24$~m, $-24$~m), and ($24$~m, $-32$~m) respectively, with arrival times labeled at each target. (D) Spatial distribution of overheating risk across the disaster zone at $t = 0$. The accompanying plot on the right shows the time-varying overheating risk profiles for different locations: \textit{Factory1}, \textit{Factory2}, \textit{Airplane1}, \textit{Airplane2}, \textit{Forest1}, and \textit{Forest2}. (E) MuJoCo simulation of bipedal robot Digit executing planned trajectory using ALIP-based controller. Planned paths are shown in light blue, orange, and magenta.}
\label{fig:map}
\end{figure*}

The experiment runs on a $100$~m $\times$ $100$~m disaster zone covering multiple mission sites including crashed airplanes, factory plants or woods on fire, a search crew base camp, a helicopter landing zone, and multiple obstacles, as illustrated in Fig.~\ref{fig:map} (A). The mission objective is to coordinate three bipedal robots to execute: (1) search and rescue operations by reaching two aircraft crash sites to rescue survivors and transport them to the helicopter for medical evacuation, (2) retrieval of critical equipment from two damaged factory plants back to the base camp, and (3) wildfire extinguishing tasks, all within specified time windows. Let $\pi^{i,j}$ denote the predicate that the $i$-th robot, $i \in \{1,2,3\}$, is at location $j$, where the locations include: $p_1$ (\textit{\textit{Factory1}}), $p_2$ (\textit{Factory2}), $p_3$ (\textit{Airplane1}), $p_4$ (\textit{Airplane2}), $p_5$ (\textit{Forest1}), $p_6$ (\textit{Forest2}), $p_7$ (\textit{Camp}), and $p_8$ (\textit{Helicopter}). The task specification for the overall mission is designed as:
\begin{align*}
\phi_{\rm{mission}} = \bigwedge_{j=1}^{5} \phi_{\rm{mission}(j)} \nonumber
\end{align*}
where the factory and airplane rescue missions require one robot visiting the task locations, staying there for a specified duration to complete the operations, and then transporting survivors or retrieving equipment back to the designated search crew base within the time window $T_w$:
\begin{align*}
&\phi_{\rm{mission}(j)} = \\
&\bigvee_{i=1}^{3} \left(\Diamond_{[0,T]}\left( \square_{[0,4]} \pi^{i,p_j} \land \Diamond_{[0, T_w]}\square_{[0,4]} \pi^{i,p_{\rm{dest}(j)}} \right)\right)
\end{align*}
where $j \in \{1, 2, 3, 4\}$ corresponds to $p_{\rm{dest}(1)}=p_6$, $p_{\rm{dest}(2)}=p_6$, $p_{\rm{dest}(3)}=p_7$, $p_{\rm{dest}(4)}=p_7$, respectively.
%
The fire extinguishing mission requires the robot to visit the forest areas and stay for a period of time to extinguish the fire:
\begin{align*}
&\phi_{\rm{mission5}} = \\
&\left(\bigvee_{i=1}^{3} \Diamond_{[0,T]} \square_{[0,4]} \pi^{i,p_4}\right) \land \left(\bigvee_{i=1}^{3} \Diamond_{[0,T]} \square_{[0,4]} \pi^{i,p_5}\right)
\end{align*}

To construct a DNF for this environment, the map is discretized into a $10\times10$ grid, excluding grid points that overlap with obstacles. The remaining points serve as vertices in the temporal graph. We build an offline-generated trajectory library to connect these vertices on the map. For each grid vertex, \xuan{nonlinear trajectory optimizations are employed to find collision-free paths connecting a vertex to all reachable neighboring vertices. Specifically, we optimize dynamically feasible footstep sequences based on the Linear Inverted Pendulum (LIP) model \citep{koolen2012capturability}. Each obstacle is conservatively approximated as a circular region that circumscribes it, and obstacle avoidance is enforced by constraining the robot's CoM trajectory to remain outside these circle regions.} We select a time discretization of $dT = 16$ seconds to ensure reasonable edge traversal time steps and maintain a manageable planning horizon $T = 50$ and $T_w = 25$. Since each robot undertakes independent mission plans, we construct $3$ separate DNFs to find a path for each robot. Each DNF contains $87$ vertices and $470$ edges per time step. The full optimization formulation includes $726$ binary variables, $81,456$ continuous variables, and $26,964$ constraints.

The objective function is designed to incorporate two types of risk-aware locomotion costs: one models the risk of falls, which is associated with terrain roughness, and the other is related to time-varying overheating risk that measures the potential for mechanical or electronic failure due to severe fire intensity. The first cost is imposed when the robot moves from one site to the next, while the second is imposed regardless of the robot's motion status (\textit{i.e.}, either in motion or in a standing pose). To model the terrain roughness, we leverage multi-modal Gaussian distributions randomly placed on the map that range from zero to the maximum roughness level. Fig.~\ref{fig:map} (B) and (C) show two roughness distribution examples. 

To account for the first type of costs, roughness values are queried along the robot's trajectory path. The probability of failure is assumed to be scaled proportionally to the terrain roughness, and then integrated along the path. We then convert it to log-probability space, making the risks additive across trajectory segments. The terrain cost for each edge is computed as the negative log-probability of successful traversal:
\begin{equation}
\theta_{\rm{traversal}}(e) = -\log(1 - p_{\rm{fail}}(e)) \nonumber
\end{equation}
where $p_{\rm{fail}}(e)$ represents the probability of failure for the traverse of the edge $e$ based on the roughness of the terrain.

To measure the second type of risk, we model disaster sources centered at each fire location, with severity decreasing exponentially with distance. These costs vary over time, following evolution patterns as shown in Fig.~\ref{fig:map} (D). \jimrev{For example, aircraft crash sites exhibit increasing fire intensity, whereas forest fires exhibit diminishing intensity over time.} Fig.~\ref{fig:map} (D) also shows the spatial distribution of the risk of overheating at the beginning. The cost function sums the severity contributions of all disaster sources at each vertex and time step. The probability of failure is proportional to the severity with an appropriate scaling factor. The cost of staying at vertex $v$ and time step $k$ is: 
\begin{equation}
\theta_{\rm{hold}}(v,k) = -\log(1 - p_{\rm{fail}}(v,k)) \nonumber
\end{equation}
where $p_{\rm{fail}}(v,k)$ is the probability of failure due to overheating at vertex $v$ and time step $k$.

\textbf{Results} 
We compare LNF and LT formulations solved by Gurobi 12.0 on two problem instances with different terrain roughness patterns. The solved trajectories are shown in Fig.~\ref{fig:map} (B) and (C). As expected, both formulations find the same globally optimal solution, with their key distinction in computational efficiency. In the first terrain configuration, the optimal solution assigns Robot $1$ to handle the forest fire missions first, then to retrieve equipment from the first factory house, and finally to return to the base camp. Robot $3$ sequentially investigates both crash sites to rescue survivors and transports them to the helicopter landing zone. While \textit{Factory2} is located near the aircraft crash sites, assigning all nearby tasks to Robot $3$ would exceed the time constraints. Therefore, Robot $2$ is tasked with the second factory house rescue mission. Time spent near \textit{Factory1} is minimized because of its high terrain roughness.

In contrast, the optimal task assignment and paths for the second terrain are adjusted to avoid the highly rough terrain region in the middle-right of the map. As a result, Robot $3$, which was previously assigned airplane rescue tasks, now takes over the forest fire missions and factory house operations. This shift causes Robot $1$ to be reassigned to airplane rescue operations and the \textit{Factory2} mission. Since the disaster severity grows at the airplane crash sites, the robots prioritize these missions to avoid prolonged exposure to escalating risks. On the other hand, because the factories demonstrate periodic risk patterns, the robots intelligently avoid high-risk areas by waiting until the danger subsides before accessing those regions. 

The computational results again reveal a significant performance advantage of LNF over LT. For the first terrain configuration, LNF achieves $4\times$ speedups in both global optimality ($165$~s vs $632$~s) and $10$\%-optimal solutions ($39$~s vs $165$~s). For the second terrain configuration, LNF finds the global optimum in $69$ seconds while LT exceeds $3000$ seconds (over $40\times$ faster), and achieves $5\times$ speedup for $10$\%-optimal solutions ($44$~s vs $213$~s). These results highlight that LNF's tighter convex relaxations ($58\%$ and $68\%$, versus $90\%$ and $91\%$ for LT) enable the solver to find lower-risk robot paths within a shorter time.

Additionally, we simulate the search and rescue scenario in a MuJoCo environment (see the attached video submission). The bipedal robots use controllers based on the Angular Momentum Linear Inverted Pendulum model \citep{gong2022zero} to execute the planned paths, as shown in Fig.~\ref{fig:map} (E). This implementation exhibits that the optimized trajectories are realistically tractable for  bipedal systems to navigate complex environments while maintaining balance and satisfying all temporal logic specifications.

\subsection{Optimal Motion Planning with \xuanrev{PWA Dynamics} using Point-Mass Model}
\label{sec:point_mass_navigation}

\begin{figure*}[h!]
    \centering
    \includegraphics[width=0.8\textwidth]{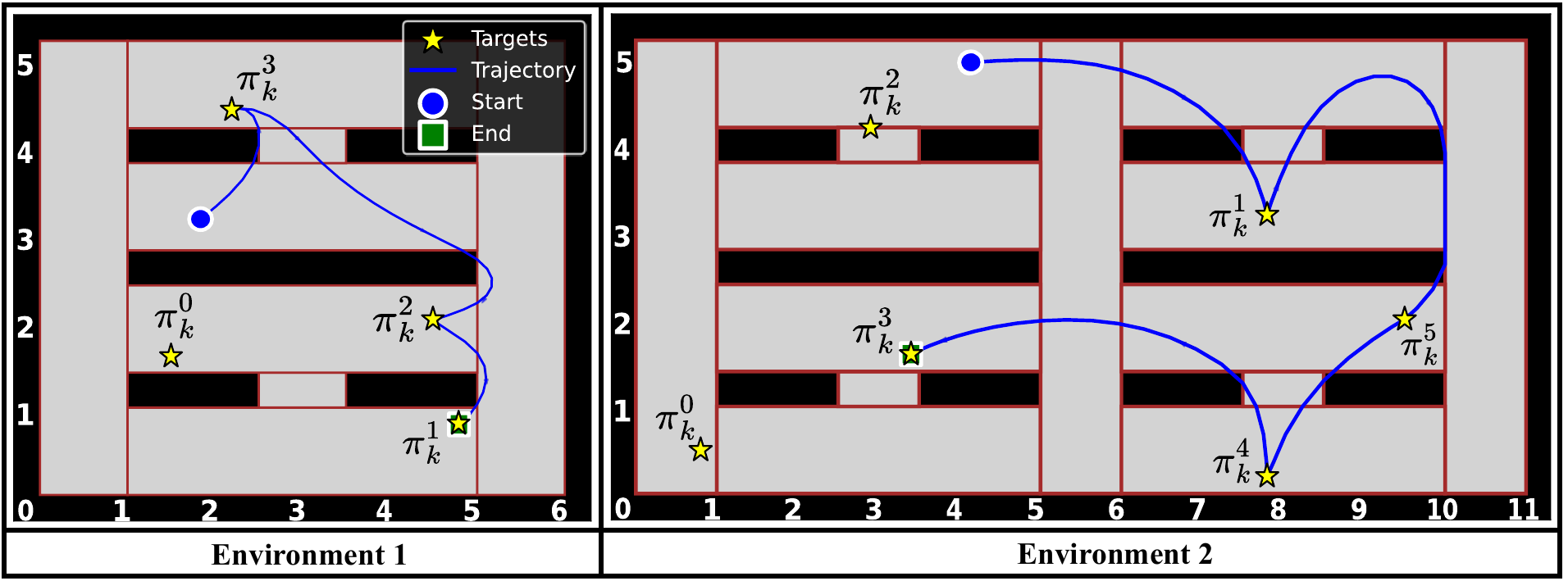}
    \caption{Point-mass navigation environments used for evaluating hybrid dynamics under STL specifications.  Gray regions represent collision-free areas, yellow stars indicate target points, and the blue trajectory shows the optimized path from start (blue circle) to end (green square) position satisfying the sequential dependency constraints. 
    }
    \label{fig:point_mass_envs}
\end{figure*}

\xuan{We evaluate our LNF formulation on motion planning with PWA dynamics under STL specifications. In this section, we feature a simple point-mass model as the underlying dynamical system,} where the system state is $\boldsymbol{x}_k = [x_k, y_k, \dot{x}_k, \dot{y}_k]^\top \in \mathbb{R}^4$ representing the 2D position and velocity of the robot. The control input $\boldsymbol{u}_k = [u_{x,k}, u_{y,k}]^\top \in \mathbb{R}^2$ represents the input forces in the $x$ and $y$ directions. We design two environments of increasing complexity: the first environment is a $6$~m $\times$ $5$~m workspace which is decomposed into $8$ collision-free convex regions, and the second environment is an $11$~m $\times$ $5$~m workspace divided into $15$ collision-free convex regions, as shown in Fig.~\ref{fig:point_mass_envs}. We adopt the big-M formulation \eqref{eqn:pwa_as_bigM} for collision-avoidance constraints and introduce binary variables $\delta^i_k \in \{0,1\}$ to indicate whether the robot occupies region $i$ at time step $k$. This guarantees that the robot’s position at each time step lies within exactly one convex region. 

The STL specifications involve sequential dependency constraints with multiple target pairs. For Environment $1$, we randomly sample $4$ targets from different convex regions and create $2$ sequential pairs, while environment $2$ is sampled $6$ targets with $3$ sequential pairs. The specification for Environment $1$ is:
\begin{equation}
\begin{aligned}
    \phi_{\text{M1}} = &(\neg \pi^0 \, \mathcal{U} \, \pi^1) \wedge (\neg \pi^2 \, \mathcal{U} \, \pi^3) \wedge \Diamond_{[0,T]} \pi^1 \wedge \Diamond_{[0,T]} \pi^3 \wedge \\
    & (\Diamond_{[0,T]} \square_{[0,3]} \pi^0 \vee \Diamond_{[0,T]} \square_{[0,3]} \pi^2)
\end{aligned}
\end{equation}
which requires that target $1$ must be visited before target $0$, and target $3$ before target $2$. The specification further requires the robot to visit all prerequisite targets ($\pi^1$ and $\pi^3$) and at least one of the subsequent targets ($\pi^0$ or $\pi^2$) must eventually be visited and occupied for $3$ time steps. Environment $2$ extends this pattern with 
\begin{equation}
\begin{aligned}
\phi_{\text{M2}} = &(\neg \pi^0 \, \mathcal{U} \, \pi^1) \wedge (\neg \pi^2 \, \mathcal{U} \, \pi^3) \wedge (\neg \pi^4 \, \mathcal{U} \, \pi^5) \wedge \\
& \Diamond_{[0,T]} \pi^1 \wedge \Diamond_{[0,T]} \pi^3 \wedge \Diamond_{[0,T]} \pi^5 \wedge \\
& (\Diamond_{[0,T]} \square_{[0,3]} \pi^0 \vee \Diamond_{[0,T]} \square_{[0,3]} \pi^2 \vee \Diamond_{[0,T]} \square_{[0,3]} \pi^4)
\end{aligned}
\end{equation}
by adding a third sequential pair while maintaining the disjunctive choice between subsequent targets. We use a discretization time step of $dT = 0.5$ seconds with planning horizons of $T = 60$ time steps for both environments.

To ensure fair comparisons, we execute $50$ trials for each environment, with initial conditions randomly sampled on the map but within a specified convex region. We also build a target visit cost $\boldsymbol{\Theta}$, incurring the cost of visiting each target, uniformly sampled between $0$ and $1$, \xuan{modeling preferences for specific task completion locations and times due to factors such as lower operational risk.} The \xuan{state} cost matrix is $\boldsymbol{Q}=\boldsymbol{0}$ and \xuan{the control cost matrix is} $\boldsymbol{R} = \text{diag}([1, 1])$ to penalize the control input. We compare our LNF formulation with the baseline LT formulation using the big-M method, which aligns with optimization-based formulations for temporal logic control, as presented in Eqn. 15 in~\citep{belta2019formal}.

\textbf{Results}
The results for point-mass navigation in two environments are shown in the first ($\phi_{\text{M1}}$) and second ($\phi_{\text{M2}}$) columns of Table~\ref{Tab:dynamics_results}. Similar to previous experiments, we report the number of binary and continuous variables, constraints, root relaxation gap, \xuanrev{peak memory consumption}, and various timing metrics. In addition, we present speed summary results that indicate the number of trials (out of $50$) where LNF is faster, LT is faster, or the performance is similar (within a $20\%$ difference).

For the point-mass environments, LNF outperforms LT across both problem sizes. In Environment 1 ($\phi_{\text{M1}}$), LNF achieves better root relaxation gaps ($36.5\%$ vs $41.0\%$) and demonstrates clear computational advantages, with LNF being faster in $44$ of $50$ trials compared to only two trials where LT is faster. The speedups range from $2.5 \times$ to $3.0 \times$ across different timing metrics. Environment 2 ($\phi_{\text{M2}}$) shows a similar pattern, but with slightly reduced advantages as problem complexity increases: LNF is faster in $36$ of $50$ trials versus $4$ for LT, with $10$ trials showing a similar computational speed. The speedups remain substantial at $1.7 \times$ to $3.3 \times$, though LT slightly outperforms LNF in finding the optimal solution ($0.9 \times$ speedup for T-Find optimal). \xuanrev{LNF also requires less peak memory in both environments, with reductions of $1.8 \times$ for $\phi_{\text{M1}}$, and $3.1 \times$ for $\phi_{\text{M2}}$.}

\subsection{Optimal Motion Planning with \xuanrev{PWA Dynamics} using Inverted Pendulum Model} \label{sec:LIP_experiment}

\begin{figure*}[h!]
    \centering
    \includegraphics[width=0.95\textwidth]{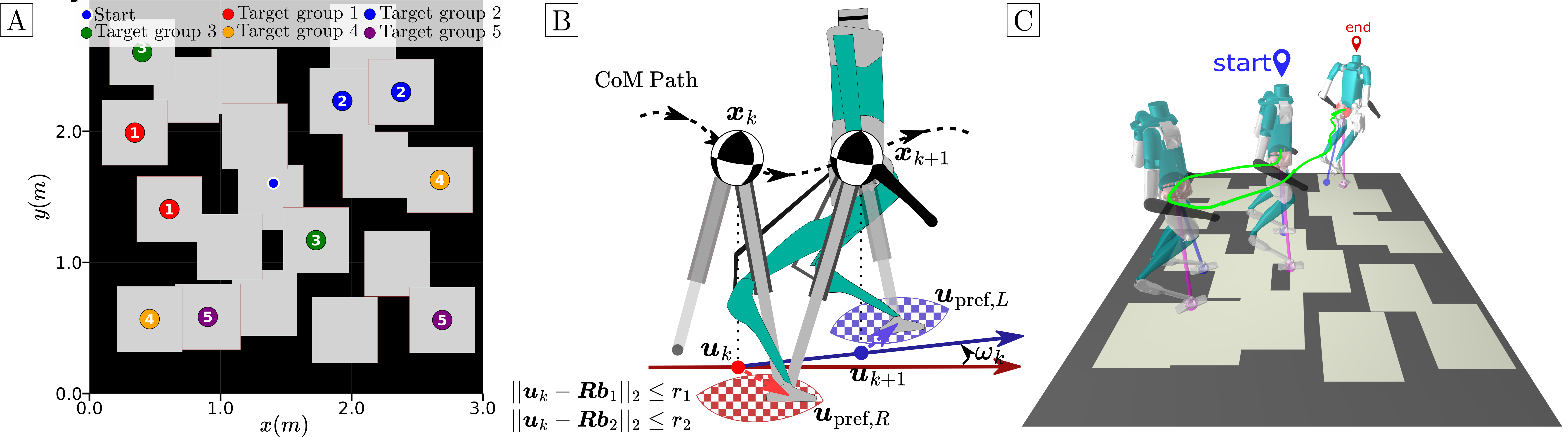}
    \caption{Bipedal locomotion planning with LIP dynamics under STL  specifications. (A) Environment setup: The $3$ m $\times$ $3$ m workspace is decomposed into $20$ safe contact regions ($0.5$ m $\times$ $0.5$ m squares with overlap, shown in light gray). Five target groups are defined, each containing $2$ regions labeled with matching colors and numbers 1-5. The STL  specification requires the robot to visit at least one target region from each group. (B) Kinematic feasibility constraints: Footstep placement is constrained by the intersection of two circles representing the reachable regions for left and right feet, with preferred footstep positions $\boldsymbol{u}_{\text{pref},(L/R)}$ indicated. (C) Planning results: The optimized CoM trajectory is shown as a green curve, with the bipedal walking gait illustrated using LIP models superimposed along the trajectory. Each LIP model shows the CoM position (center) and corresponding left and right foot contact locations. 
    }
    \label{fig:LIP_envs}
\end{figure*}

To further evaluate more realistic robot models, we study bipedal footstep planning using a 3-D LIP model \xuan{\citep{deits2014footstep}. While existing LIP-based approaches such as \citep{deits2014footstep} typically plan optimal footsteps toward predefined goal states, our framework generates footstep sequences to reach several locations of interest under temporal logic specifications.} The system state $\boldsymbol{x}_k = [x, y, \dot{x}, \dot{y}, \theta]^\top \in \mathbb{R}^5$ includes the CoM position, velocity, and heading angle. The control input $\boldsymbol{u}_k = [u_x, u_y, \omega]^\top \in \mathbb{R}^3$ represents the footstep placement relative to the CoM in the $x$ and $y$ directions, and the angular velocity for the heading angle change. We use the discrete time LIP dynamics developed by \citep{narkhede2022sequential}, \xuan{where the dynamics represent walking-step-to-walking-step evolution of the robot's CoM:}

\begin{subequations}
For $i \in \{x, y\}$:
\begin{align}
\begin{bmatrix} s_{k+1}^{i} \\ \dot{s}_{k+1}^{i} \end{bmatrix} &= \begin{bmatrix} 1 & \frac{1}{\omega} \sinh(\omega dT) \\ 0 & \cosh(\omega dT)\end{bmatrix} \begin{bmatrix} s_{k}^{i} \\ \dot{s}_k^{i} \end{bmatrix} + \begin{bmatrix} 1 - \cosh(\omega dT)\\ -\omega \sinh(\omega dT) \end{bmatrix} u_{k}^{i} \\
\theta_{k+1} &= \theta_k + dT\omega_k
\end{align}
\label{eqn:LIP}\\
\end{subequations}
%
\noindent \xuan{The discrete transition from the $k$-th step to the $(k+1)$-th step represents a contact transition, where $H$ is the CoM height, $\omega = \sqrt{g/H}$ is the natural frequency of the pendulum, and $dT$ is the time interval between consecutive footsteps.} We introduce additional binary variables to discretize the heading angle into $n_\theta=4$ segments, and approximate the sinusoidal functions using piecewise linear functions  \citep{deits2014footstep}:
\begin{subequations}
\begin{align}
\sum_{q=1}^{n_\theta} \lambda_{k}^{q} &= 1, \quad \lambda_{k}^{q} \in \{0,1\} \\
\theta_{\min}^{q} - M(1-\lambda_{k}^{q}) &\leq \theta_k \leq \theta_{\max}^{q} + M(1-\lambda_{k}^{q}) \\
|s_k - (m_{s}^{q} \theta_k + b_{s}^{q})| &\leq M(1-\lambda_{k}^{q}) \\
|c_k - (m_{c}^{q} \theta_k + b_{c}^{q})| &\leq M(1-\lambda_{k}^{q})
\end{align}
\label{eqn:piecewise_trig}\\
\end{subequations}
\noindent where $s_k = \sin(\theta_k)$, $c_k = \cos(\theta_k)$, and $\lambda_{k}^{q}$ indicates if the segment $q$ is active at time step $k$. The interval $[\theta_{\min}^{q}, \theta_{\max}^{q}]$ defines the angular range of segment $q$, while $(m_{s}^{q}, b_{s}^{q})$, $(m_{c}^{q}, b_{c}^{q})$ are the linear approximation coefficients of the sine and cosine functions, respectively, within that segment.

\xuan{For each foot, two circular bounds $\boldsymbol{b}_{1}$ and $\boldsymbol{b}_{2}$, with radii $r_1$ and $r_2$, are used to delineate the feasible footstep region. The contact location $\boldsymbol{u}_k$ must lie within the intersection of these two circles:}
\begin{equation}
\|\boldsymbol{u}_k - \boldsymbol{R}(\theta_k) \boldsymbol{b}_{1}\|_2 \leq r_1 \quad \text{and} \quad \|\boldsymbol{u}_k - \boldsymbol{R}(\theta_k) \boldsymbol{b}_{2}\|_2 \leq r_2
\label{eqn:kinematics}
\end{equation}

\noindent To ensure locomotion safety, we define $20$ discrete foot contact regions within which the robot can safely place its footsteps. Regions outside these designated areas are considered unsafe due to factors such as slippery surfaces, excessively rough terrain, or unsteppable objects. These safe regions are uniformly distributed within a $3$~m $\times$ $3$~m workspace with some overlap. \xuan{The safe contact regions are represented by the light-gray boxes in Fig.~\ref{fig:LIP_envs} (A).} We enforce the restriction that each footstep must be placed within one of these predefined safe regions.

The PWA dynamics formulation for safe bipedal locomotion incorporates several types of constraints: LIP dynamics \eqref{eqn:LIP}, kinematics feasibility \eqref{eqn:kinematics}, discretized orientation \eqref{eqn:piecewise_trig}, and safe contact regions that ensure foot placement occurs within predefined safe areas using big-M formulations.

\renewcommand{\arraystretch}{1.2}
\newcolumntype{C}[1]{>{\centering\let\newline\\\arraybackslash\hspace{0pt}}m{#1}}
\begin{table*}[h!]
\setlength{\tabcolsep}{2pt}
\centering
\caption{Computation results for PWA with temporal logic specifications in $50$ trials (timeout limit: $10{,}000$ seconds)}
\label{Tab:dynamics_results}
\footnotesize
\begin{tabular}{@{}l|l|C{1.2cm}C{1.0cm}C{0.9cm}|C{1.2cm}C{1.0cm}C{0.9cm}|C{1.2cm}C{1.0cm}C{0.9cm}|C{1.2cm}C{1.0cm}C{0.9cm}@{}}
\Xhline{3\arrayrulewidth}
\multicolumn{2}{c|}{} &
\multicolumn{3}{c|}{\begin{tabular}[c]{c}$\phi_{\text{M1}}$ \\ T=$60$, $4$ targets\end{tabular}} &
\multicolumn{3}{c|}{\begin{tabular}[c]{c}$\phi_{\text{M2}}$ \\ T=$60$, $6$ targets\end{tabular}} &
\multicolumn{3}{c|}{\begin{tabular}[c]{c}$\phi_{\text{LIP}_1}$ \\ T=$25$, $10$ targets\end{tabular}} &
\multicolumn{3}{c}{\begin{tabular}[c]{c}\xuanrev{$\phi_{\text{LIP}_2}$} \\ \xuanrev{T=$25$, $10$ targets}\end{tabular}} \\ \cline{3-14}
\multicolumn{2}{c|}{} & \multirow{2}{*}{LNF} & \multirow{2}{*}{LT} & \multirow{2}{*}{\begin{tabular}[c]{@{}c@{}}\xuanrev{Improv}\\ \xuanrev{-ement} \end{tabular}} & \multirow{2}{*}{LNF} & \multirow{2}{*}{LT} & \multirow{2}{*}{\begin{tabular}[c]{@{}c@{}}\xuanrev{Improv}\\ \xuanrev{-ement} \end{tabular}} & \multirow{2}{*}{LNF} & \multirow{2}{*}{LT} & \multirow{2}{*}{\begin{tabular}[c]{@{}c@{}}\xuanrev{Improv}\\ \xuanrev{-ement} \end{tabular}} & \multirow{2}{*}{\xuanrev{LNF}} & \multirow{2}{*}{\xuanrev{LT}} & \multirow{2}{*}{\begin{tabular}[c]{@{}c@{}}\xuanrev{Improv}\\ \xuanrev{-ement} \end{tabular}} \\
\multicolumn{2}{c|}{} & & & & & & & & & & & & \\ \hline
\multicolumn{2}{l|}{\# bin. vars} & $732$ & $732$ & N.A. & $1281$ & $1281$ & N.A. & $624$ & $624$ & N.A. & \xuanrev{$624$} & \xuanrev{$624$} & \xuanrev{N.A.} \\ \hline
\multicolumn{2}{l|}{\# cont. vars} & $598$ & $602$ & N.A. & $715$ & $720$ & N.A. & $497$ & $503$ & N.A. & \xuanrev{$497$} & \xuanrev{$508$} & \xuanrev{N.A.} \\ \hline
\multicolumn{2}{l|}{\# constr.} & $6899$ & $14724$ & N.A. & $11847$ & $23261$ & N.A. & $3300$ & $3778$ & N.A. & \xuanrev{$3525$} & \xuanrev{$6423$} & \xuanrev{N.A.} \\ \hline
\multicolumn{2}{l|}{\multirow{3}{*}{$G_r$ (\%)}} & $\mathbf{36.5}$ & $41.0$ & & $\mathbf{54.0}$ & $57.3$ & & $\mathbf{14.4}$ & $34.8$ & & \xuanrev{$\mathbf{26.4}$} & \xuanrev{$50.2$} & \\
\multicolumn{2}{l|}{} & $\pm$ & $\pm$ & N.A. & $\pm$ & $\pm$ & N.A. & $\pm$ & $\pm$ & N.A. & \xuanrev{$\pm$} & \xuanrev{$\pm$} & N.A. \\
\multicolumn{2}{l|}{} & $12.2$ & $9.9$ & & $8.8$ & $9.4$ & & $6.6$ & $7.3$ & & \xuanrev{$6.0$} & \xuanrev{$3.2$} & \\ \hline
\multirow{15}{*}{Gurobi} & \multirow{3}{*}{T-Find ($10$\%)} & $\mathbf{2.1}$ & $5.7$ & & $\mathbf{2.9}$ & $4.9$ & & $\mathbf{17.6}$ & $23.0$ & & \xuanrev{$\mathbf{175.2}$} & \xuanrev{$183.5$} & \\
& & $\pm$ & $\pm$ & $\mathbf{2.7\times}$ & $\pm$ & $\pm$ & $\mathbf{1.7\times}$ & $\pm$ & $\pm$ & $\mathbf{1.3\times}$ & \xuanrev{$\pm$} & \xuanrev{$\pm$} & \xuanrev{$\mathbf{1.0\times}$} \\
& & $0.9$ & $2.1$ & & $1.4$ & $0.4$ & & $6.6$ & $18.7$ & & \xuanrev{$58.5$} & \xuanrev{$107.9$} & \\ \cline{2-14}
& \multirow{3}{*}{T-Find (opt.)} & $\mathbf{7.2}$ & $18.0$ & & $32.8$ & $\mathbf{28.4}$ & & $111.8$ & $\mathbf{99.6}$ & & \xuanrev{$\mathbf{571.8}$} & \xuanrev{$592.1$} & \\
& & $\pm$ & $\pm$ & $\mathbf{2.5\times}$ & $\pm$ & $\pm$ & $\mathbf{0.9\times}$ & $\pm$ & $\pm$ & $\mathbf{0.9\times}$ & \xuanrev{$\pm$} & \xuanrev{$\pm$} & \xuanrev{$\mathbf{1.0\times}$} \\
& & $5.8$ & $11.7$ & & $26.6$ & $23.6$ & & $76.5$ & $70.1$ & & \xuanrev{$379.5$} & \xuanrev{$342.3$} & \\ \cline{2-14}
& \multirow{3}{*}{T-Prove ($10$\%)} & $\mathbf{32.9}$ & $88.6$ & & $\mathbf{98.7}$ & $323.0$ & & $\mathbf{31.7}$ & $37.5$ & & \xuanrev{$\mathbf{261.5}$} & \xuanrev{$338.6$} & \\
& & $\pm$ & $\pm$ & $\mathbf{2.7\times}$ & $\pm$ & $\pm$ & $\mathbf{3.3\times}$ & $\pm$ & $\pm$ & $\mathbf{1.2\times}$ & \xuanrev{$\pm$} & \xuanrev{$\pm$} & \xuanrev{$\mathbf{1.3\times}$} \\
& & $26.2$ & $67.6$ & & $76.2$ & $279.4$ & & $28.9$ & $33.3$ & & \xuanrev{$120.1$} & \xuanrev{$198.9$} & \\ \cline{2-14}
& \multirow{3}{*}{T-Prove (optimal)} & $\mathbf{48.0}$ & $142.2$ & & $\mathbf{138.2}$ & $348.0$ & & $163.5$ & $\mathbf{115.6}$ & & \xuanrev{$\mathbf{625.2}$} & \xuanrev{$722.0$} & \\
& & $\pm$ & $\pm$ & $\mathbf{3.0\times}$ & $\pm$ & $\pm$ & $\mathbf{2.5\times}$ & $\pm$ & $\pm$ & $\mathbf{0.7\times}$ & \xuanrev{$\pm$} & \xuanrev{$\pm$} & \xuanrev{$\mathbf{1.2\times}$} \\
& & $37.2$ & $101.6$ & & $103.9$ & $299.0$ & & $126.9$ & $87.3$ & & \xuanrev{$375.3$} & \xuanrev{$349.6$} & \\ 
\cline{2-14}
& \multirow{3}{*}{\xuanrev{Peak Mem (MB)}} & \xuanrev{$\mathbf{495}$} & \xuanrev{$892$} & & \xuanrev{$\mathbf{1242}$} & \xuanrev{$3896$} & & \xuanrev{$\mathbf{814}$} & \xuanrev{$884$} & & \xuanrev{$\mathbf{539}$} & \xuanrev{$641$} & \\
& & $\pm$ & $\pm$ & \xuanrev{$\mathbf{1.8\times}$} & $\pm$ & $\pm$ & \xuanrev{$\mathbf{3.1\times}$} & $\pm$ & $\pm$ & \xuanrev{$\mathbf{1.1\times}$} & \xuanrev{$\pm$} & \xuanrev{$\pm$} & \xuanrev{$\mathbf{1.2\times}$} \\
& & \xuanrev{$69$} & \xuanrev{$107$} & & \xuanrev{$31$} & \xuanrev{$70$} & & \xuanrev{$152$} & \xuanrev{$133$} & & \xuanrev{$186$} & \xuanrev{$57$} & \\
\hline
\multirow{3}{*}{CPLEX} & \multirow{3}{*}{T-Prove (opt.)} & $\mathbf{483.9}$ & $894.6$ & & \multirow{3}{*}{\textit{Timeout}} & \multirow{3}{*}{\textit{Timeout}} & & \multirow{3}{*}{\textit{Timeout}} & \multirow{3}{*}{\textit{Timeout}} & & \multirow{3}{*}{\xuanrev{\textit{Timeout}}} & \multirow{3}{*}{\xuanrev{\textit{Timeout}}} & \\
& & $\pm$ & $\pm$ & $\mathbf{1.8\times}$ & & & N.A. & & & N.A. & & & \xuanrev{N.A.} \\
& & $228.3$ & $232.8$ & & & & & & & & & & \\ \hline \hline
\multicolumn{2}{l|}{\multirow{2}{*}{Speed summary}} & LNF faster & LT faster & Similar & LNF faster & LT faster & Similar & LNF faster & LT faster & Similar & \xuanrev{LNF faster} & \xuanrev{LT faster} & \xuanrev{Similar} \\
\multicolumn{2}{l|}{} & $44/50$ & $2/50$ & $4/50$ & $36/50$ & $4/50$ & $10/50$ & $23/50$ & $20/50$ & $7/50$ & \xuanrev{$25/50$} & \xuanrev{$12/50$} & \xuanrev{$13/50$} \\ \Xhline{3\arrayrulewidth} 
\end{tabular}
\vspace{3pt}
\footnotesize
\noindent \textit{Note: Speed comparison criteria: faster indicates $>20\%$ speedup, similar speed indicates $<20\%$ difference.}
\end{table*}

\begin{table}[t!]
\centering
\caption{Key parameters for the LIP experiment}
\adjustbox{width=\columnwidth}{
\begin{tabular}{lcc}
\toprule
Parameter & Symbol & Value \\
\midrule
Height of COM & $H$ & $1.0$~m \\
Time discretization interval & $dT$ & $0.5$~s \\
Natural frequency & $\omega$ & $\sqrt{g/H} = 3.13$ rad/s \\
Preferred left/right footstep & $\boldsymbol{u}_{\text{pref},(L/R)}$ & $[0, \pm0.15]$~m \\
Circle 1 center (left/right foot) & $\boldsymbol{b}_{1}$ & $[0, \pm 1]^\top$~m \\
Circle 2 center (left/right foot) & $\boldsymbol{b}_{2}$ & $[0, \mp 2.5]^\top$~m \\
Circle 1 radius & $r_1$ & $0.9$~m \\
Circle 2 radius & $r_2$ & $3.1$~m \\
\bottomrule
\end{tabular}}
\label{tab:lip_parameters}
\end{table}

\xuan{Additionally, we predefine the desired contact positions $\boldsymbol{u}_{\text{pref},(L/R)}$ relative to the CoM for each foot to maintain nominal foot placement. Fig.~\ref{fig:LIP_envs} (B) illustrates the feasible footstep region defined by the intersection of the kinematics constraint circles along with these preferred contact positions. The objective function minimizes the deviation between actual footstep placements and these preferred positions, while also incorporating region occupancy costs:}
\begin{align}
f_{\rm obj} = \sum_{k=0}^{N-1} \|\boldsymbol{u}_k - \boldsymbol{u}_{\text{pref},(L/R)}\|^2 + \boldsymbol{\Theta}^\top \boldsymbol{z}^{\pi}
\end{align}
Similar to all previous experiments, $\boldsymbol{\Theta}$ \xuan{represents the predicate costs uniformly sampled from [0, 1], modeling preferences for completing tasks at specific locations and times due to factors such as reduced operational risk.} \xuan{The key parameters for the LIP model are summarized in Table~\ref{tab:lip_parameters}.}

\xuanrev{The STL specifications require the bipedal robot to satisfy conditions involving the $5$ target groups within the planning horizon of $T = 25$ time steps with $dT = 0.5$~s. Each group contains $2$ target regions, each corresponding to a specific safe contact region. The robot must place at least two consecutive footsteps within a selected region for the visit to be considered complete. The target group assignments are illustrated in Fig.~\ref{fig:LIP_envs} (A).}

\xuanrev{We investigate two different types of temporal logic specifications. The first specification, $\phi_{\text{LIP1}}$, requires the robot to visit at least one target region from each of the five distinct target groups:
\begin{equation}
\phi_{\text{LIP1}} = \bigwedge_{i=1}^{5} \left( \bigvee_{j \in G_i} \Diamond_{[0,T]} \square_{[0,1]} \pi^j \right)
\label{eqn:lip_spec1}
\end{equation}
where $G_i$ denotes the set of region indices in the target group $i$ (with $|G_i| = 2$ for all $i$), and $\pi^j$ denotes the predicate indicating that the robot occupies region $j$.}

\xuanrev{The second specification, $\phi_{\text{LIP2}}$, introduces a sequential dependency between the two regions within each group, requiring the robot to visit the prerequisite region before dwelling  the corresponding target region:
\begin{equation}
\phi_{\text{LIP2}} = \bigwedge_{i=1}^{5} \left( (\neg \pi^{a_i} \, \mathcal{U} \, \pi^{b_i}) \wedge \Diamond_{[0,T]} \square_{[0,1]} \pi^{a_i} \right)
\label{eqn:lip_spec2}
\end{equation}
where $a_i$ and $b_i$ denote the target and prerequisite region indices in group $i$, respectively. 
}

\xuanrev{Similar to the point-mass case, we randomly generate $50$ trials by uniformly sampling initial positions within one of the safe regions. 
}


\textbf{Results} The optimized locomotion plan, including the CoM trajectory and foothold locations, is shown in Fig.~\ref{fig:LIP_envs} (C). \xuanrev{Computational efficiency results are presented in the third ($\phi_{\rm LIP1}$) and fourth ($\phi_{\rm LIP2}$) columns of Table~\ref{Tab:dynamics_results}. While LNF continues to achieve significantly tighter root relaxation gaps for both specifications ($14.4\%$ vs $34.8\%$ for $\phi_{\text{LIP1}}$ and $26.4\%$ vs $50.2\%$ for $\phi_{\text{LIP2}}$), the computational advantages become more modest compared with the point-mass experiments. For $\phi_{\text{LIP1}}$, the speed summary shows a nearly even split: LNF is faster in $23$ out of $50$ trials, LT is faster in $20$ trials, and $7$ trials show similar performance, with individual timing metrics ranging from $0.7 \times$ to $1.3 \times$. For $\phi_{\text{LIP2}}$, the split tilts more toward LNF, which is faster in $25$ of the $50$ trials,  whereas LT is faster in $12$, with the remaining $13$ trials showing similar performance and speedups ranging from $1.0 \times$ to $1.3 \times$. Across both specifications, neither method consistently outperforms the other in solving time. Even so, LNF consumes less peak memory than LT for both specifications, reducing peak memory usage by $1.1 \times$ for $\phi_{\text{LIP1}}$ and $1.2 \times$ for $\phi_{\text{LIP2}}$.} This diminished computational advantage occurs because the complex hybrid dynamics formulation, including piecewise linear trigonometric approximations, kinematics feasibility constraints, and footstep placement restrictions, becomes the primary computational bottleneck. While LNF's tighter logic formulation still provides benefits, the dynamics constraints now dominate the overall solution time, limiting the impact of logic formulation improvements on total computational performance.

\textbf{Discussion} These results on hybrid dynamical models, such as the point-mass and LIP models, combined with our earlier planning experiments on temporal graph structures, reveal a clear trend: the simpler the dynamical model, the greater the advantage provided by our LNF formulation. 
In the graph-based planning problems, where dynamics are abstracted as simple transitions between discrete states, LNF achieves the most substantial speedups of up to several orders of magnitude. The point-mass navigation problem with big-M constraints represents an intermediate complexity, yielding $2-3\times$ speedups. The LIP model for locomotion, with its complex trigonometric approximations, shows the most modest improvements despite its tighter relaxation gaps. These results suggest that LNF's computational benefits are most pronounced when temporal logic constraints constitute the primary computational bottleneck. \xuan{When the dynamics become complex, decomposition approaches such as Benders decomposition \citep{ren2025accelerating} can still leverage LNF's computational advantages by solving the temporal logic and simplified dynamics in the master problem while delegating the complex dynamics to the subproblems.}






\subsection{Infeasible Case Detection}
\label{Sec:infeasibility}
\xuanrev{In this section, we evaluate the effectiveness of LNF versus LT in detecting infeasible problems. We construct infeasible instances from a subset of the problem families used in previous studies, including single-robot multi-target planning with $N_g=5$, $\varphi_{\text{vrptw}}$ and $\varphi_{\text{sequential}}$ with $3$ and $6$ robots, point-mass navigation ($\phi_{\text{M2}}$), and bipedal LIP locomotion ($\phi_{\text{LIP1}}$, $\phi_{\text{LIP2}}$). For each problem family, we progressively shorten the planning horizon $T$ until approximately $50\%$ of the instances become infeasible, while keeping all other factors, including the environment setup, task specifications, and costs, randomized as before. This ensures that infeasibility is non-trivial and cannot be detected immediately. For each infeasible instance, we record the number of B\&B nodes explored, the time required to prove infeasibility, and the peak memory consumption. The results are reported as the \textit{median $\pm$ median absolute deviation} across $50$ trials.}

\xuanrev{The results in Table~\ref{tab:infeasibility} show that LNF certifies infeasibility faster than LT in every family. On the graph-based DNF problems, LNF proves infeasibility up to $12.8\times$ faster, while exploring roughly an order of magnitude fewer nodes. The gains are more modest for the hybrid dynamics problems ($1.5\times$ for $\phi_{\text{M2}}$, $1.6\times$ for $\phi_{\text{LIP1}}$, and $1.1\times$ for $\phi_{\text{LIP2}}$), where the complex dynamics formulation becomes the dominant bottleneck. Across all instances, LNF also reduces peak memory by $1.2\times$ to $2.4\times$. Since no incumbent is ever found in these cases, we attribute the performance gap to LNF's tighter convex relaxations, which enable the solver to prune branches and exhaust the search tree more efficiently.}

\renewcommand{\arraystretch}{1.2}
\begin{table*}[t!]
\color{black}  
\setlength{\tabcolsep}{2pt}
\centering
\caption{Computation results for proving infeasibility across different problem instances}
\label{tab:infeasibility}
\footnotesize
\begin{tabular}{@{}l|C{1.0cm}C{1.0cm}|C{1.8cm}C{1.8cm}C{0.9cm}|C{1.8cm}C{1.8cm}C{0.9cm}@{}}
\Xhline{3\arrayrulewidth}
\multirow{2}{*}{Problem instance} & \multicolumn{2}{c|}{Nodes explored} & \multicolumn{3}{c|}{T-Prove infeasibility (s)} & \multicolumn{3}{c}{Peak Memory (MB)} \\ \cline{2-9}
 & LNF & LT & LNF & LT & \begin{tabular}[c]{@{}c@{}}Improv\\-ement\end{tabular} & LNF & LT & \begin{tabular}[c]{@{}c@{}}Improv\\-ement\end{tabular} \\
\Xhline{1.5\arrayrulewidth}
Single robot, $N_g=5$, $T{=}50$ & $\mathbf{420}$ & $1029$ & $\mathbf{160}\pm94$ & $399\pm215$ & $\mathbf{2.5\times}$ & $\mathbf{1738}\pm409$ & $2540\pm455$ & $\mathbf{1.5\times}$ \\
\hline
$\varphi_{\text{vrptw}}$, $3$ robots, $T{=}30$ & $\mathbf{979}$ & $9314$ & $\mathbf{5.9}\pm2.3$ & $81.6\pm38.3$ & $\mathbf{12.8\times}$ & $\mathbf{428}\pm97$ & $904\pm120$ & $\mathbf{2.1\times}$ \\
\hline
$\varphi_{\text{vrptw}}$, $6$ robots, $T{=}30$ & $\mathbf{2233}$ & $13214$ & $\mathbf{38.7}\pm12.6$ & $371.0\pm202.3$ & $\mathbf{9.6\times}$ & $\mathbf{781}\pm90$ & $1061\pm72$ & $\mathbf{1.4\times}$ \\
\hline
$\varphi_{\text{sequential}}$, $3$ robots, $T{=}30$ & $\mathbf{899}$ & $2103$ & $\mathbf{5.7}\pm0.8$ & $19.6\pm2.7$ & $\mathbf{3.4\times}$ & $\mathbf{404}\pm22$ & $592\pm68$ & $\mathbf{1.5\times}$ \\
\hline
$\varphi_{\text{sequential}}$, $6$ robots, $T{=}30$ & $\mathbf{10580}$ & $39178$ & $\mathbf{147.0}\pm49.8$ & $580.0\pm298.9$ & $\mathbf{3.9\times}$ & $\mathbf{852}\pm87$ & $2059\pm34$ & $\mathbf{2.4\times}$ \\
\hline
$\phi_{\text{M2}}$, $T{=}40$ & $\mathbf{24}$ & $1240$ & $\mathbf{2.4}\pm0.9$ & $3.6\pm1.5$ & $\mathbf{1.5\times}$ & $\mathbf{108}\pm18$ & $150\pm11$ & $\mathbf{1.4\times}$ \\
\hline
$\phi_{\text{LIP1}}$, $T{=}15$ & $874$ & $\mathbf{627}$ & $\mathbf{1.21}\pm0.56$ & $1.88\pm0.83$ & $\mathbf{1.6\times}$ & $\mathbf{214}\pm8$ & $289\pm10$ & $\mathbf{1.4\times}$ \\
\hline
$\phi_{\text{LIP2}}$, $T{=}20$ & $3613$ & $\mathbf{2540}$ & $\mathbf{36.7}\pm10.2$ & $41.4\pm25.2$ & $\mathbf{1.1\times}$ & $\mathbf{343}\pm10$ & $421\pm26$ & $\mathbf{1.2\times}$ \\
\Xhline{3\arrayrulewidth}
\end{tabular}
\end{table*}

\subsection{Ablation: Choice of Binary Variable Encoding}
\label{Sec:ablation_encoding}
\xuanrev{We examine two complementary options for imposing binary variables through an
ablation study. The first option enforces the atomic predicate variables
$\boldsymbol{z}^{\pi}$ to be binary while leaving the edge variables
$\boldsymbol{y}$ continuous, as adopted in Lemma~\ref{lemma:dnf_tight_lp} and
Formulation~\ref{eqn:logic_bigM_formulation}. The second keeps
$\boldsymbol{z}^{\pi}$ continuous and instead requires the LNF edge variables
$\boldsymbol{y}$ to be binary, as stated in Remark~\ref{remark:binary_encoding}.
Both choices yield equivalent MICPs. We compare the two encodings on three
representative problems ($\varphi_{\text{vrptw}}$ with $3$ robots,
$\varphi_{\text{sequential}}$ with $3$ robots, and the point mass navigation task
$\varphi_{\text{M1}}$), reported under the $z$-bin and $y$-bin columns of
Table~\ref{Tab:encoding_comparison}, respectively. The two approaches differ only in the number of
binary variables. Overall, the two encodings deliver comparable solving speed,
with $y$-bin mildly faster. It is emphasized that the benefit of tighter convex relaxation is independent of the choice of imposing binary variables. In practice, a user can select the encoding with the smaller binary-variable
count.}

\renewcommand{\arraystretch}{1.3}
\newcolumntype{C}[1]{>{\centering\let\newline\\\arraybackslash\hspace{0pt}}m{#1}}
\begin{table*}[t]
\color{black}
\setlength{\tabcolsep}{3pt}
\centering
\footnotesize
\caption{Computation results across three problems and two LNF binary encoding choices}
\label{Tab:encoding_comparison}
\begin{tabular}{@{}l|C{1.4cm}C{1.4cm}|C{1.4cm}C{1.4cm}|C{1.4cm}C{1.4cm}@{}}
\Xhline{3\arrayrulewidth}
 &
\multicolumn{2}{c|}{\begin{tabular}[c]{c} $\varphi_{\text{vrptw}}$, $3$ Robots \end{tabular}} &
\multicolumn{2}{c|}{\begin{tabular}[c]{c} $\varphi_{\text{sequential}}$, $3$ Robots \end{tabular}} &
\multicolumn{2}{c}{\begin{tabular}[c]{c} $\varphi_{\text{M1}}$ \end{tabular}} \\
\cline{2-7}
 & $z$-bin & $y$-bin & $z$-bin & $y$-bin & $z$-bin & $y$-bin \\ \hline
\# of binary vars   & $490$   & $460$   & $475$   & $465$   & $732$   & $1080$  \\ \hline
\# of constraints   & $9878$  & $9878$  & $10328$ & $10328$ & $7312$  & $7312$  \\ \hline
T-Prove (opt.)       & $12.9\pm6.4$ & $\mathbf{10.4}\pm5.0$ & $22.6\pm18.1$ & $\mathbf{16.7}\pm12.7$ & $\mathbf{28.9}\pm11.5$ & $30.0\pm9.9$ \\ \hline
T-Prove (opt.) min   & $7.8$ & $\mathbf{4.6}$ & $4.1$ & $\mathbf{2.9}$ & $\mathbf{17.4}$ & $18.9$ \\ \hline
T-Prove (opt.) max   & $152.9$ & $\mathbf{146.6}$ & $\mathbf{132.5}$ & $147.8$ & $\mathbf{707.5}$ & $988.2$ \\
\Xhline{3\arrayrulewidth}
\end{tabular}
\end{table*}

\subsection{Ablation: Isolating the Source of LNF's Tighter Convex Relaxation}
\label{Sec:ablation_source}
\xuanrev{In this section, we present a second ablation study to verify that the aggregated constraints~\eqref{eqn:lnf_2} and~\eqref{eqn:lnf_3} are exactly the constraints introduced by our
network-flow construction together with the subsequent Fourier-Motzkin elimination, and that they alone account for the tighter relaxation of LNF over LT. We use the single-robot multi-target problem of Section~\ref{Sec:temporal_graph_experiment} with the specification~\eqref{eqn:multi_target_spec}, an instance of the conjunctive-disjunctive form $\varphi = \bigwedge_{l=1}^{L}\bigvee_{m=1}^{M_l}\bigwedge_{n=1}^{N_{m,l}}\pi_{m,n}^l$ used in the proof of Theorem~\ref{theorem1}. Rather than constructing the general LNF formulation, we build the cuts~\eqref{eqn:lnf_2}-\eqref{eqn:lnf_3} directly inside the LT structure for this specific problem. We first set aside the obstacle term $\square_{[0,T]} \neg \text{obstacle}$, which grounds into single-predicate conjuncts rather than disjunctions and therefore induces no aggregated constraint. The remaining $N_g$ reach requirements carry the relevant structure: the top level is the conjunction over target groups $k$, the middle level is a disjunction indexed by the time-target pair $(\tau, j)$, and the bottom level is a conjunction whose predicate set $P_{m,l} = \{\text{target}^j[\tau],\ \text{target}^j[\tau+1]\}$ contains two elements owing to the unit dwell $\square_{[0,1]}$.}

\xuanrev{To form the LNF cuts, we examine, within each disjunction node, the branches that share a common predicate. Because the unit dwell $\square_{[0,1]}$ forces each branch to span two consecutive time steps $\{\tau,\tau+1\}$, a target predicate $\text{target}_k^j[\tau+1]$ can appear only in the branch that dwells from $\tau$ and the branch that dwells from $\tau+1$, namely $(\tau,j)$ and $(\tau+1,j)$; it is therefore shared by at most two branches. For every predicate $\pi$ shared by more than one branch, we introduce a single aggregated inequality bounding its indicator by the sum of the indicators of all branches that contain it, $z^{\pi}\ge\sum_{m:\,\pi\in P_{m,k}} z^{\varphi_{m,k}}$ for a non-negative predicate, and the complementary $z^{\pi}\le 1-\sum_{m:\,\pi\in P_{m,k}} z^{\varphi_{m,k}}$ for a negated one. These aggregated inequalities are identical to constraints~\eqref{eqn:lnf_2}--\eqref{eqn:lnf_3} that naturally arise from our network-flow construction and Fourier--Motzkin elimination, and we add them to the LT formulation alongside its original branch--predicate constraints~\eqref{eqn:lt_4}--\eqref{eqn:lt_5}.}

\xuanrev{We evaluate the aggregated inequalities introduced in the preceding paragraph by comparing the original LNF, the original LT, and the LT augmented with these aggregated inequalities (LT+cut) across problem sizes $N_g\in\{2,3,4,5\}$, running 5 randomized instances per $N_g$; results are collected in Table~\ref{Tab:lnf_cuts}. The root relaxation gap of LT+cut matches that of LNF exactly across all problem sizes --- for example, $11.0\%$ versus $32.3\%$ for LT at $N_g=2$, and $51.8\%$ versus $65.9\%$ for LT at $N_g=5$ --- confirming that the aggregated inequalities~\eqref{eqn:lnf_2}--\eqref{eqn:lnf_3} are precisely the source of LNF's tighter convex relaxation. While LT+cut recovers the same relaxation tightness as LNF, it does so at the cost of a modestly larger constraint count, since the original branch--predicate
constraints~\eqref{eqn:lt_4}--\eqref{eqn:lt_5} are retained. In terms of solving speed, LT+cut is comparable to LNF and both are substantially faster than the original LT. Together, these results demonstrate that the aggregated inequalities arising from the network-flow construction and Fourier--Motzkin elimination are the key mechanism behind LNF's computational advantage.}

\renewcommand{\arraystretch}{1.3}
\newcolumntype{C}[1]{>{\centering\let\newline\\\arraybackslash\hspace{0pt}}m{#1}}
\begin{table*}[t]
\color{black}
\setlength{\tabcolsep}{3pt}
\centering
\footnotesize
\caption{Recovering LNF's relaxation tightness inside the Logic Tree via LNF's aggregated inequalities}
\label{Tab:lnf_cuts}
\resizebox{\textwidth}{!}{%
\begin{tabular}{@{}l|C{1.5cm}C{1.5cm}C{1.5cm}|C{1.5cm}C{1.5cm}C{1.5cm}|C{1.5cm}C{1.5cm}C{1.5cm}|C{1.5cm}C{1.5cm}C{1.5cm}@{}}
\Xhline{3\arrayrulewidth}
 &\multicolumn{3}{c|}{\begin{tabular}[c]{c} $N_g=2$ \end{tabular}} &\multicolumn{3}{c|}{\begin{tabular}[c]{c} $N_g=3$ \end{tabular}} &\multicolumn{3}{c|}{\begin{tabular}[c]{c} $N_g=4$ \end{tabular}} &\multicolumn{3}{c}{\begin{tabular}[c]{c} $N_g=5$ \end{tabular}} \\
\cline{2-13}
 & LNF & LT & LT+cut & LNF & LT & LT+cut & LNF & LT & LT+cut & LNF & LT & LT+cut \\ \hline
\# of binary vars   & $178$ & $178$ & $178$ & $417$ & $417$ & $417$ & $756$ & $756$ & $756$ & $1195$ & $1195$ & $1195$ \\ \hline
\# of cont.\ vars   & $37619$ & $37621$ & $37621$ & $58678$ & $58681$ & $58681$ & $79697$ & $79701$ & $79701$ & $100676$ & $100681$ & $100681$ \\ \hline
\# of constraints   & $9930$ & $10195$ & $10285$ & $15330$ & $15967$ & $16192$ & $20850$ & $22019$ & $22439$ & $26490$ & $28351$ & $29026$ \\ \hline
$G_r$ (\%)          & $\mathbf{11.0}\pm13.4$ & $32.3\pm12.6$ & $\mathbf{11.0}\pm13.4$ & $\mathbf{24.8}\pm8.8$ & $47.0\pm6.8$ & $\mathbf{24.8}\pm8.8$ & $\mathbf{37.5}\pm8.0$ & $56.6\pm4.9$ & $\mathbf{37.5}\pm8.0$ & $\mathbf{51.8}\pm9.3$ & $65.9\pm6.0$ & $\mathbf{51.8}\pm9.3$ \\ \hline
T-Prove (opt.)       & $\mathbf{0.1}\pm0.0$ & $0.1\pm0.0$ & $\mathbf{0.1}\pm0.0$ & $0.7\pm0.2$ & $1.8\pm0.2$ & $\mathbf{0.6}\pm0.1$ & $8.6\pm2.5$ & $18.7\pm4.6$ & $\mathbf{8.3}\pm1.6$ & $\mathbf{96.6}\pm30.4$ & $465.0\pm139.0$ & $148\pm73.9$ \\
\Xhline{3\arrayrulewidth}
\end{tabular}}
\end{table*}

\subsection{Real-Time Replanning with Multiple Quadrupedal Robots}
\label{sec:experiment_hardware}
\begin{figure*}[!htbp]
    \centering
    \includegraphics[width=0.95\textwidth]{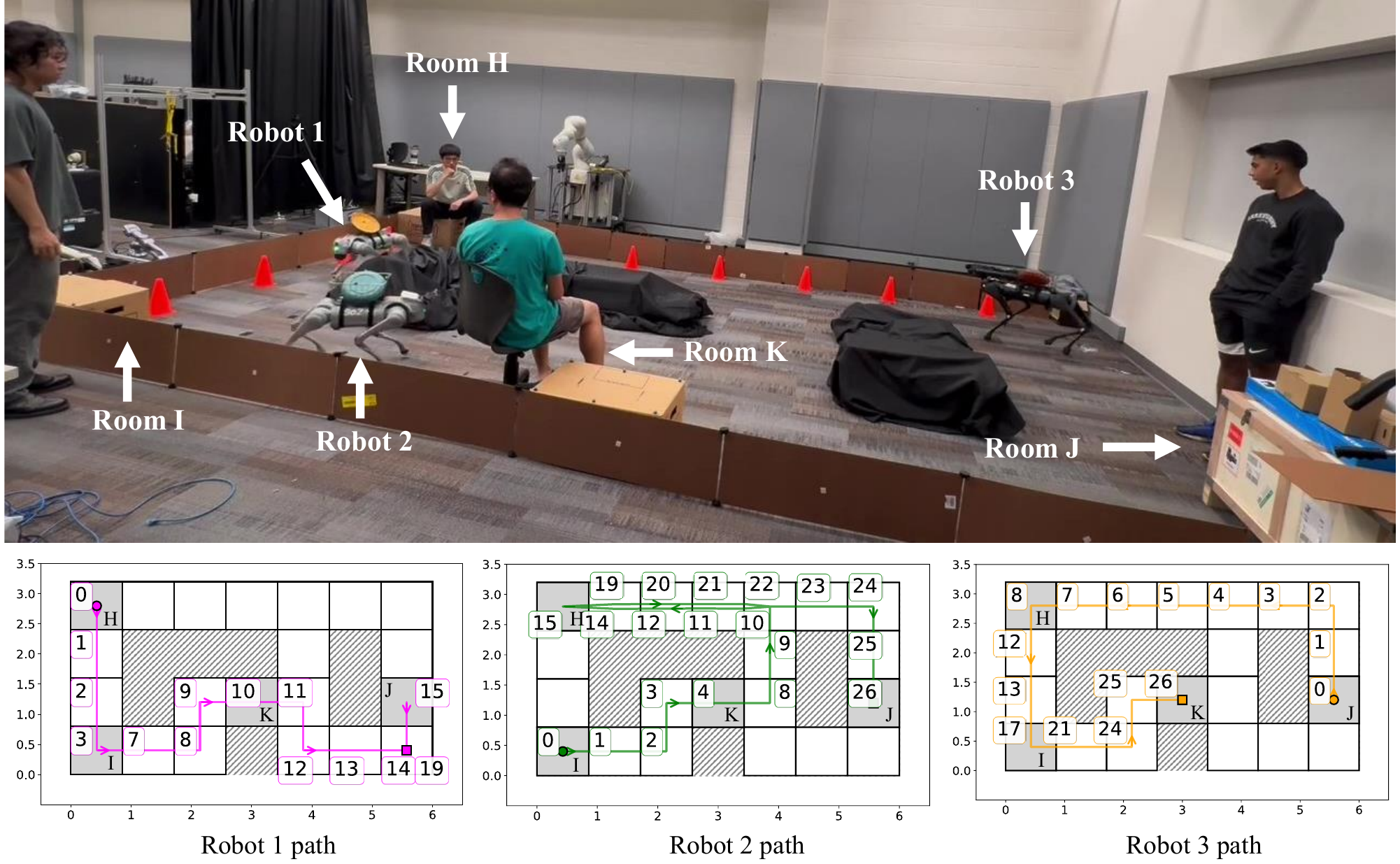}
    \caption{Top: Physical robot demonstration setup showing three robots with labeled task rooms H, I, J, and K. Bottom: Computed trajectories for Robot $1$ (magenta), Robot $2$ (green), and Robot $3$ (orange) with time annotations ($k=0$ to $29$) showing the planned paths through the grid. Arrows indicate direction of motion along each path. 
    }
\label{fig:hardware_1}
\end{figure*}

\begin{figure*}[!htbp]
    \centering
    \includegraphics[width=0.8\textwidth]{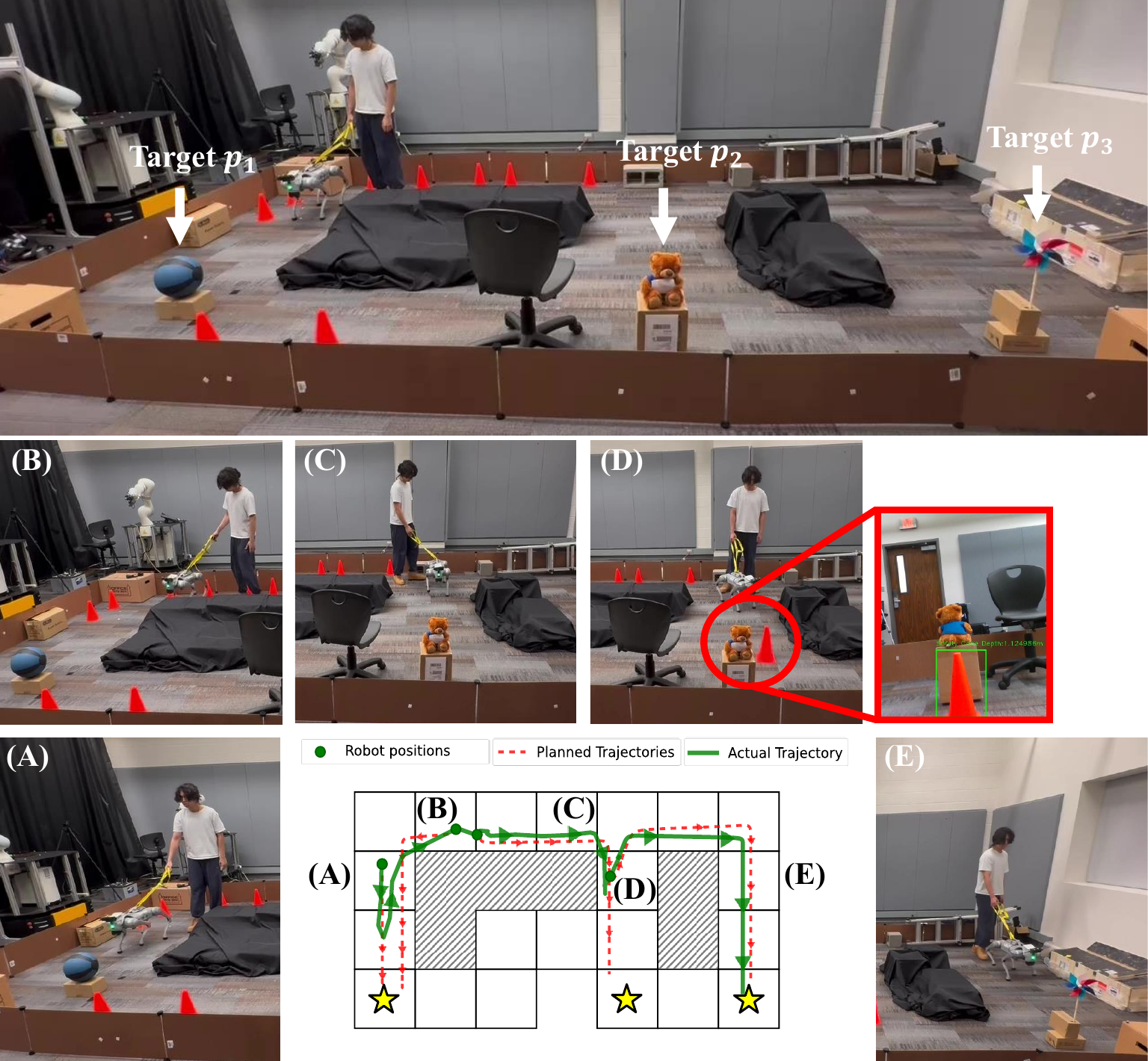}
    \caption{A quadrupedal robot navigates toward three potential target locations ($p_1$, $p_2$, $p_3$) with continuous replanning under STL specifications. The central trajectory plot shows the robot's planned paths (red dotted lines) at several key instants with green dots indicating robot positions, and actual executed trajectory (green solid line). (A) The robot initially plans a path toward target $p_1$. (B) A human operator physically drags the robot backward, and the replanning drives the robot toward target $p_2$ (C) as a more optimal choice given the new position. (D) A cone obstacle is placed in the robot's path, which is recognized by the onboard RGB-D camera (highlighted in red box), making target $p_2$ infeasible. (E) The robot automatically replans and successfully reaches the new optimal goal, i.e., target $p_3$. 
    }
\label{fig:hardware_2}
\end{figure*}

We demonstrate the temporal logic motion planning as well as the real-time replanning capabilities of our LNF framework with two hardware experiments. The first experiment involves three quadrupedal robots targeting object delivery tasks with sequential ordering and collision avoidance requirements (see Fig.~\ref{fig:hardware_1}). The second experiment studies a single quadrupedal robot that must visit a disjunction of targets while dynamically avoiding unforeseen triangular traffic cones detected by an onboard camera (see Fig.~\ref{fig:hardware_2}), showing instant replanning capabilities under dynamically changing environmental conditions.

The hardware experiments are run by Unitree quadrupedal robots with an  overhead motion capture system providing real-time robot position and orientation status. The robots navigate in a confined workspace populated with static obstacles. A Euclidean-space PID controller regulates the linear velocities in the horizontal plane and the yaw angular velocity to accurately track the planned trajectories.

For the first experiment in Fig.~\ref{fig:hardware_1}, we implement Formulation~\ref{eqn:logic_dnf_formulation} for multi-robot coordination. We set up a workspace with dimensions $6$ m $\times$ $3$ m , discretized into 21 regions of equal size. The environment designated four areas, labeled $H$, $I$, $J$, and $K$, to represent \jimrev{spots} where the staff collect different components to assemble products. 
Three quadrupedal robots carry the necessary objects with delivery deadlines $T_1$, $T_2$, and $T_3$, respectively. The robots start from the designated positions: Robot $1$ at room $H$, Robot $2$ at room $I$, and Robot $3$ at room $J$. The assembly processes require robots to arrive at each room in a specific sequence to ensure proper assembly: room $H$ requires Robot $3$ to arrive first, followed by Robot $2$; room $I$ requires Robot $1$ to be the first one to show up and then Robot $3$, room $J$ requires Robot $1$ first followed by Robot $2$, and room $K$ requires Robot $2$ first followed by Robot $3$. For each visit, the robot stays for $2$ time units to allow the staff to pick up the object from the robots. The temporal logic specification of this task is:
\begin{align*}
&\phi_{\text{task}} = \\
&(\Diamond_{[0,T_1]} \Box_{[0,2]} \pi^{1 \rightarrow I}) \wedge (\Diamond_{[0,T_1]} \Box_{[0,2]} \pi^{1 \rightarrow J}) \wedge \\
&(\Diamond_{[0,T_2]} \Box_{[0,2]} \pi^{2 \rightarrow H}) \wedge (\Diamond_{[0,T_2]} \Box_{[0,2]} \pi^{2 \rightarrow J}) \wedge \\
&(\Diamond_{[0,T_2]} \Box_{[0,2]} \pi^{2 \rightarrow K}) \wedge (\Diamond_{[0,T_3]} \Box_{[0,2]} \pi^{3 \rightarrow H}) \wedge \\
&(\Diamond_{[0,T_3]} \Box_{[0,2]} \pi^{3 \rightarrow I}) \wedge (\Diamond_{[0,T_3]} \Box_{[0,2]} \pi^{3 \rightarrow K}) \wedge \\
&(\neg \pi^{2 \rightarrow H} \, \mathcal{U} \, \pi^{3 \rightarrow H}) \wedge (\neg \pi^{2 \rightarrow J} \, \mathcal{U} \, \pi^{1 \rightarrow J}) \wedge \\ &(\neg \pi^{3 \rightarrow I} \, \mathcal{U} \, \pi^{1 \rightarrow I}) \wedge (\neg \pi^{3 \rightarrow K} \, \mathcal{U} \, \pi^{2 \rightarrow K})
\end{align*}
where $\pi^{1 \rightarrow I}$ represents Robot $1$ visiting room $I$, and so on. 

The four rooms are positioned in separate locations connected by narrow hallways with a T-shaped junction at the center. Robots need to coordinate their passing order at the junction to avoid congestion. 
We assign costs to both the robot’s movement through hallways and its waiting time in rooms. These time-varying costs reflect factors such as rooms having limited staff availability at certain times (\textit{e.g.}, during breaks) and hallways experiencing varying congestion throughout the day.

In the first experiment, three Unitree quadruped robots 
are deployed to execute the planned trajectories,  where each robot is assigned a temporal graph. Each cell in Fig. \ref{fig:hardware_1}  is represented by a vertex in the graph. Adjacent cells are connected by edges. We set the capacity of any edge to 1 such that two robots cannot simultaneously traverse the same edge or occupy the same cell. 
We choose a planning horizon $T=30$ and build the \xuanrev{corresponding} DNF. To prevent collisions between robots, we implement the conflict set constraints described in \xuanrev{Section~\ref{sec:dnf_2}}. Specifically, we define conflict sets for each vertex containing robot-vertex pairs that prevent multiple robots from occupying the same location, and conflict sets for each bidirectional edge containing robot-edge pairs that prevent head-to-head collisions.

\textbf{Results}
We solve the problem using Gurobi 12.0, which contains $6,054$ variables ($5,655$ binary, $399$ continuous) and $5,829$ constraints. The optimizer finds the globally optimal trajectories for all three robots within $8$ seconds, as shown in Fig.~\ref{fig:hardware_1}. In addition, we deploy the optimized trajectories on real quadrupedal robots (see the submission video). The hardware demonstration validates that the robots successfully deliver components following the sequential ordering constraints and timing requirements specified in the STL specification, while avoiding collisions with other robots.

For the second experiment, we use \xuanrev{exactly the same} environment as the first experiment and \xuanrev{implement LNF in continuous configuration space} for real-time temporal logic replanning. A single quadrupedal robot must select from $3$ disjunctive goals. Rather than using Formulation~\ref{eqn:logic_bigM_formulation}, we describe the robot's CoM trajectory through Bézier curves, as remarked in Remark~\ref{remark1}. Only CoM motion is needed to plan paths for quadrupedal robots, and using Bézier curves reduces problem dimensionality by representing smooth trajectories through a small number of control points. Note that the specific choice of dynamics formulation does not affect the validation of our method, as the LNF is the core contribution that we validate.

We use two Bézier curves: one degree-3 Bézier curve for $x-y$ position and one degree-1 curve for time scaling, to model robot motion. Collision avoidance with the environment is ensured through the convex hull property of Bézier curves. The formulation includes initial position and velocity constraints enforced through the $0$-th and $1$-st order derivatives of the position Bézier curve, which also enforces the robot's initial orientation by assuming it aligns with the velocity direction. This approach significantly reduces problem dimensionality by representing smooth trajectories through a small number of control points. For details of the Bézier curve formulation, we refer readers to the work in \citep{marcucci2023motion}.

Integration with LNF is achieved using binary edge variables $y_{ij}^k$ as predicate variables, similar to the DNF formulation in \eqref{eqn:predicate_dnf}. To enable this mapping, we add constraints ensuring that traversing each region takes a fixed duration $\Delta t$, so that the discrete time steps of $y_{ij}^k$ correspond to actual time intervals $[k\Delta t, (k+1)\Delta t)$. In future work, we plan to relax this uniform timing constraint to allow variable region traversal times. We designate $3$ target regions $p_1$, $p_2$, $p_3$, and the STL specification requires the robot to visit at least one of them with a minimum dwell time. The temporal logic specification is:
\begin{align}
\phi_{\text{replanning}} = \bigvee_{i \in \{p_1, p_2, p_3\}} \Diamond_{[0,T]} \Box_{[0,1]} \pi^i
\end{align}

The planner runs in a model predictive control fashion, continuously replanning as initial conditions and the environment change. We introduce two types of disturbances to demonstrate adaptive goal selection based on optimization feasibility and cost: manual perturbations where a person drags the robot to different locations, and traffic control cone obstacles suddenly showing up during robot motion. The Go2 robot is equipped with an Intel RealSense RGB-D camera, which detects cone positions in the 3D space. The cone positions are transformed into global coordinates using the robot's own pose, providing the absolute location of cones in the environment. The replanning algorithm then adds constraints $y_{ij}^k = 0,$ $\forall k$ to any edge entering cells within a 1-meter blocking radius of the cone, removing those regions from the feasible path options. The objective function minimizes the path length while applying an exponential time penalty $\alpha^k$ ($\alpha > 1$) for delayed goal completion. \xuan{Specifically, when the robot reaches the goal region and remains there at time step $k$, a penalty of $\alpha^k$ is added to the objective, discouraging late arrivals at target locations.} 
The optimization model is built only once and resolved with changing initial conditions, leveraging Gurobi's internal warm-start capability to accelerate the computation. The warm-start provides a speedup of around $10\%$ for this specific case. 

\textbf{Results}
We conduct multiple trials as demonstrated in the submitted video, where the robot initially plans to pursue a goal and automatically switches targets due to human interference and dynamic obstacles. Throughout the process, the planner continuously operates and generates a series of trajectories in real-time. Fig.~\ref{fig:hardware_2} shows one representative behavior. Initially, the planner drives the robot toward target $1$ (subfigure (A)). When a person drags the robot back significantly, moving it farther from target $1$ (subfigure (B)), the robot automatically switches to pursue target $2$ as that new target becomes more optimal (subfigure (C)). During the approach to target $2$, a person places a cone obstacle in the robot's path, making target $2$ infeasible (subfigure (D)). The robot quickly responds and redirects itself to target $3$ instead (subfigure (E)). Overall, the system maintains a $2-10$ Hz replanning frequency.

It is worth noting that the continuous replanning demonstrated here is designed to showcase the system's rapid response capabilities. In practice, event-driven replanning triggered by obstacle detection or human intention recognition would be more appropriate, as our current implementation can introduce unnecessary replanning causing the robot to oscillate between different trajectories.

\section{Discussion} 
\label{Sec:discussion}
\subsection{LNF Performance with Solver Presolve and Heuristics}

Modern MIP solvers employ presolve techniques — such as the introduction of valid inequalities — to tighten formulations and reduce the number of variables and constraints. These procedures benefit both LNF and LT formulations. In our experiments, the reported convex relaxation gaps are measured after presolve, as they reflect the actual formulations processed by the solver. Our experiments demonstrate that the advantages of LNF formulations, endowed by their tighter convex relaxations and reduced number of constraints, hold consistently with or without presolve.

While LNF provides tighter relaxations, MIP solvers continue to rely on heuristics to identify feasible binary solutions. Although we have not systematically investigated this, a hybrid formulation incorporating LT constraints \eqref{eqn:lt_4} and \eqref{eqn:lt_5} into the LNF formulation may potentially enhance solver heuristics for handling complex logic specifications.




\subsection{Incorporating Robustness Objectives}
\xuanrev{Our LNF formulation can be readily extended to incorporate STL robustness metrics, which quantify how strongly a specification is satisfied. As given in Table~\ref{tab:robustness}, the robustness of an atomic predicate is the signed margin $\rho^{\pi}(\boldsymbol{x},k) = (\boldsymbol{a}^{\pi})^\top
\boldsymbol{x}_k + b^{\pi}$. For the general class of PWA dynamics in continuous
configuration space (Section~\ref{sec:continuous_dynamics_bigM},
Formulation~\ref{eqn:logic_bigM_formulation}), the state $\boldsymbol{x}_k$ is a decision variable, so this margin is linear in the optimization variables. We introduce a single scalar variable $\rho$ denoting the minimal robustness across the active predicates, linked to each predicate through $\rho \leq \rho^{\pi}(\boldsymbol{x},k) + M(1 - z^{\pi^i_k})$ for all $\pi^i \in \Pi$ and time step $k$. Guaranteeing a minimal robustness level then reduces to the single linear constraint $\rho \geq \rho_{\min}$, while maximizing robustness amounts to appending the linear term $-\lambda \rho$,
$\lambda > 0$, to the objective $f_{\text{obj}}(\boldsymbol{\xi}, \boldsymbol{z}^{\pi})$, as done in Eqn.~(15) of~\citet{belta2019formal}.}

\xuanrev{For the temporal graph and DNF abstraction in Section~\ref{Sec:temporal_graph_experiment}, the configuration space is discretized, and the dynamics reduce to transitions between vertices at fixed physical locations. To approximately impose robustness as a coarse surrogate, we may discretize each predicate-satisfying region into finer patches, and for each patch, represented by a vertex $p$ corresponding to a physical location $\boldsymbol{p}$ inside this patch, we may use a constant robustness margin defined as $\rho^{\pi}(\boldsymbol{p}) = (\boldsymbol{a}^{\pi})^\top \boldsymbol{p} +
b^{\pi}$, assigned as a cost of remaining at $\boldsymbol{p}$ that decreases as the
margin increases. For example, patches near the region boundary are penalized more than patches in the interior, reflecting deeper satisfaction of the predicate being safer against disturbances. The robustness term in the objective hence takes the form $-\lambda\sum_{\pi\in\Pi}\sum_{k=0}^{T}\rho^{\pi}(\boldsymbol{p})\,z^{\pi}_k$, which fits the existing linear term $\boldsymbol{\Theta}^\top \boldsymbol{z}^{\pi}$. This cost thereby biases the planner toward more robust paths within the discrete approximation.}

\section{Conclusion and Future Work} 
\label{Sec:conclusion}
In this paper, we present LNF, a novel optimization formulation that encodes temporal logic specifications as network flow constraints inspired by the GCS framework, along with a network-flow-based Fourier-Motzkin elimination procedure that projects the formulation onto a lower-dimensional space. LNF achieves provably tighter convex relaxations compared to the standard LT approach while reducing the number of constraints. Experimental results demonstrate computational speedups ranging from $2-3\times$ for complex dynamics to orders of magnitude for Temporal-Graph-based planning problems. Hardware experiments with quadrupedal robots validate the framework's real-time replanning capabilities under dynamically changing environmental conditions.

Future research directions include integrating LNF with decomposition approaches such as Benders Decomposition \citep{ren2025accelerating}, where LNF focuses on solving simplified dynamics in the master problem while complicated dynamics are handled in subproblems. \xuanrev{Another potential direction is to explicitly decompose the specification along the time axis~\citep{yu2023model}, solving each time-disjoint sub-task over a shorter horizon while using Benders cuts to construct the required terminal feasible sets lazily and online rather than enumerating them in full offline, since their set representation can grow exponentially in the number of temporal operators.} \xuanrev{Additionally, developing theoretical bounds on the relaxation gap for specific problem classes would provide more performance guarantees. One tractable route is a worst-case analysis: starting from an arbitrary feasible point of the convex relaxation and using the tightness of the aggregated constraints~\eqref{eqn:lnf_2} and~\eqref{eqn:lnf_3} to substitute the predicate variables $\boldsymbol{z}^{\pi}$ with the edge variables $y_{m,l}$, the relaxed objective can be related to the integral optimum.} The network-flow-based Fourier-Motzkin elimination procedure also shows promise for simplifying other GCS formulations beyond temporal logic, suggesting broader applicability across motion planning and hybrid control problems.

\begin{acks}
The authors are grateful to Prof. Samuel Coogan for his valuable support and guidance throughout this work. The authors would also like to thank the members of the Laboratory for Intelligent Decision and Autonomous Robots (LIDAR) and other colleagues at the Georgia Institute of Technology for their valuable discussions and assistance with the experiments.
\end{acks}

\subsection*{\normalsize\sagesf\bfseries Author Contributions}
\begin{refsize}\noindent 
All authors contributed to the study. Xuan Lin: Conceptualization, Methodology, Formal analysis, Investigation, Software, Project administration, Writing – original draft. Jiming Ren: Methodology, Software, Visualization, Writing – review \& editing. Yandong Luo: Investigation, Software, Writing – review \& editing. Weijun Xie: Supervision. Ye Zhao: Funding acquisition, Resources, Supervision. All authors reviewed and approved the final version of the manuscript.
\end{refsize}

\subsection*{\normalsize\sagesf\bfseries Ethical Considerations}
\begin{refsize}\noindent Not applicable.\end{refsize}

\subsection*{\normalsize\sagesf\bfseries Consent to Participate}
\begin{refsize}\noindent Not applicable.\end{refsize}

\subsection*{\normalsize\sagesf\bfseries Consent for Publication}  
\begin{refsize}\noindent Not applicable.\end{refsize}

\begin{dci}
The author(s) declared no potential conflicts of interest with respect to the research, authorship, and/or publication of this article.
\end{dci}

\begin{funding}
The author(s) disclosed receipt of the following financial support for the research, authorship, and/or publication of this article: This work was supported in part by the Office of Naval Research (ONR) under Grant N000142312223; in part by the National Science Foundation (NSF) under Grant IIS-1924978, Grant CMMI-2144309, and Grant FRR-2328254.
\end{funding}


{
\bibliographystyle{SageH}
\bibliography{references}
}

\appendix
\section{Index to Multimedia Extensions}

Archives of IJRR multimedia extensions published prior to 2014 can be found at http://www.ijrr.org; after 2014 all videos are available on the IJRR YouTube channel at http://www.youtube.com/user/ijrrmultimedia

\begin{table}[h]
\footnotesize
\centering
\caption*{\raggedright Table of Multimedia Extensions}
\begin{tabular}{clp{4.5cm}}
\toprule
Extension & Media Type & Description \\
\midrule
1 & Video & Demonstrations of experimental results, including: \newline
(1) Multi-robot search and rescue using trajectory libraries; \newline
(2) Bipedal locomotion planning with Linear Inverted Pendulum dynamics under temporal logic constraints; \newline
(3) Hardware demonstrations on multiple quadrupedal robots temporal logic planning; \newline
(4) Hardware demonstrations on a single quadrupedal robot real-time re-planning under disturbances. \\
\bottomrule
\end{tabular}
\end{table}

\section{Constructing LNFs from LTs}
\label{appendix_LT_to_LNF}
\jimrev{We briefly introduce Algorithm~\ref{Algorithm:build_node} that converts an LT to an LNF, although a similar strategy can be applied directly to STL specifications without first building an intermediate LT. Let $e \in \mathcal{E}$ denote an arbitrary edge in the LNF, and for each vertex $v \in \mathcal{V}$, let $\mathcal{E}^{\rm in}_{v}$ and $\mathcal{E}^{\rm out}_{v}$ denote its sets of incoming and outgoing edges, respectively.  The algorithm is initialized by calling $\Call{BuildNode}{T^{\varphi_0}, v_{s}, e, P_{e}}$, where $v_s$ is the source vertex, $e \in \mathcal{E}^{\rm out}_{v_s}$ is a ``dangling" outgoing edge with no assigned target, and $P_{e} = \varnothing$ is an empty predicate set associated with $e$. Starting from the root of the LT, the algorithm performs a post-order traversal: it processes conjunction nodes by threading the same edge and predicate set through all children, and it processes disjunction nodes by instantiating a new vertex once all of that node's children have been visited (line \ref{new_node}). Because new edges are only ever created from an existing disjunction node to vertices generated afterward, the post-order traversal naturally yields a topological ordering, guaranteeing that the resulting LNF is acyclic. The algorithm terminates by returning the target vertex $v_t$.}

\begin{algorithm}[!ht]
	\caption{$\protect \Call{BuildNode}{}$}
        \label{Algorithm:build_node}
         \textbf{Input}: An LT node $T^{\varphi_i}$, a vertex $v$, an outgoing edge $e$ of $v$, and a predicate set $P_e$ for edge $e$.\\
         \textbf{Output}: A vertex $v$, an outgoing edge $e$ of $v$, and a predicate set $P_e$ for edge $e$.
	\begin{algorithmic}[1]
    \If{$T^{\varphi_i} = \pi^\varphi$ is a leaf node}
        \State Add $\pi^\varphi$ to $P_e$.
        \State \textbf{return} $v$, $e$, $P_e$
    \EndIf
    \If{$\circ(T^{\varphi_i}) = \wedge$}
        \For{each child $T^{\varphi_j}$ of $T^{\varphi_i}$}
        \State $v$, $e$, $P_e$ =  \Call{BuildNode}{$T^{\varphi_j}$, $v$, $e$, $P_e$}
        \EndFor
        \State \textbf{return} $v$, $e$, $P_e$
    \EndIf
    \If{$\circ(T^{\varphi_i}) = \vee$}
    \State (\textit{Assume $n$ to be the number of subnodes of $T^{\varphi_i}$})
    \State Duplicate $e$, $P_e$ for $n-1$ times, denote as $e_j$, $P_{e,j}$, \phantom . \phantom . \phantom . where $j=1,\ldots,n$.
    \For{each child $T^{\varphi_j}$ of $T^{\varphi_i}$}
    \State $v$, $e_{o,j}$, $P_{o,j}$ = \Call{BuildNode}{$T^{\varphi_j}$, $v$, $e_j$, $P_{e,j}$}
    \EndFor
    \State Initialize a new vertex $v_{\varphi_i}$, an outgoing edge $e_{\varphi_i}$, \phantom .  \phantom .  \phantom .  \phantom . and a set $P_{e_{\varphi_i}}=\varnothing$ for the edge $e_{\varphi_i}$ \label{new_node} 
    \State Assign $\mathcal{E}_{v_{\varphi_i}}^{\rm in}$ = \{$e_{o,j}$, $j=1,\ldots,n$\}.
    \State \textbf{return} $v_{\varphi_i}$, $e_{\varphi_i}$, $P_{e_{\varphi_i}}$.
    \EndIf
    
	\end{algorithmic} 
\end{algorithm} 

\section{Rigorous Process of Network-flow-based Fourier-Motzkin Elimination}
\label{appendix_project_one_by_one}
In this appendix, we present the rigorous F-M elimination procedure that projects the original GCS-inspired formulation onto a lower-dimensional space with fewer continuous variables and constraints. Through systematic variables elimination, this process generates a formulation that is mathematically equivalent to the original model, preserving the same feasible region and optimal solution, and maintaining the same tightness of convex relaxation.

Consider the flow variables $\boldsymbol{\omega}_e$ for edges $e \in \mathcal{E}^{\rm{out}}_{v_s}$ originating from the source vertex $v_s$. We eliminate these edge flow variables one edge at a time, treating each $\boldsymbol{\omega}_e \in [0,1]^{|\Pi|}$ as a batch to be removed.

For edge $e_1 \in \mathcal{E}^{\rm{out}}_{v_s}$ that connects to vertex $v_1$, we collect all constraints involving $\boldsymbol{\omega}_{e_1}$. From the convex set constraints \eqref{eqn:edge}, and the flow conservation at the source \eqref{eqn:input-b}, we have:
\begin{align}
y_{e_1}\boldsymbol{v}^+_{e_1} &\leq \boldsymbol{\omega}_{e_1} \leq \boldsymbol{1}_{|\Pi|} - y_{e_1}\boldsymbol{v}^-_{e_1} \\ \nonumber
\boldsymbol{z}^{\pi} &= \boldsymbol{\omega}_{e_1} + \sum_{e \in \mathcal{E}^{\rm{out}}_{v_s}\backslash\{e_1\}} \boldsymbol{\omega}_e \nonumber
\end{align}
Let $v_1$ denote the vertex that $e_1$ enters, i.e., $e_1 \in \mathcal{E}^{\rm{in}}_{v_1}$. From the flow conservation constraint \eqref{eqn:vertex-b} at $v_1$:
\begin{equation}
\boldsymbol{\omega}_{e_1} + \sum_{e \in \mathcal{E}^{\rm{in}}_{v_1}\backslash\{e_1\}} \boldsymbol{\omega}_e = \sum_{e \in \mathcal{E}^{\rm{out}}_{v_1}} \boldsymbol{\omega}_e \nonumber
\end{equation}

Collecting all constraints involving $\boldsymbol{\omega}_{e_1}$, we have the following lower bounds:
\begin{subequations}
\begin{align}
\boldsymbol{\omega}_{e_1} &\geq y_{e_1}\boldsymbol{v}^+_{e_1} \label{completeFM:lb0}\\
\boldsymbol{\omega}_{e_1} &\geq \boldsymbol{z}^{\pi} - \sum_{e \in \mathcal{E}^{\rm{out}}_{v_s}\backslash\{e_1\}} \boldsymbol{\omega}_e \label{completeFM:lb1}\\
\boldsymbol{\omega}_{e_1} &\geq \sum_{e \in \mathcal{E}^{\rm{out}}_{v_1}} \boldsymbol{\omega}_e - \sum_{e \in \mathcal{E}^{\rm{in}}_{v_1}\backslash\{e_1\}} \boldsymbol{\omega}_e \label{completeFM:lb2}
\end{align}
\end{subequations}
and upper bounds:
\begin{subequations}
\begin{align}
\boldsymbol{\omega}_{e_1} &\leq \boldsymbol{1}_{|\Pi|} - y_{e_1}\boldsymbol{v}^-_{e_1} \label{completeFM:ub0}\\
\boldsymbol{\omega}_{e_1} &\leq \boldsymbol{z}^{\pi} - \sum_{e \in \mathcal{E}^{\rm{out}}_{v_s}\backslash\{e_1\}} \boldsymbol{\omega}_e \label{completeFM:ub1}\\
\boldsymbol{\omega}_{e_1} &\leq \sum_{e \in \mathcal{E}^{\rm{out}}_{v_1}} \boldsymbol{\omega}_e - \sum_{e \in \mathcal{E}^{\rm{in}}_{v_1}\backslash\{e_1\}} \boldsymbol{\omega}_e \label{completeFM:ub2}
\end{align}
\end{subequations}
Following standard F-M elimination, variables $\boldsymbol{\omega}_{e_1}$ are removed by requiring that each lower bound must be less than or equal to each upper bound. The constraint $y_{e_1}\boldsymbol{v}^+_{e_1} \leq \boldsymbol{1}_{|\Pi|} - y_{e_1}\boldsymbol{v}^-_{e_1}$ is always satisfied by construction when $y_{e_1} \in \{0,1\}$ and $\boldsymbol{v}^+_{e_1}, \boldsymbol{v}^-_{e_1}$ have disjoint non-zero entries. Therefore, we obtain the following 5 meaningful constraints:
\allowdisplaybreaks
\begin{subequations}
\begin{align}
y_{e_1}\boldsymbol{v}^+_{e_1} &\leq \boldsymbol{z}^{\pi} - \sum_{e \in \mathcal{E}^{\rm{out}}_{v_s}\backslash\{e_1\}} \boldsymbol{\omega}_e \label{completeFM:out1}\\
\boldsymbol{z}^{\pi} - \sum_{e \in \mathcal{E}^{\rm{out}}_{v_s}\backslash\{e_1\}} \boldsymbol{\omega}_e &\leq \boldsymbol{1}_{|\Pi|} - y_{e_1}\boldsymbol{v}^-_{e_1} \label{completeFM:out2}\\
y_{e_1}\boldsymbol{v}^+_{e_1} &\leq \sum_{e \in \mathcal{E}^{\rm{out}}_{v_1}} \boldsymbol{\omega}_e - \sum_{e \in \mathcal{E}^{\rm{in}}_{v_1}\backslash\{e_1\}} \boldsymbol{\omega}_e \label{completeFM:out3}\\
\sum_{e \in \mathcal{E}^{\rm{out}}_{v_1}} \boldsymbol{\omega}_e - \sum_{e \in \mathcal{E}^{\rm{in}}_{v_1}\backslash\{e_1\}} \boldsymbol{\omega}_e &\leq \boldsymbol{1}_{|\Pi|} - y_{e_1}\boldsymbol{v}^-_{e_1} \label{completeFM:out4}\\
\boldsymbol{z}^{\pi} - \sum_{e \in \mathcal{E}^{\rm{out}}_{v_s}\backslash\{e_1\}} \boldsymbol{\omega}_e &= \sum_{e \in \mathcal{E}^{\rm{out}}_{v_1}} \boldsymbol{\omega}_e - \sum_{e \in \mathcal{E}^{\rm{in}}_{v_1}\backslash\{e_1\}} \boldsymbol{\omega}_e \label{completeFM:out5}
\end{align}
\end{subequations}
The original flow conservation constraint (11b) is replaced by \eqref{completeFM:out1} and \eqref{completeFM:out2}. When we proceed to eliminate $\boldsymbol{\omega}_{e_2}$ for $e_2 \in \mathcal{E}^{\rm{out}}_{v_s}\backslash\{e_1\}$, \eqref{completeFM:out2} provides lower bounds on $\boldsymbol{\omega}_{e_2}$ that replace \eqref{completeFM:lb1}:
\begin{equation}
\boldsymbol{\omega}_{e_2} \geq \boldsymbol{z}^{\pi} - (\boldsymbol{1}_{|\Pi|} - y_{e_1}\boldsymbol{v}^-_{e_1}) - \sum_{e \in \mathcal{E}^{\rm{out}}_{v_s}\backslash\{e_1,e_2\}} \boldsymbol{\omega}_e
\end{equation}
Constraint \eqref{completeFM:out1} provides upper bounds that replace \eqref{completeFM:ub1}:
\begin{equation}
\boldsymbol{\omega}_{e_2} \leq \boldsymbol{z}^{\pi} - y_{e_1}\boldsymbol{v}^+_{e_1} - \sum_{e \in \mathcal{E}^{\rm{out}}_{v_s}\backslash\{e_1,e_2\}} \boldsymbol{\omega}_e
\end{equation}
When combining with $\boldsymbol{\omega}_{e_2}$'s own convex set constraints, the F-M elimination produces new constraints where the bounds from $v_s$ now include both $y_{e_1}\boldsymbol{v}^+_{e_1}$ and $y_{e_2}\boldsymbol{v}^+_{e_2}$ in the lower bounds, and both $(\boldsymbol{1}_{|\Pi|} - y_{e_1}\boldsymbol{v}^-_{e_1})$ and $(\boldsymbol{1}_{|\Pi|} - y_{e_2}\boldsymbol{v}^-_{e_2})$ in the upper bounds. The same pattern continues for the rest of the edges in $\mathcal{E}^{\rm{out}}_{v_s}$: each elimination adds the corresponding $y_e\boldsymbol{v}^+_e$ terms to the lower bound and $(\boldsymbol{1}_{|\Pi|} - y_e\boldsymbol{v}^-_e)$ terms to the upper bound. After eliminating all edges from $v_s$, we obtain exactly constraint (13) from the main text:
\begin{align}
\boldsymbol{z}^{\pi} \geq \sum_{e \in \mathcal{E}^{\rm{out}}_{v_s}} y_e\boldsymbol{v}^+_e, \quad
\boldsymbol{z}^{\pi} \leq \sum_{e \in \mathcal{E}^{\rm{out}}_{v_s}} (\boldsymbol{1}_{|\Pi|} - y_e\boldsymbol{v}^-_e) \label{completeFM:con1}
\end{align}

Furthermore, the original constraint \eqref{eqn:vertex-b} at $v_1$ is replaced by \eqref{completeFM:out3} and \eqref{completeFM:out4}. As in the proof of Theorem~\ref{theorem1}, we assume all edges from $v_s$ enter the same vertex $v_1$, i.e., $\mathcal{E}^{\rm{out}}_{v_s} = \mathcal{E}^{\rm{in}}_{v_1}$ (see Fig.~\ref{fig:F-M}). With this assumption, constraint \eqref{completeFM:out4} provides lower bounds on $\boldsymbol{\omega}_{e_2}$ that replace \eqref{completeFM:lb2}:
\begin{equation}
\boldsymbol{\omega}_{e_2} \geq \sum_{e \in \mathcal{E}^{\rm{out}}_{v_1}} \boldsymbol{\omega}_e - (\boldsymbol{1}_{|\Pi|} - y_{e_1}\boldsymbol{v}^-_{e_1}) - \sum_{e \in \mathcal{E}^{\rm{in}}_{v_1}\backslash\{e_1,e_2\}} \boldsymbol{\omega}_e 
\end{equation}
Constraint \eqref{completeFM:out3} provides upper bounds that replace \eqref{completeFM:ub2}:
\begin{equation}
\boldsymbol{\omega}_{e_2} \leq \sum_{e \in \mathcal{E}^{\rm{out}}_{v_1}} \boldsymbol{\omega}_e - y_{e_1}\boldsymbol{v}^+_{e_1} - \sum_{e \in \mathcal{E}^{\rm{in}}_{v_1}\backslash\{e_1,e_2\}} \boldsymbol{\omega}_e 
\end{equation}

The same pattern continues for the rest of the edges in $\mathcal{E}^{\rm{out}}_{v_s}$. After eliminating all edges from $v_s$ (which are also all edges entering $v_1$), we obtain:
\begin{align}
\sum_{e \in \mathcal{E}^{\rm{in}}_{v_1}} y_e\boldsymbol{v}^+_e \leq \sum_{e \in \mathcal{E}^{\rm{out}}_{v_1}} \boldsymbol{\omega}_e \leq \sum_{e \in \mathcal{E}^{\rm{in}}_{v_1}} (\boldsymbol{1}_{|\Pi|} - y_e\boldsymbol{v}^-_e) \label{completeFM:con3}
\end{align}

In addition, the equality constraint \eqref{completeFM:out5} simplifies to:
\begin{equation}
\boldsymbol{z}^{\pi} = \sum_{e \in \mathcal{E}^{\rm{out}}_{v_1}} \boldsymbol{\omega}_e \label{completeFM:con5}
\end{equation}

\xuan{The elimination process up to $\mathcal{E}^{\rm{out}}_{v_s}$ removes \eqref{eqn:edge} for edges $e \in \mathcal{E}^{\rm{out}}_{v_s}$, \eqref{eqn:input-b} at $v_s$, \eqref{eqn:vertex-b} at $v_1$, and replace them with the new constraints \eqref{completeFM:con1}, \eqref{completeFM:con3}, and \eqref{completeFM:con5}.}

We proceed to exclude the flow variables $\boldsymbol{\omega}_e$ for edges $e \in \mathcal{E}^{\rm{out}}_{v_1}$. Consider an edge $e_1' \in \mathcal{E}^{\rm{out}}_{v_1}$ that connects to vertex $v_2$. We have the following constraints involving this edge $e_1'$: constraints \eqref{completeFM:con3}, \eqref{completeFM:con5}, and the original flow conservation constraint (10b) at the subsequent vertex $v_2$. Under a similar assumption as before that all edges outgoing from $v_1$ enter the same vertex $v_2$, i.e., $\mathcal{E}^{\rm{out}}_{v_1} = \mathcal{E}^{\rm{in}}_{v_2}$, we obtain bounds similar to those for $\boldsymbol{\omega}_{e_1}$ with substitutions $e_1 \rightarrow e_1'$, $v_s \rightarrow v_1$, and $v_1 \rightarrow v_2$. Specifically, we have three lower bounds analogous to \eqref{completeFM:lb0}, \eqref{completeFM:lb1}, and \eqref{completeFM:lb2}, three upper bounds analogous to \eqref{completeFM:ub0}, \eqref{completeFM:ub1}, and \eqref{completeFM:ub2}, plus one additional lower bound and one additional upper bound from \eqref{completeFM:con3}:
\begin{subequations}
\begin{align}
\boldsymbol{\omega}_{e_1'} \geq \sum_{e \in \mathcal{E}^{\rm{in}}_{v_1}} y_e\boldsymbol{v}^+_e - \sum_{e \in \mathcal{E}^{\rm{out}}_{v_1} \setminus \{e_1'\}} \boldsymbol{\omega}_e \label{completeFM:add1} \\
\boldsymbol{\omega}_{e_1'} \leq \sum_{e \in \mathcal{E}^{\rm{in}}_{v_1}} (\boldsymbol{1}_{|\Pi|} - y_e\boldsymbol{v}^-_e) - \sum_{e \in \mathcal{E}^{\rm{out}}_{v_1} \setminus \{e_1'\}} \boldsymbol{\omega}_e \label{completeFM:add2}
\end{align}
\end{subequations}

For the first three lower bounds and first three upper bounds, the F-M elimination process at vertex $v_1$ gives the same results as at the source vertex $v_s$, generating constraints analogous to \eqref{completeFM:con1}, \eqref{completeFM:con3}, and \eqref{completeFM:con5}.

However, in this case we have additional bounds \eqref{completeFM:add1} and \eqref{completeFM:add2}, which generate 4 additional constraints:
\begin{subequations}
\begin{align}
y_{e_1'}\boldsymbol{v}^+_{e_1'} \leq \sum_{e \in \mathcal{E}^{\rm{in}}_{v_1}} (\boldsymbol{1}_{|\Pi|} - y_e\boldsymbol{v}^-_e) - \sum_{e \in \mathcal{E}^{\rm{out}}_{v_1} \setminus \{e_1'\}} \boldsymbol{\omega}_e \label{eqn:completeFM:con21} \\
\sum_{e \in \mathcal{E}^{\rm{in}}_{v_1}} y_e\boldsymbol{v}^+_e - \sum_{e \in \mathcal{E}^{\rm{out}}_{v_1} \setminus \{e_1'\}} \boldsymbol{\omega}_e \leq \boldsymbol{1}_{|\Pi|} - y_{e_1'}\boldsymbol{v}^-_{e_1'} \label{eqn:completeFM:con22} \\
\sum_{e \in \mathcal{E}^{\rm{in}}_{v_1}} y_e\boldsymbol{v}^+_e \leq \boldsymbol{z}^{\pi} \leq \sum_{e \in \mathcal{E}^{\rm{in}}_{v_1}} (\boldsymbol{1}_{|\Pi|} - y_e\boldsymbol{v}^-_e) \label{eqn:completeFM:con23} \\
\sum_{e \in \mathcal{E}^{\rm{in}}_{v_1}} y_e\boldsymbol{v}^+_e \leq \sum_{e \in \mathcal{E}^{\rm{out}}_{v_2}} \boldsymbol{\omega}_e \leq \sum_{e \in \mathcal{E}^{\rm{in}}_{v_1}} (\boldsymbol{1}_{|\Pi|} - y_e\boldsymbol{v}^-_e) \label{eqn:completeFM:con24}
\end{align}
\end{subequations}
We first observe that \eqref{eqn:completeFM:con23} is redundant, as it is identical to constraint~\eqref{completeFM:con1} given our assumption that $\mathcal{E}^{\rm{out}}_{v_s} = \mathcal{E}^{\rm{in}}_{v_1}$.

Additionally, constraint~\eqref{eqn:completeFM:con24} is independent of any other flow variables in $\mathcal{E}^{\rm{out}}_{v_1}$, and has a similar form to constraint~\eqref{completeFM:con3} from the elimination at $v_s$, with the only difference being that $\sum_{e \in \mathcal{E}^{\rm{out}}_{v_2}} \boldsymbol{\omega}_e$ replaces $\sum_{e \in \mathcal{E}^{\rm{out}}_{v_1}} \boldsymbol{\omega}_e$.

Regarding constraints~\eqref{eqn:completeFM:con21} and~\eqref{eqn:completeFM:con22}, when we proceed to get rid of $\boldsymbol{\omega}_{e_2'}$ for $e_2' \in \mathcal{E}^{\rm{out}}_{v_1} \setminus \{e_1'\}$, they provide bounds on $\boldsymbol{\omega}_{e_2'}$. When combining with $\boldsymbol{\omega}_{e_2'}$'s own convex set constraints, the F-M elimination yields new constraints where $y_{e_2'}\boldsymbol{v}^+_{e_2'}$ is added to the accumulated lower bounds and $(\boldsymbol{1}_{|\Pi|} - y_{e_2'}\boldsymbol{v}^-_{e_2'})$ is added to the accumulated upper bounds. The same pattern continues for the rest of the edges $e \in \mathcal{E}^{\rm{out}}_{v_1}$. After eliminating all edges from $\mathcal{E}^{\rm{out}}_{v_1}$, we obtain two constraints in the form of \eqref{eqn:ineq_contradict} in the main content.


\xuan{The elimination process up to $\mathcal{E}^{\rm{out}}_{v_1}$ removes constraints~\eqref{eqn:edge} for edges $e \in \mathcal{E}^{\rm{out}}_{v_1}$, \eqref{completeFM:con5} at $v_1$, and \eqref{eqn:vertex-b} at $v_2$, retains \eqref{completeFM:con3} from the previous step at $v_s$, and replaces them with new constraints similar to \eqref{completeFM:con1}, \eqref{completeFM:con3}, and \eqref{completeFM:con5} (with $v_s$ replaced by $v_1$ and $v_1$ replaced by $v_2$), along with \eqref{eqn:ineq_contradict}.}

Continuing this elimination process through the network until reaching the final vertex before $v_t$, denoted $v_{t-1}$, we systematically eliminate all convex set constraints \eqref{eqn:edge} for edges $e \in \mathcal{E}$, all flow conservation constraints \eqref{eqn:vertex-b} at vertices $v \in \mathcal{V} \setminus \{v_s, v_t\}$, and the input flow constraint \eqref{eqn:input-b} at $v_s$. 

At each intermediate vertex $v_i$, we also obtain new constraints analogous to \eqref{completeFM:con3}, but with $\sum_{e \in \mathcal{E}^{\rm{out}}_{v_i}} \boldsymbol{\omega}_e$ in place of $\sum_{e \in \mathcal{E}^{\rm{out}}_{v_1}} \boldsymbol{\omega}_e$.
These constraints collect the bounds from all previous vertices. When we reach $v_{t-1}$, the flow conservation ensures $\sum_{e \in \mathcal{E}^{\rm{out}}_{v_{t-1}}} \boldsymbol{\omega}_e = \boldsymbol{z}^{\pi}$. At this point, all these accumulated constraints become redundant as they simply provide bounds on $\boldsymbol{z}^{\pi}$ that have already been captured by \eqref{completeFM:con1} derived at each $v_i$, and therefore can be eliminated from the final formulation.

We finally obtain constraints in the form of \eqref{completeFM:con1} at each vertex $v \in \mathcal{V} \setminus \{v_t\}$, and constraints in the form of \eqref{eqn:ineq_contradict} at each vertex $v \in \mathcal{V} \setminus \{v_s, v_t\}$, thus achieving the simplified formulation presented in Section~\ref{sect:variable_elimination} of the main text.

\section{Flow Variable Elimination Preserves the Convex Relaxation Tightness}
\label{appendix_epigraph}
In this section, we present a lemma showing that replacing constraint \eqref{eqn:edge} and \eqref{eqn:input-b} with \eqref{eqn:new_ineq_flow_conserv} preserves the tightness of the convex relaxation.

\begin{lemma}
Consider the following optimization problem (denoted by problem A) with flow variables:
\begin{alignat}{2}
\hspace{1.8em} & \hspace{-1.8em} \underset{\substack{\boldsymbol{z}^{\pi}, y_e, \boldsymbol{\omega}_e \text{ for } e \in \mathcal{E}^{\rm{out}}_{v_s}, \\ \boldsymbol{\xi}, \text{ other } \boldsymbol{y}, \boldsymbol{\omega}}}{\textit{minimize}} \hspace{-2em} && \hspace{2em} \quad f_{\text{obj}}(\boldsymbol{\xi}, \boldsymbol{z}^{\pi}) \nonumber\\
&\textit{s.t.} \quad && \boldsymbol{z}^{\pi} = \sum_{e \in \mathcal{E}^{\rm{out}}_{v_s}} \boldsymbol{\omega}_e \tag{A1} \\
& && \boldsymbol{\omega}_e \geq y_e \boldsymbol{v}^+_e, \quad \forall e \in \mathcal{E}^{\rm{out}}_{v_s} \tag{A2} \\
& && \text{other constraints} \nonumber\\
& && \boldsymbol{z}^{\pi}, y_e \text{ are binary variables} \nonumber
\end{alignat}
This problem has the same tightness of convex relaxation as the following problem after flow variable elimination (denoted by problem B):
\begin{alignat}{2}
\hspace{1.1em} & \hspace{-1.1em} \underset{\substack{\boldsymbol{z}^{\pi}, y_e \text{ for } e \in \mathcal{E}^{\rm{out}}_{v_s}, \\ \boldsymbol{\xi}, \text{ other } \boldsymbol{y}, \boldsymbol{\omega}}}{\textit{minimize}} \hspace{-2em} && \hspace{2em} \quad f_{\text{obj}}(\boldsymbol{\xi}, \boldsymbol{z}^{\pi}) \nonumber \\
&\textit{s.t.} \quad && \boldsymbol{z}^{\pi} \geq \sum_{e \in \mathcal{E}^{\rm{out}}_{v_s}} y_e \boldsymbol{v}^+_e \tag{B1} \\
& && \text{other constraints} \nonumber \\
& && \boldsymbol{z}^{\pi}, y_e \text{ are binary variables} \nonumber
\end{alignat}
\end{lemma}
\begin{proof}
    Consider the convex relaxation of problems A and B, where the binary constraints are removed. We show that the optimal cost $f^*_A = f^*_B$. 
    
    \xuan{To establish that $f^*_A \geq f^*_B$, consider an optimal solution to problem A that satisfies constraints (A1) and (A2). Since constraints (A1) and (A2) can be combined to obtain constraint (B1), any feasible solution to problem A provides a feasible solution to problem B with the same values for all shared variables, including $\boldsymbol{z}^{\pi*}$ and $y_e^*$. Therefore, $f^*_A \geq f^*_B$.}

    \xuan{We next show that $f^*_A \leq f^*_B$. Consider an optimal solution to problem B containing variables $(\boldsymbol{z}^{\pi*}, y_e^*, \ldots)$ that satisfy constraint (B1). We show that there exist flow variables $\boldsymbol{\omega}_e$ such that $(\boldsymbol{z}^{\pi*}, y_e^*, \boldsymbol{\omega}_e, \ldots)$ satisfies constraints (A1) and (A2) of problem A.}

    \xuan{To construct such $\boldsymbol{\omega}_e$, define the slack variable $\boldsymbol{\delta} = \boldsymbol{z}^{\pi*} - \sum_{e \in \mathcal{E}^{\rm{out}}_{v_s}} y_e^* \boldsymbol{v}^+_e \geq \boldsymbol{0}$, where the inequality follows from (B1). Let $\boldsymbol{\omega}_e = y_e^* \boldsymbol{v}^+_e + \boldsymbol{\delta}/|\mathcal{E}^{\rm{out}}_{v_s}|$. This construction satisfies constraint (A1):}
    \begin{align*}
    \sum_{e \in \mathcal{E}^{\rm{out}}_{v_s}} \boldsymbol{\omega}_e &= \sum_{e \in \mathcal{E}^{\rm{out}}_{v_s}} \left(y_e^* \boldsymbol{v}^+_e + \boldsymbol{\delta}/|\mathcal{E}^{\rm{out}}_{v_s}|\right) \\
    &= \sum_{e \in \mathcal{E}^{\rm{out}}_{v_s}} y_e^* \boldsymbol{v}^+_e + \boldsymbol{\delta} = \boldsymbol{z}^{\pi*}
    \end{align*}
    \xuan{and constraint (A2):}
    \begin{align*}
    \boldsymbol{\omega}_e = y_e^* \boldsymbol{v}^+_e + \boldsymbol{\delta}/|\mathcal{E}^{\rm{out}}_{v_s}| \geq y_e^* \boldsymbol{v}^+_e
    \end{align*}
    
    \xuan{Since the objective function is independent of $\boldsymbol{\omega}_e$, any optimal solution to problem B gives a feasible solution to problem A with the same objective value. Therefore, $f^*_A \leq f^*_B$.}\qed
\end{proof}

\section{Fourier-Motzkin Elimination for GCS Formulation of Minimum-Time Linear Control}
\label{app:gcs_application}
In this appendix, we provide an example demonstrating how the proposed F-M elimination process can be applied to simplify GCS formulations for problems beyond temporal logic planning. We illustrate this by applying the F-M elimination to the minimum-time control problem for discrete-time linear systems with double integrator dynamics, following the example from Chapter 10.1 of \citep{marcucci2024graphs}. Consider the discrete-time linear dynamical system:
\begin{equation}
    \boldsymbol{s}_{k+1} = \boldsymbol{A}\boldsymbol{s}_k + \boldsymbol{B}\boldsymbol{a}_k \label{eqn:min_time_control_dynamics}
\end{equation}
where $\boldsymbol{s}_k \in \mathbb{R}^n$ and $\boldsymbol{a}_k \in \mathbb{R}^m$ are the system state and control action at time step $k \in \mathbb{Z}_{\geq 0}$. For the double integrator dynamics with discretization step $0.01$, the system matrices are:
\begin{equation}
\boldsymbol{A} = \begin{bmatrix} 1 & 0.01 \\ 0 & 1 \end{bmatrix}, \quad 
\boldsymbol{B} = \begin{bmatrix} 0 \\ 0.01 \end{bmatrix}
\end{equation}

Given initial conditions $\boldsymbol{s}_0 = \hat{\boldsymbol{s}}$, the goal is to drive the system to a target set $\mathcal{T} = \{\boldsymbol{0}\}$ (origin) in the minimum number of time steps $k \leq \bar{K}$, subject to constraints $(\boldsymbol{s}_k, \boldsymbol{a}_k) \in \mathcal{D}$ where $\mathcal{D}$ is a compact convex set. For this problem, we specify $\mathcal{D}$ by element-wise bounds $\boldsymbol{s}_{\text{lb}} \leq \boldsymbol{s}_k \leq \boldsymbol{s}_{\text{ub}}$ and $\boldsymbol{a}_{\text{lb}} \leq \boldsymbol{a}_k \leq \boldsymbol{a}_{\text{ub}}$, where $\boldsymbol{s}_{\text{lb}}, \boldsymbol{s}_{\text{ub}} \in \mathbb{R}^n$ and $\boldsymbol{a}_{\text{lb}}, \boldsymbol{a}_{\text{ub}} \in \mathbb{R}^m$.

\begin{figure*}[t!]
    \centering
    \includegraphics[width=0.85\textwidth]{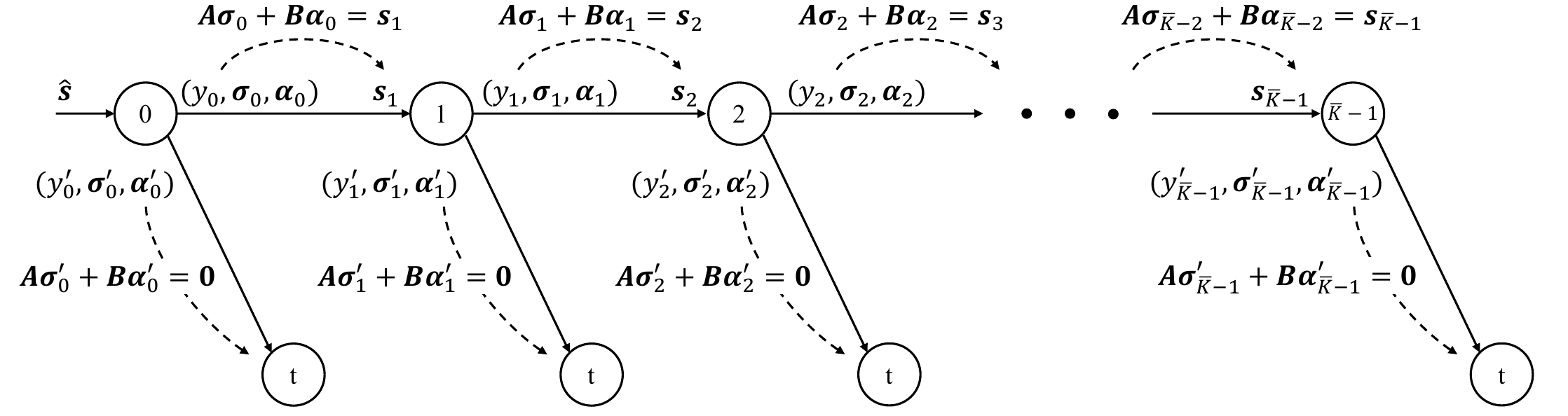}
     \caption{GCS representation for minimum-time control. The graph structure shows vertices $\{0, 1, \ldots, \bar{K}-1, t\}$ with edge variables $y_k$ and $y'_k$, flow variables $\boldsymbol{\sigma}_k, \boldsymbol{\alpha}_k, \boldsymbol{\sigma}'_k, \boldsymbol{\alpha}'_k$, initial state $\hat{\boldsymbol{s}}$ at vertex 0, and target state at vertex $t$.}
\label{fig:min_time_control_gcs}
\end{figure*} 

Following \citep{marcucci2024graphs}, the GCS formulation constructs a graph with vertices $\mathcal{V} = \{0, 1, \ldots, \bar{K}-1, t\}$ where each vertex $v = 0, \ldots, \bar{K}-1$ has outgoing edges to vertex $v+1$ and to the target vertex $t$. The formulation introduces binary edge variables $y_e \in \{0,1\}$ indicating if edge $e$ is traversed, and continuous variables representing state-control pairs $(\boldsymbol{s}_v, \boldsymbol{a}_v) \in \mathcal{D}$ at each vertex, with the state $\boldsymbol{s}_0 = \hat{\boldsymbol{s}}$ at the source vertex and $\boldsymbol{s}_t \in \mathcal{T}$ at the target vertex $t$. Figure~\ref{fig:min_time_control_gcs} illustrates this graph structure.

The complete GCS formulation is:
\begin{subequations}
\begin{alignat}{2}
&\underset{}{\textit{minimize}} \hspace{-1em} && \hspace{1em} \quad  1 + \sum_{k=0}^{\bar{K}-2} y_k \nonumber \\
&\textit{s.t.} \quad && y_0 + y'_0 = 1 \label{eq:gcs_init}\\
& && y_{\bar{K}-1} = 0 \label{eq:gcs_term}\\
& && \forall k = 0, \ldots, \bar{K}-2: \nonumber \\
& && \quad y_{k+1} + y'_{k+1} = y_k \label{eq:gcs_flow}\\
& && \boldsymbol{\sigma}_0 + \boldsymbol{\sigma}'_0 = \hat{\boldsymbol{s}} \label{eq:gcs_initial_state}\\
& && \forall k = 0, \ldots, \bar{K}-1: \label{eq:gcs_convex_bound_constraints}\\
& && \quad y_k \boldsymbol{s}_{\text{lb}} \leq \boldsymbol{\sigma}_k \leq y_k \boldsymbol{s}_{\text{ub}} \nonumber\\
& && \quad y_k \boldsymbol{a}_{\text{lb}} \leq \boldsymbol{\alpha}_k \leq y_k \boldsymbol{a}_{\text{ub}} \nonumber\\
& && \quad y'_k \boldsymbol{s}_{\text{lb}} \leq \boldsymbol{\sigma}'_k \leq y'_k \boldsymbol{s}_{\text{ub}} \nonumber\\
& && \quad y'_k \boldsymbol{a}_{\text{lb}} \leq \boldsymbol{\alpha}'_k \leq y'_k \boldsymbol{a}_{\text{ub}} \nonumber\\
& && \forall k = 0, \ldots, \bar{K}-2: \nonumber\\
& && \quad \boldsymbol{A}\boldsymbol{\sigma}_k + \boldsymbol{B}\boldsymbol{\alpha}_k = \boldsymbol{\sigma}_{k+1} + \boldsymbol{\sigma}'_{k+1} \label{eq:gcs_dynamics}\\
& && \forall k = 0, \ldots, \bar{K}-1: \nonumber\\
& && \quad \boldsymbol{A}\boldsymbol{\sigma}'_k + \boldsymbol{B}\boldsymbol{\alpha}'_k = \boldsymbol{0} \label{eq:gcs_target}
\end{alignat}
\label{eqn:min_time_GCS}
\end{subequations}

\noindent where $y_k$ is the edge variable from vertex $k$ to vertex $k+1$, $y'_k$ is the edge variable from vertex $k$ to the target vertex $t$. The variables $\boldsymbol{\sigma}_k, \boldsymbol{\alpha}_k$ are the state and control flow variables paired with $y_k$, while $\boldsymbol{\sigma}'_k, \boldsymbol{\alpha}'_k$ are the state and control flow variables paired with $y'_k$. The actual system state and control at time step $k$ are reconstructed as:
\begin{equation}
(\boldsymbol{s}_k, \boldsymbol{a}_k) = (\boldsymbol{\sigma}_k, \boldsymbol{\alpha}_k) + (\boldsymbol{\sigma}'_k, \boldsymbol{\alpha}'_k) \label{eqn:min_time_control_state_reconstruction}
\end{equation}
For comparison, the baseline formulation used by \citep{marcucci2024graphs} is:
\begin{subequations}
\begin{alignat}{2}
&\underset{}{\textit{minimize}} \hspace{-1em} && \hspace{1em} \quad 1 + \sum_{k=0}^{\bar{K}-2} y_k \nonumber \\
&\textit{s.t.} \quad && \boldsymbol{s}_0 = \hat{\boldsymbol{s}} \label{eq:baseline_init}\\
& && \boldsymbol{s}_{\bar{K}} = \boldsymbol{0} \label{eq:baseline_target}\\
& && \boldsymbol{a}_{\text{lb}} \leq \boldsymbol{a}_0 \leq \boldsymbol{a}_{\text{ub}} \label{eq:baseline_init_control}\\
& && \forall k = 1, \ldots, \bar{K}-2: \label{eq:baseline_bound_constraints}\\
& && \quad y_{k-1} \boldsymbol{s}_{\text{lb}} \leq \boldsymbol{s}_k \leq y_{k-1} \boldsymbol{s}_{\text{ub}} \nonumber\\
& && \quad y_{k-1} \boldsymbol{a}_{\text{lb}} \leq \boldsymbol{a}_k \leq y_{k-1} \boldsymbol{a}_{\text{ub}} \nonumber\\
& && \forall k = 0, \ldots, \bar{K}-1 \nonumber\\
& && \quad \boldsymbol{s}_{k+1} = \boldsymbol{A}\boldsymbol{s}_k + \boldsymbol{B}\boldsymbol{a}_k \label{eq:baseline_dynamics}
\end{alignat}
\label{eqn:min_time_baseline}\\
\end{subequations}
%
\noindent While GCS achieves tighter convex relaxations than the baseline formulation, this advantage comes at the cost of additional flow variables. These extra variables offset the benefits of tighter relaxations, resulting in no computational speed improvement over the baseline for this problem. However, by applying F-M elimination, we can remove the flow variables from the GCS formulation while preserving the relaxation tightness, resulting in a formulation with nearly the same number of variables as the baseline but with tighter convex relaxations.

Since this problem involves linear dynamics as described in \eqref{eqn:min_time_control_dynamics}, the flow conservation constraints propagate the dynamics through the network flow structure via \xuanrev{Eqns.} \eqref{eq:gcs_dynamics} and \eqref{eq:gcs_target}. However, eliminating the flow variables would disrupt this dynamic propagation. To preserve the dynamic propagation, we reintroduce state and control variables $\boldsymbol{s}_k$ and $\boldsymbol{a}_k$ that satisfy the dynamics \eqref{eqn:min_time_control_dynamics} and their bounds \eqref{eq:baseline_bound_constraints}. The state variables are labeled at each vertex in Fig.~\ref{fig:min_time_control_gcs}.

Starting with $k=0$, we systematically eliminate the flow variables $\boldsymbol{\sigma}_0$, $\boldsymbol{\alpha}_0$, $\boldsymbol{\sigma}'_0$, and $\boldsymbol{\alpha}'_0$. The constraints involving these variables include the initial state constraint \eqref{eq:gcs_initial_state}, the dynamics constraints \eqref{eq:gcs_dynamics} and \eqref{eq:gcs_target} at $k=0$, and the convex bounds \eqref{eq:gcs_convex_bound_constraints} at $k=0$. Since the dynamics constraints naturally involve $\boldsymbol{A}\boldsymbol{\sigma}_0$, $\boldsymbol{B}\boldsymbol{\alpha}_0$, $\boldsymbol{A}\boldsymbol{\sigma}'_0$, and $\boldsymbol{B}\boldsymbol{\alpha}'_0$, we multiply the bounds by $\boldsymbol{A}$ or $\boldsymbol{B}$ accordingly to align with these compound terms. This multiplication preserves inequality directions for the double integrator system since all elements of $\boldsymbol{A}$ and $\boldsymbol{B}$ are non-negative. The resulting aligned constraints are:
\begin{subequations}
\begin{align}
& \boldsymbol{A}\boldsymbol{\sigma}_0 + \boldsymbol{A}\boldsymbol{\sigma}'_0 = \boldsymbol{A}\hat{\boldsymbol{s}} \\
& \boldsymbol{A}\boldsymbol{\sigma}_0 + \boldsymbol{B}\boldsymbol{\alpha}_0 = \boldsymbol{s}_1 \\
& \boldsymbol{A}\boldsymbol{\sigma}'_0 + \boldsymbol{B}\boldsymbol{\alpha}'_0 = \boldsymbol{0} \label{eqn:min_time_control_origin_dynamics} \\
y_0 &\boldsymbol{A} \boldsymbol{s}_{\text{lb}} \leq \boldsymbol{A} \boldsymbol{\sigma}_0 \leq y_0 \boldsymbol{A} \boldsymbol{s}_{\text{ub}}\\
y'_0 &\boldsymbol{A} \boldsymbol{s}_{\text{lb}} \leq \boldsymbol{A} \boldsymbol{\sigma}'_0 \leq y'_0 \boldsymbol{A} \boldsymbol{s}_{\text{ub}} \label{eqn:aligned_bound_sigma0_prime}\\
y_0 &\boldsymbol{B} \boldsymbol{a}_{\text{lb}} \leq \boldsymbol{B} \boldsymbol{\alpha}_0 \leq y_0 \boldsymbol{B} \boldsymbol{a}_{\text{ub}} \label{eqn:aligned_bound_alpha0}\\ 
y'_0 &\boldsymbol{B} \boldsymbol{a}_{\text{lb}} \leq \boldsymbol{B} \boldsymbol{\alpha}'_0 \leq y'_0 \boldsymbol{B} \boldsymbol{a}_{\text{ub}} \label{eqn:aligned_bound_alpha0_prime}
\end{align}
\end{subequations}
\begin{table*}[h]
\centering
\caption{Minimum-time control for double integrator on 63 initial conditions}
\label{table:min_time_results}
\footnotesize
\begin{tabular}{lcccc}
\toprule
\textbf{Formulation} & \textbf{Variables} & \textbf{Constraints} & \textbf{Relaxation Gap (\%)} & \textbf{Solving Time (s)} \\
& (Binary, Continuous) & & Mean [Min, Max] & Mean [Max] \\
\midrule
GCS \eqref{eqn:min_time_GCS} & $(500, 1500)$ & $4251$ & $33.4$ $[4.1, 64.8]$ & $0.32$ $[2.27]$ \\
F-M Simplified GCS \eqref{eqn:min_time_GCS_simplified} & $(500, 752)$ & $2499$ & $33.5$ $[4.1, 64.8]$ & $0.08$ $[0.45]$ \\
Baseline \eqref{eqn:min_time_baseline} & $(250, 752)$ & $1994$ & $40.7$ $[19.1, 68.5]$ & $0.07$ $[0.78]$ \\
\bottomrule
\end{tabular}
\end{table*}
We begin by canceling out $\boldsymbol{\sigma}_0$. Applying the standard F-M procedure, all lower and upper bounds for $\boldsymbol{A}\boldsymbol{\sigma}_0$ from the aligned constraints are enumerated, then valid inequalities are generated by requiring each lower bound to be less than or equal to each upper bound. This process yields the following non-trivial constraints:
\begin{subequations}\label{eq:sigma0_elimination}
\begin{align}
y_0 \boldsymbol{A}\boldsymbol{s}_{\text{lb}} + \boldsymbol{A}\boldsymbol{\sigma}'_0 &\leq \boldsymbol{A}\hat{\boldsymbol{s}} \\
y_0 \boldsymbol{A}\boldsymbol{s}_{\text{lb}} &\leq \boldsymbol{s}_1 - \boldsymbol{B}\boldsymbol{\alpha}_0 \\
\boldsymbol{A}\hat{\boldsymbol{s}} - \boldsymbol{A}\boldsymbol{\sigma}'_0 &\leq y_0 \boldsymbol{A}\boldsymbol{s}_{\text{ub}} \\
\boldsymbol{A}\hat{\boldsymbol{s}} - \boldsymbol{A}\boldsymbol{\sigma}'_0 &\leq \boldsymbol{s}_1 - \boldsymbol{B}\boldsymbol{\alpha}_0 \\
\boldsymbol{s}_1 - \boldsymbol{B}\boldsymbol{\alpha}_0 &\leq y_0 \boldsymbol{A}\boldsymbol{s}_{\text{ub}} \\
\boldsymbol{s}_1 - \boldsymbol{B}\boldsymbol{\alpha}_0 &\leq \boldsymbol{A}\hat{\boldsymbol{s}} - \boldsymbol{A}\boldsymbol{\sigma}'_0
\end{align}
\end{subequations}

Next, we eliminate $\boldsymbol{\alpha}_0$ using the constraints from \eqref{eq:sigma0_elimination} along with the convex bounds \eqref{eqn:aligned_bound_alpha0} for $\boldsymbol{B}\boldsymbol{\alpha}_0$. Applying F-M elimination gives the following non-trivial constraints:
\begin{subequations}\label{eq:alpha0_elimination}
\begin{align}
y_0 \boldsymbol{B}\boldsymbol{a}_{\text{lb}} - \boldsymbol{A}\boldsymbol{\sigma}'_0 &\leq \boldsymbol{B}\boldsymbol{a}_0 \label{eq:alpha0_elimination_b}\\
\boldsymbol{B}\boldsymbol{a}_0 &\leq - \boldsymbol{A}\boldsymbol{\sigma}'_0 + y_0 \boldsymbol{B}\boldsymbol{a}_{\text{ub}} \label{eq:alpha0_elimination_d}
\end{align}
\end{subequations}

Following that, we remove $\boldsymbol{\sigma}'_0$ using the constraints from \eqref{eq:alpha0_elimination} along with the convex bounds \eqref{eqn:aligned_bound_sigma0_prime} and the dynamics constraint \eqref{eqn:min_time_control_origin_dynamics}. Applying F-M elimination gives the following non-trivial constraints:
\begin{subequations}\label{eq:sigma0_prime_elimination}
\begin{align}
y'_0 \boldsymbol{A}\boldsymbol{s}_{\text{lb}} + \boldsymbol{B}\boldsymbol{a}_0 &\leq y_0 \boldsymbol{B}\boldsymbol{a}_{\text{ub}} \label{eq:sigma0_prime_elimination_a}\\
y'_0 \boldsymbol{A}\boldsymbol{s}_{\text{lb}} &\leq -\boldsymbol{B}\boldsymbol{\alpha}'_0 \label{eq:sigma0_prime_elimination_b}\\
\boldsymbol{A}\hat{\boldsymbol{s}} - y_0 \boldsymbol{A}\boldsymbol{s}_{\text{ub}} &\leq -\boldsymbol{B}\boldsymbol{\alpha}'_0 \label{eq:sigma0_prime_elimination_c}\\
y_0 \boldsymbol{B}\boldsymbol{a}_{\text{lb}} &\leq \boldsymbol{B}\boldsymbol{a}_0 + y'_0 \boldsymbol{A}\boldsymbol{s}_{\text{ub}} \label{eq:sigma0_prime_elimination_d}\\
y_0 \boldsymbol{B}\boldsymbol{a}_{\text{lb}} &\leq \boldsymbol{B}\boldsymbol{a}_0 -\boldsymbol{B}\boldsymbol{\alpha}'_0 \label{eq:sigma0_prime_elimination_e}\\
-\boldsymbol{B}\boldsymbol{\alpha}'_0 &\leq y'_0 \boldsymbol{A}\boldsymbol{s}_{\text{ub}} \label{eq:sigma0_prime_elimination_f}\\
-\boldsymbol{B}\boldsymbol{\alpha}'_0 &\leq \boldsymbol{A}\hat{\boldsymbol{s}} - y_0 \boldsymbol{A}\boldsymbol{s}_{\text{lb}} \label{eq:sigma0_prime_elimination_g}\\
-\boldsymbol{B}\boldsymbol{\alpha}'_0 + \boldsymbol{B}\boldsymbol{a}_0 &\leq y_0 \boldsymbol{B}\boldsymbol{a}_{\text{ub}} \label{eq:sigma0_prime_elimination_h}
\end{align}
\end{subequations}
Among these inequalities, constraints \eqref{eq:sigma0_prime_elimination_a} and \eqref{eq:sigma0_prime_elimination_d} are valid inequalities for the final formulation.

Finally, we eliminate $\boldsymbol{\alpha}'_0$ from \eqref{eq:sigma0_prime_elimination} (other than \eqref{eq:sigma0_prime_elimination_a} and \eqref{eq:sigma0_prime_elimination_d}) along with the convex bounds \eqref{eqn:aligned_bound_alpha0_prime}. F-M gives two additional valid inequalities:
\begin{subequations}\label{eq:alpha0_prime_elimination}
\begin{align}
\boldsymbol{A}\hat{\boldsymbol{s}} + y'_0 \boldsymbol{B}\boldsymbol{a}_{\text{lb}} &\leq y_0 \boldsymbol{A}\boldsymbol{s}_{\text{ub}} \label{eq:alpha0_prime_elimination_a} \\
y_0 \boldsymbol{A}\boldsymbol{s}_{\text{lb}} &\leq \boldsymbol{A}\hat{\boldsymbol{s}} + y'_0 \boldsymbol{B}\boldsymbol{a}_{\text{ub}} \label{eq:alpha0_prime_elimination_b}
\end{align}
\end{subequations}
Constraints \eqref{eq:alpha0_prime_elimination_a} and \eqref{eq:alpha0_prime_elimination_b}, together with \eqref{eq:sigma0_prime_elimination_a}, and \eqref{eq:sigma0_prime_elimination_d}, form the complete set of valid inequalities that tightens the formulation for $k=0$.

Following an identical pattern, we repeat this elimination process to remove flow variables $\boldsymbol{\sigma}_k$, $\boldsymbol{\alpha}_k$, $\boldsymbol{\sigma}'_k$, and $\boldsymbol{\alpha}'_k$ for $k=1, 2, \ldots, \bar{K}-1$. After completing the elimination at all time steps, we obtain the F-M simplified GCS formulation:
\begin{subequations}
\begin{alignat}{2}
&\underset{}{\textit{minimize}} \hspace{-1em} && \hspace{1em} \quad 1 + \sum_{k=0}^{\bar{K}-2} y_k \nonumber \\
&\textit{s.t.} \quad && y_0 + y'_0 = 1 \label{eq:simplified_init}\\
& && y_{\bar{K}-1} = 0 \label{eq:simplified_term}\\
& && \forall k = 0, \ldots, \bar{K}-2: \nonumber \\
& && \quad y_{k+1} + y'_{k+1} = y_k \label{eq:simplified_flow}\\
& && \boldsymbol{s}_0 = \hat{\boldsymbol{s}}, \quad \boldsymbol{s}_{\bar{K}} = \boldsymbol{0} \label{eq:simplified_initial_state}\\
& && \boldsymbol{a}_{\text{lb}} \leq \boldsymbol{a}_0 \leq \boldsymbol{a}_{\text{ub}} \\
& && \forall k = 1, \ldots, \bar{K}-2: \\
& && \quad y_{k-1} \boldsymbol{s}_{\text{lb}} \leq \boldsymbol{s}_k \leq y_{k-1} \boldsymbol{s}_{\text{ub}} \nonumber\\
& && \quad y_{k-1} \boldsymbol{a}_{\text{lb}} \leq \boldsymbol{a}_k \leq y_{k-1} \boldsymbol{a}_{\text{ub}} \nonumber\\
& && \forall k = 0, \ldots, \bar{K}-1 \nonumber\\
& && \quad \boldsymbol{s}_{k+1} = \boldsymbol{A}\boldsymbol{s}_k + \boldsymbol{B}\boldsymbol{a}_k \\
& && \forall k = 0, \ldots, \bar{K}-1: \nonumber \\
& && \quad y'_k \boldsymbol{A}\boldsymbol{s}_{\text{lb}} + \boldsymbol{B}\boldsymbol{a}_k \leq y_k \boldsymbol{B}\boldsymbol{a}_{\text{ub}} \label{eq:simplified_tightening3}\\
& && \quad y_k \boldsymbol{B}\boldsymbol{a}_{\text{lb}} \leq y'_k \boldsymbol{A}\boldsymbol{s}_{\text{ub}} + \boldsymbol{B}\boldsymbol{a}_k \label{eq:simplified_tightening4}\\
& && \quad \boldsymbol{A}\boldsymbol{s}_k + y'_k \boldsymbol{B}\boldsymbol{a}_{\text{lb}} \leq y_k \boldsymbol{A}\boldsymbol{s}_{\text{ub}} \label{eq:simplified_tightening5}\\
& && \quad y_k \boldsymbol{A}\boldsymbol{s}_{\text{lb}} \leq \boldsymbol{A}\boldsymbol{s}_k + y'_k \boldsymbol{B}\boldsymbol{a}_{\text{ub}} \label{eq:simplified_tightening6}
\end{alignat}
\label{eqn:min_time_GCS_simplified}
\end{subequations}
Comparing \eqref{eqn:min_time_GCS_simplified} with the baseline \eqref{eqn:min_time_baseline}, we observe that the network flow structure used by GCS effectively introduces constraints \eqref{eq:simplified_flow} and \eqref{eq:simplified_tightening3}--\eqref{eq:simplified_tightening6} that tighten the formulation upon the baseline. During our tests, we find that the tightening effect of constraints \eqref{eq:simplified_tightening3}--\eqref{eq:simplified_tightening6} is minimal, hence we exclude them from our experimental formulation. The major tightening effect comes from \eqref{eq:simplified_flow}, which implicitly enforces the valid inequality $y_{k+1} \leq y_k$. Through F-M elimination, we obtain \eqref{eqn:min_time_GCS_simplified}, a formulation equivalent to \eqref{eqn:min_time_GCS} but with significantly fewer variables and constraints.

We use Gurobi 12.0 to run 63 experiments on a grid of initial conditions spanning positions from $-0.8$ to $0.8$ (at $0.2$ intervals), velocities from $-0.75$ to $0.75$ (at $0.25$ intervals), and time horizon $\bar{K}=250$. We compare the GCS \eqref{eqn:min_time_GCS}, F-M simplified GCS \eqref{eqn:min_time_GCS_simplified}, and baseline \eqref{eqn:min_time_baseline} formulations. The results are shown in Table~\ref{table:min_time_results}. F-M simplified GCS achieves a $33.5$\% mean relaxation gap, nearly identical to the original GCS formulation's $33.4$\%, with the $0.1$\% difference attributable to excluding the tightening constraints \eqref{eq:simplified_tightening3}--\eqref{eq:simplified_tightening6}. The baseline formulation exhibits a substantially weaker mean relaxation gap of $40.7$\%. The number of continuous variables in \eqref{eqn:min_time_GCS_simplified} is identical to the baseline \eqref{eqn:min_time_baseline}, and significantly less than the original GCS. The additional constraints \eqref{eq:simplified_flow} increase the number of constraints over the baseline. In terms of computation time, the F-M simplification increases the solving time over the original GCS. While the average solving time for the F-M simplified formulation is comparable to the baseline, the maximum solving time is less.

\xuan{Our F-M procedure presented in this appendix has limitations. It requires that the state matrices $\boldsymbol{A}$ and $\boldsymbol{B}$ contain only non-negative elements, and that the convex bounds \eqref{eq:gcs_convex_bound_constraints} can be easily written to isolate the flow variables. These conditions do not hold for the general class of GCS formulations, though coordinate transformations might extend applicability in some cases. When these conditions are satisfied, we have a pipeline to identify strong valid inequalities for the specific problem class: starting with GCS to obtain tight convex relaxations, then applying F-M elimination to remove the flow variables.}

\section{Soundness and Completeness}
\label{appendix_soundness}
In this appendix, we provide soundness and completeness guarantees for our LNF formulation. 
\xuanrev{For all proofs presented, we assume the temporal logic specification takes the conjunctive-disjunctive form $\varphi = \bigwedge_{l=1}^{L} \bigvee_{m=1}^{M_l} \bigwedge_{n=1}^{N_{m,l}} \pi^l_{m,n}$ used in the proof of Theorem~\ref{theorem1}, where $\pi^l_{m,n}$ is a possibly negated predicate. Any propositional logic formula can be converted to this form through iterative application of distributivity and De Morgan's laws \citep{jackson2004clause}. We also assume the same form of objective as in Theorem~\ref{theorem1}, which depends only on $\boldsymbol{\xi}$ and $\boldsymbol{z}^\pi$ and not on the logic-formulation variables.}
\subsection{Soundness}

\begin{theorem}
    If the complete formulation returns solution $(\boldsymbol{\xi}, \boldsymbol{y}, \boldsymbol{z}^{\pi})$, then the solution $\boldsymbol{z}^{\pi}$ satisfies the temporal logic formula $\varphi$, the solution $\boldsymbol{\xi}$ satisfies the underlying system dynamics, and the combination of them satisfies the atomic predicate constraint ~\eqref{eqn:predicate}.  
\end{theorem}

\begin{proof}
    \xuanrev{Let $\varphi = \bigwedge_{l=1}^{L}\bigvee_{m=1}^{M_l}\bigwedge_{n=1}^{N_{m,l}}\pi^l_{m,n}$. For a formula in this form, satisfaction requires that for each $l = 1,\dots,L$, at least one of the conjunctive clauses $\bigwedge_{n=1}^{N_{m,l}}\pi^l_{m,n}$, $m = 1,\dots,M_l$, is satisfied. In our LNF formulation, constraint \eqref{eqn:lnf_1} ensures that within each section $l$, exactly one path (corresponding to one conjunctive clause) is selected.}

    \xuanrev{When a specific path $m$ is selected in section $l$ (i.e., $y_{m,l} = 1$), all predicates in the corresponding conjunctive clause must be satisfied. For each non-negated predicate $\pi_j \in P_{m,l}$, constraint \eqref{eqn:lnf_2} gives $z^{\pi_j} \ge \sum_{m':\,\pi_j \in P_{m',l}} y_{m',l} \ge y_{m,l} = 1$, hence $z^{\pi_j} = 1$. Similarly, for each negated predicate $\neg\pi_j \in P_{m,l}$, constraint \eqref{eqn:lnf_3} gives $z^{\pi_j} \le 1 - \sum_{m':\,\neg\pi_j \in P_{m',l}} y_{m',l} \le 1 - y_{m,l} = 0$, hence $z^{\pi_j} = 0$. Therefore every predicate in the selected clause $\bigwedge_{n=1}^{N_{m,l}}\pi^l_{m,n}$ is satisfied, so the disjunction $\bigvee_{m=1}^{M_l}\bigwedge_{n=1}^{N_{m,l}}\pi^l_{m,n}$ in section $l$ holds; since this is true for every $l = 1,\dots,L$, the full specification $\varphi$ is satisfied.}

    \xuanrev{For paths not selected (i.e., $y_{m,l} = 0$), \eqref{eqn:lnf_5} prevents all predicates in $P_{m,l}$ from being simultaneously satisfied, as this would make the right-hand side equal to $1$.}

    \xuanrev{Finally, \eqref{eqn:lnf_4} guarantees that the underlying dynamics and atomic predicates are satisfied.}
\end{proof}

\subsection{Completeness}

We show that our planner is complete, meaning it always finds the optimal solution when one exists. 

\begin{theorem}
\label{thm:completeness1}
    If there exists a feasible trajectory satisfying the temporal logic constraint, it corresponds to a set of feasible solutions $(\boldsymbol{\xi}, \boldsymbol{y}, \boldsymbol{z}^{\pi})$ of the optimization formulation.
\end{theorem} 

\begin{proof}

\xuanrev{Let $\varphi = \bigwedge_{l=1}^{L}\bigvee_{m=1}^{M_l}\bigwedge_{n=1}^{N_{m,l}}\pi^l_{m,n}$. Assume there exists a feasible trajectory $\boldsymbol{\xi}$ that satisfies $\varphi$. We construct $\boldsymbol{z}^{\pi}$ from this feasible $\boldsymbol{\xi}$ as follows. For each atomic predicate $\pi_j \in \Pi$ and each time step $k$, define $z^{\pi^j_k} = 1$ if and only if $({\boldsymbol{a}^{\pi_j}})^\top \boldsymbol{x}_k + b^{\pi_j} \geq 0$, and $z^{\pi^j_k} = 0$ otherwise. This uniquely determines $\boldsymbol{z}^{\pi}$.}

\xuanrev{We show that there exists $\boldsymbol{y}$ that, together with the $\boldsymbol{z}^{\pi}$ constructed above, satisfies constraints \eqref{eqn:lnf_1}, \eqref{eqn:lnf_2}, and \eqref{eqn:lnf_3}. Since $\varphi$ is a conjunction over $l$, for each $l = 1,\dots,L$ the trajectory satisfies at least one conjunctive clause $\bigwedge_{n=1}^{N_{m,l}}\pi^l_{m,n}$ for some $m \in \{1,\dots,M_l\}$; we set $y_{m,l} = 1$ for one such $m$ and $y_{m,l} = 0$ for all other $m$. This construction immediately satisfies \eqref{eqn:lnf_1} since $\sum_{m=1}^{M_l} y_{m,l} = 1$ for each $l$.}


\xuanrev{For constraint \eqref{eqn:lnf_2}, consider any non-negated predicate $\pi_j \in P_{m,l}$ in the selected clause of section $l$. Since $\boldsymbol{\xi}$ satisfies this clause, $\pi_j$ holds, so $z^{\pi^j_k} = 1$ by the construction of $\boldsymbol{z}^{\pi}$. Because only the selected branch has $y_{m,l}=1$, the right-hand side of \eqref{eqn:lnf_2} equals 1, and \eqref{eqn:lnf_2} is satisfied. A similar argument holds for \eqref{eqn:lnf_3}.}

\xuanrev{For any predicate $\pi_j$ not in the selected clause of section $l$ (neither $\pi_j$ nor $\neg\pi_j$), the only branch with $y_{m,l}=1$ does not contain $\pi_j$, so the right-hand-side sums in \eqref{eqn:lnf_2} and \eqref{eqn:lnf_3} vanish, reducing them to $z^{\pi^j_k} \geq 0$ and $z^{\pi^j_k} \leq 1$, respectively, which hold trivially.}



Since our formulation is a mixed-integer program solved with Branch-and-Bound, the solver guarantees finding this feasible solution.
\end{proof}

\begin{theorem}
    If there exists an optimal trajectory satisfying the temporal logic constraint, it corresponds to a set of optimal solutions $(\boldsymbol{\xi}^*, \boldsymbol{y}^*, \boldsymbol{z}^{\pi*})$ of the optimization formulation.
\end{theorem} 

\begin{proof}
In the proof of Theorem \ref{thm:completeness1}, we have already shown that any solution of $\boldsymbol{\xi}^*$ with uniquely determined $\boldsymbol{z}^{\pi*}$ corresponds to (potentially more than one) set of feasible $\boldsymbol{y}$'s. Therefore, this theorem is evident since the objective function is independent of $\boldsymbol{y}$.
\end{proof}

\end{document}